\documentclass[a4paper,12pt, authoryear]{elsarticle}
\usepackage{amsmath, amssymb}
\usepackage{bm}
\usepackage[table, dvipsnames]{xcolor}
\usepackage{booktabs}
\usepackage{hyperref}
\usepackage{orcidlink}
\usepackage{graphicx}
\usepackage{wrapfig}
\usepackage{tikz}
\usepackage{pgf}
\usepackage{lmodern}
\usepackage{placeins}
\usepackage{pdfpages}
\usetikzlibrary{shapes.geometric}
\usetikzlibrary{fit,positioning}
\usetikzlibrary{arrows}

\usepackage{array}
\newcolumntype{L}[1]{>{\raggedleft\arraybackslash\hspace{0pt}}p{#1}}
\newcolumntype{R}[1]{>{\raggedright\arraybackslash\hspace{0pt}}p{#1}}

\usepackage[left=2cm, right=2cm, bottom=3cm, top=3cm]{geometry}
\usepackage[ruled,vlined, linesnumbered]{algorithm2e}
\SetKwRepeat{Do}{do}{while}
\SetKw{Return}{Return}
\SetKwComment{Comment}{\#{} }{}

\renewcommand{\vec}[1]{\ensuremath{\bm{\mathrm{#1}}}}
\newcommand{\mat}[1]{\ensuremath{\bm{\mathrm{#1}}}}
\newcommand{\transpose}[1]{\ensuremath{#1^\top}}

\newcommand{\review}[1]{{\color{black} #1}}
\newcommand{\secondreview}[1]{{\color{black} #1}}
\newcommand{\thirdreview}[1]{{\color{black} #1}}
\newcommand{\fourthreview}[1]{{\color{black} #1}}

\begin{document}
\begin{frontmatter}
\title{Online Multivariate Regularized Distributional Regression for High-dimensional Probabilistic Electricity Price Forecasting}
\author[sk,ude]{Simon Hirsch\textsuperscript{\orcidlink{0009-0008-1409-9677}}}
\ead{simon.hirsch@statkraft.com}
\ead{simon.hirsch@stud.uni-due.de}
\affiliation[sk]{
    organization={Statkraft Trading GmbH},
    country={Germany},
}
\affiliation[ude]{
    organization={University of Duisburg-Essen, House of Energy Markets and Finance},
    country={Germany},
}

\date{This Version: October 6, 2026}

\begin{abstract}
    \noindent \review{
        Probabilistic electricity price forecasting (PEPF) is vital for short-term electricity markets, yet the multivariate nature of day-ahead prices — spanning 24 consecutive hours — remains underexplored. At the same time, real-time decision-making requires methods that are both accurate and fast.
        We introduce an online algorithm for multivariate distributional regression models, allowing efficient modeling of the conditional means, variances, and dependence structures of electricity prices. The approach combines multivariate distributional regression with online coordinate descent and LASSO-type regularization (absolute shrinkage and selection operator), enabling scalable estimation in high-dimensional covariate spaces. Additionally, we propose a regularized estimation path over increasingly complex dependence structures, allowing for early stopping and avoiding overfitting.
        \secondreview{
            In a case study using historical data from the German day-ahead market, the proposed method yields interpretable and well-calibrated joint prediction intervals for the 24-dimensional price distribution and provides robust performance across a range of proper scoring rules. The results underscore the importance of modeling the dependence structure of electricity prices.
        }
        Furthermore, we analyze the trade-off between predictive accuracy and computational costs for batch and online estimation and provide a high-performing open-source Python implementation in the \texttt{ondil} package.
    }
\end{abstract}
\end{frontmatter}

\section{Introduction}

Short-term electricity markets play a key role in the integration of renewable energy sources and flexible generation in the electricity system. In Germany, the day-ahead auction is the major venue for physically delivered electricity. Trading volumes have grown with the increase of renewable generation capacity. To optimize decision-making and bidding strategies, market participants need accurate price forecasts. \review{Since} electricity prices are characterized by high volatility, positive and negative spikes and skewness, research and industry have moved toward probabilistic electricity price forecasting (PEPF) to account for their stochastic nature \cite[see e.g.][]{nowotarski2018recent, dexter2024probabilistic}. \review{With 24 hourly prices per day, electricity prices are multivariate time series with a potentially complex dependency structure.} However, the multivariate dimension has received little attention for PEPF so far, while being of high importance for market participants in the context of the optimization of flexible assets and portfolio management \citep{lohndorf2023value, pena2024hedging, beykirch2022bidding, beykirch2024value}. At the same time, the increasing availability of high-frequency data and the need for real-time decision-making in energy markets require online estimation methods for efficient model updating. This work presents an online, multivariate distributional regression model, which we apply to probabilistic day-ahead electricity price forecasting in Germany. Our work is among the first to treat the 24-dimensional hourly electricity prices as a multivariate distribution and the first to treat the problem in a strict online estimation setting, which makes the complex, high-dimensional distributional learning problem feasible on standard laptops. \thirdreview{Our results emphasize the importance of modeling the dependence structure for multivariate probabilistic forecasting of electricity prices.}

\begin{wrapfigure}{r}{0.5\textwidth}
    \centering
    \includegraphics[width=\linewidth]{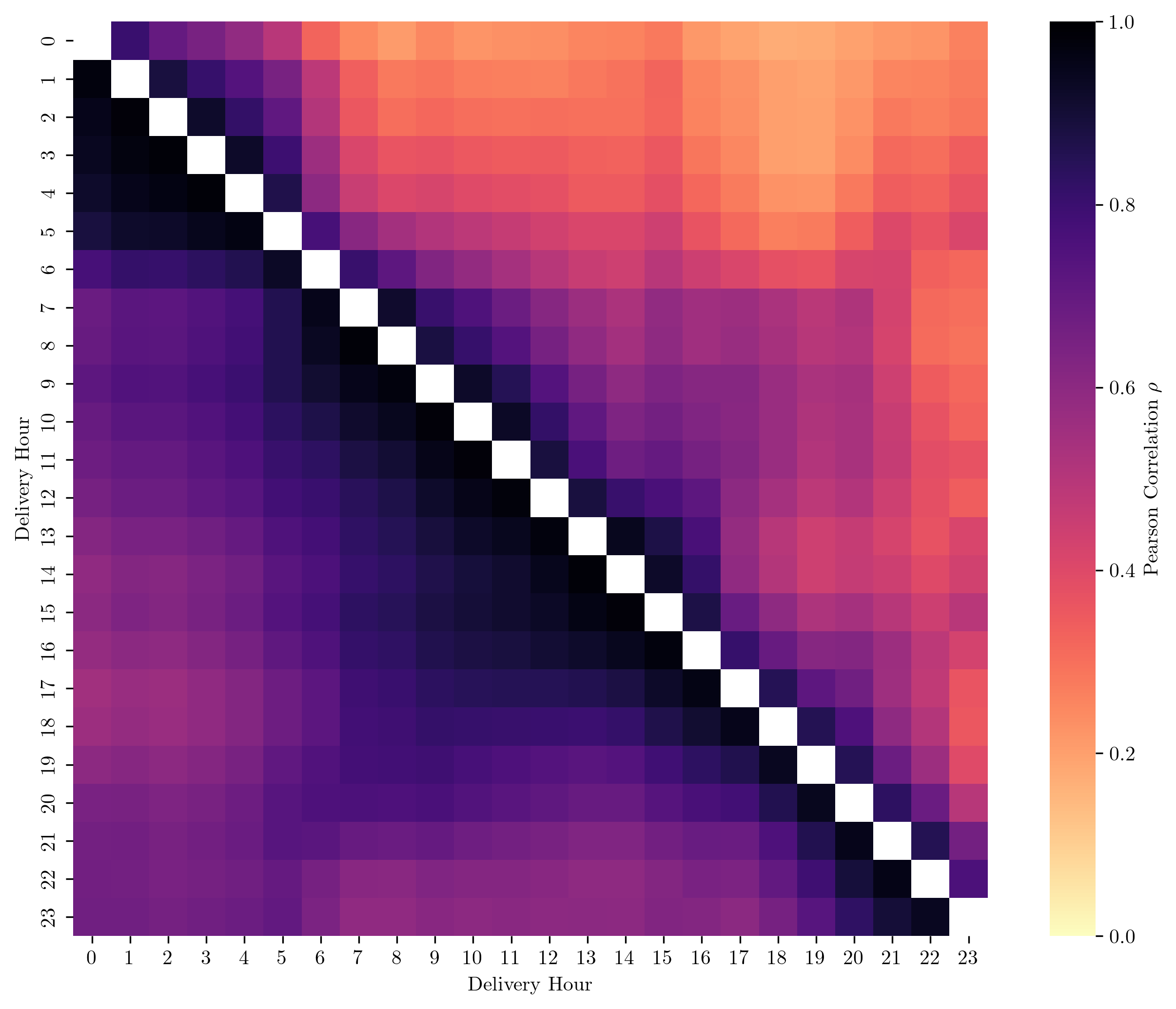}
    \caption{Correlation Matrix for day-ahead electricity prices $P_{d, h}$ in Germany. The lower triangle gives the hourly Pearson correlation $\rho$ for electricity prices. The upper triangle gives the hourly correlation of residuals $\varepsilon_{d, h} = P_{d, h} - \hat{\mu}_{d,h}$ for a standard LASSO-estimated autoregressive model \cite[LEAR, see e.g.][and Eq. \ref{eq:mean_model}]{lago2021forecasting}. The high degree of residual correlation, especially around the noon hours is visible. All correlation coefficients are statistically significant to the $\alpha = 0.01$ confidence level.}
    \label{fig:electricity_prices_spot_correlation}
\end{wrapfigure}

The literature on PEPF has evaluated a wide range of different statistical and machine learning methods, such as quantile regression, \secondreview{generalized autoregressive conditional heteroskedasticity models \citep[GARCH,][]{nowotarski2018recent, bille2023forecasting, marcjasz2023distributional},} conformal prediction methods, \citep[\secondreview{CP}, see e.g.][]{kath2021conformal, zaffran2022adaptive, lipiecki2024postprocessing, brusaferri2024line}, distributional regression and neural network approaches \citep[e.g.][]{ziel2021gamlss, marcjasz2023distributional, brusaferri2024nbmlss, hirsch2024online}. However, these works treat each delivery hour as \emph{independent}, univariate time series as in \cite{ziel2018day}.
\thirdreview{Electricity prices are known for annual, weekly and daily seasonal effects, high volatility and dependence \citep{ziel2018day}.} Figure \ref{fig:electricity_prices_spot_correlation} shows the correlation matrix of the raw electricity prices \review{(lower triangular) and} the residual correlation for a LASSO-estimated AutoRegressive model for the electricity price \citep[\secondreview{LEAR, upper triangular, }][]{lago2021forecasting}. We see a strong, statistically significant remaining residual cross-correlation, indicating that the resulting marginal error distributions, which are conditional on the mean, are not independent. On top of the statistical motivation, \cite{beykirch2022bidding, beykirch2024value} describe the need for predicting joint distributions for the optimization of schedules and bidding curves in energy markets, further examples are provided by \cite{pena2024hedging, lohndorf2023value}. 

Work on \emph{multivariate} probabilistic forecasting for day-ahead electricity prices is sparse in the literature and the majority of the existing work, e.g. \cite{maciejowska2024multiple, berrisch2024multivariate, han2023probabilistic, mashlakov2021assessing} and \cite{agakishiev2025multivariate}, does not evaluate multivariate scoring rules such as the \secondreview{Variogram score (VS), Dawid-Sebastiani-Score (DSS) or Energy Score (ES), but focus on the evaluation of the marginals of the multivariate distribution through the continuous ranked probability score (CRPS).} This reduces the problem to modeling 24 marginal distributions, taking only lagged cross-information into account. To the best knowledge of the author, only two studies truly model and evaluate the  multivariate dependence structure. \cite{janke2020probabilistic} approach the issue through implicit generative Copula models. \cite{grothe2023point} employ the Schaake shuffle, a post-processing method for point forecasts. On the contrary, in the fields of probabilistic weather, renewable production \citep{bjerregaard2021introduction, soerensen2022recent, kolkmann2024modeling} and probabilistic load forecasting \citep{gioia2022additive, browell2022covariance} truly multivariate forecasting approaches have gained more attention.

The goal of distributional regression or ``regression beyond the mean" \citep{kneib2023rage,klein2024distributional} is modeling not only the conditional expectation, but all distribution parameters of the assumed parametric response distribution conditional on explanatory variables. The most prominent model in this regard is the original Generalized Additive Model for Location, Scale and Shape \citep[GAMLSS, ][]{rigby2005generalized}, of which numerous extensions have been developed over the last years \citep{kock2023truly,kneib2023rage,muschinski2022cholesky} and distributional deep neural networks \citep[DDNN, e.g.][]{klein2021marginally, klein2023deep, rugamer2024semi}.  \secondreview{Due to the direct modeling of the variable's distribution}, this method is well suited for the generation of probabilistic forecasts and has been successfully applied in energy markets \citep{muniain2020probabilistic,gioia2022additive, serinaldi2011distributional,brusaferri2024nbmlss,marcjasz2023distributional}. \review{A drawback of fully distributional models is the computational effort. The need for efficient estimation approaches for (multivariate) distributional regression models has been recognized by \cite{umlauf2025scalable} and \cite{gioia2025scalable}, who propose efficient batch estimation approaches.}

For environments with large amounts of continuously incoming data, such as energy markets, online learning describes the task of updating the model given new data, without falling back on previous samples. Formally, in the strict online setting, after having seen $N$ samples of our data set, we fit a model, predict for step $N + 1$. Subsequently, we receive the realized values for $N+1$ and update our model, taking into account only the new row $N+1$. This approach allows an efficient processing of high-velocity data and results in greatly decreased computational effort. The principle is outlined in Figure \ref{fig:online_forecasting_study}. \thirdreview{
    The computational benefits of online learning are also of advantage in settings with low volumes of data due to the time saved in experimentation and model evaluation. Commonly, one needs to fit a larger number of bad models before one finds a good one. An analyst with limited time budget can therefore iterate a lot quicker if she/he employs online methods instead of repeated batch fitting.
} Online learning for \secondreview{the absolute shrinkage and selection operator (LASSO) regularized regression} for the mean has been introduced in \cite{angelosante2009online, angelosante2010online} and \cite{messner2019online}. Univariate approaches suitable for probabilistic forecasting based stochastic gradient descent have been developed for specific distributions, \cite[see e.g.][]{pierrot2021adaptive}, conformal prediction \citep[see e.g.][]{zaffran2022adaptive,gibbs2021adaptive,gibbs2024conformal} and the generic online distributional regression in \cite{hirsch2024online}. However, in the multivariate case, the literature remains sparse and focused on unconditional distributions and Copulae \citep[see e.g.][]{dasgupta2007line, zhao2022online, landgrebe2020online}.

We add to the literature by presenting a generic, online, regularized, multivariate distributional regression model, allowing to model all distribution parameters conditional on explanatory variables and validate the approach in a forecasting study for the day-ahead electricity market in Germany. Our paper is one the first to tackle the issue of truly multivariate probabilistic electricity price forecasting. \review{The contribution of this paper is therefore threefold:
\begin{itemize}
    \item \emph{Methodological Contribution:} We develop a regularized online estimation method for multivariate distributional regression based on the univariate work by \cite{hirsch2024online}. Our algorithm allows for two layers of regularization: \begin{itemize}
        \item By leveraging online coordinate descent and LASSO-type penalties for each individual distribution parameter, we allow for high-dimensional covariate spaces. 
        \item By exploiting the structure in scale matrix of the multivariate distribution, we develop a path-based estimation along increasingly complex dependency structures, allowing for parsimonious estimation and early stopping. We validate the trade-off between model complexity and predictive accuracy in our application study.
    \end{itemize} 
    Our algorithm is generic and can be applied to any parametric multivariate distribution, as long as the (log-)likelihood function and its derivatives are available. We implement the multivariate normal and $t$-distributions with three different parameterizations of the scale matrix. Further distributions can be easily added.
    \item \emph{Applied Contribution:} We apply the method to probabilistic forecasting of the joint distribution of \thirdreview{German day-ahead} electricity prices. \thirdreview{Our study is among the first to evaluate the full multivariate distribution through strictly proper scoring rules and the calibration of the 24-dimensional joint prediction bands (JPB) for a wide range of benchmark models and including the volatile years 2021 to 2023 in our test set.}
    \begin{itemize}
        \item \secondreview{We show that the multivariate distributional regression models, which allow modeling all distributional parameters, i.e. the mean, but also the dependence structure,  conditional on explanatory variables such as renewable in-feed or past prices provide well-calibrated prediction bands and \thirdreview{competitive} forecasting performance in terms of the Log Score (LS), Variogram Score (VS) and Dawid-Sebastiani Score (DSS), \thirdreview{while performance with respect to other metrics remains mixed.}}
        \item We showcase the computational advantage of online estimation by benchmarking with repeated batch estimation on various re-estimation frequencies. These results provide an ``efficient frontier" of computational costs against predictive accuracy.
    \end{itemize}
    \item \emph{Software:} We provide a high-performing Python implementation using just-in-time compilation and providing a familiar, \texttt{scikit-learn}-like API to facilitate the usage of our package for other researchers. We contributed our code to the \texttt{ondil} package by \cite{hirsch2024online}.
\end{itemize}
Thereby, our contribution is valuable both from a methodological and applied perspective in research and industry alike. Reproduction code can be found on \texttt{GitHub}.\footnote{See: \url{https://github.com/simon-hirsch/online-mv-distreg}.} The remainder of the paper is structured as usual: The following Section \ref{sec:online_multivariate_distreg} introduces the multivariate, online, regularized distributional regression model. Section \ref{sec:forecasting_study} introduces the forecasting study, the used data and discusses multivariate forecast evaluation in detail. Section~\ref{sec:results} presents our results and analyzes the computational costs. {Section \ref{sec:conclusion}} concludes the paper.}

\section{Online Multivariate Distribution Regression}\label{sec:online_multivariate_distreg}

\review{The following subsections introduce the building blocks for the multivariate, online distributional regression algorithm. The general structure of the algorithm and also the flow of the following section is outlined in Figure~\ref{fig:paper_overview}. First, we like to introduce some intuition to distributional regression. Distributional regression or ``regression beyond the mean" \citep{kneib2023rage} aims at modeling not only the conditional expectation, but all distribution parameters of the assumed parametric response distribution conditioned on explanatory variables. Generally, we aim to model 
$$
\mat{y} \sim \mathcal{F}(\vec{\theta}_1, ..., \vec{\theta}_K) \quad \text{where} \quad g_k(\vec{\theta}_k) = \mat{X}_k\vec{\beta}_k
$$
that is, we model the parameters $\vec{\theta}_k$ of the distribution $\mathcal{F}$ of the response variable $\mat{y}$ as regression based on the covariates in $\mat{X}$. \review{The link function $g(\cdot)$ ensures that the distribution parameters are in their domain.} Naturally, this distributional model is aligned with our goal of probabilistic forecasting. 

\begin{figure}[htb]
    \centering
    \includegraphics[width=\linewidth]{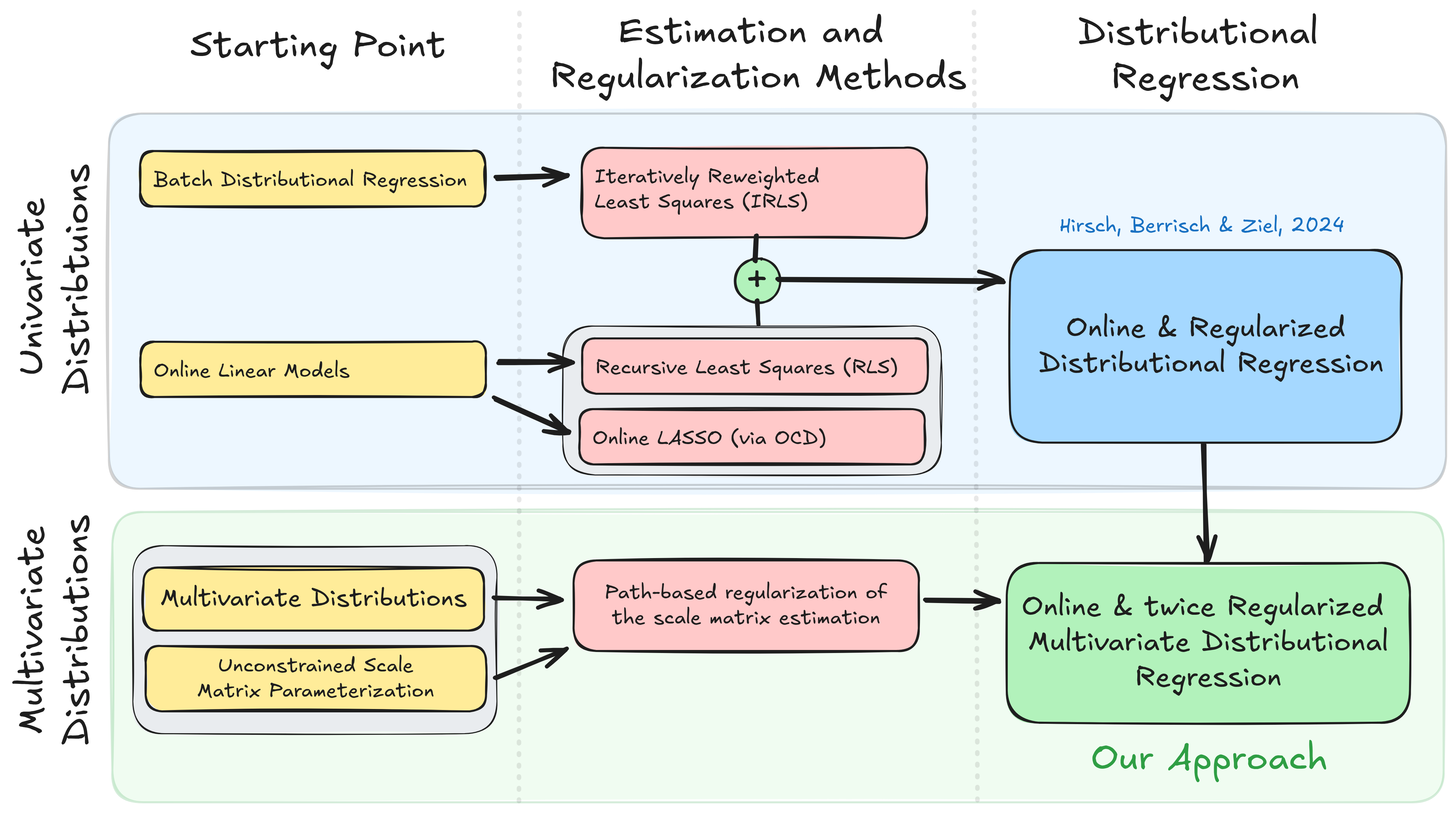}
    \caption{Online Distributional Regression. For univariate distributions (top panel), online distributional regression has been introduced by \citep{hirsch2024online} by a combination of the iteratively reweighted least squares algorithm with online coordinate descent. We extend the approach to multivariate distributions by utilizing unconstrained scale matrix parameterizations and introducing a path-based regularization along increasingly complex dependency structures.}
    \label{fig:paper_overview}
\end{figure}

\secondreview{Section~\ref{sec:distributional_regression_setting} gives a formal introduction to the univariate and multivariate case. Section \ref{sec:online_estimation} introduces our main contribution, the multivariate extension of online distributional regression. We further develop a path-based estimation along a sequence of the increasingly complex scale matrices, which allows for a regularized estimation and early stopping in Section~\ref{sec:path_based_estimation}.} }

We denote scalar float and integer values as lowercase letters (e.g. $a$), constants as large letters (e.g. $T$) vectors as bold, upright lower case letters (e.g. $\vec{v}$) and matrices as bold upper case letters (e.g. $\mat{A}$). The calligraphic $\mathcal{F}$ and $\mathcal{D}$ are reserved for (arbitrary) distributions, $\mathcal{N}$ denotes the normal distribution and $\mathcal{L}$ denotes the likelihood; other calligraphic letters (usually) denote index sets. Subscript values are usually indices in matrices, which we start with 0. Superscript indices (in square brackets) denote iterations and/or the number of samples received in the online setting.

\review{\subsection{Distributional Regression Setting} \label{sec:distributional_regression_setting}}

\thirdreview{In multivariate distributional regression}, we are interested in learning the conditional distribution parameters of the $D$-dimensional response {variable $\mat{Y} = (\vec{y}_{1}, ..., \vec{y}_{D})$}, conditional on the covariate \secondreview{matrix} $\mat{X}$, by adopting a multivariate parametric distribution $\mat{y}_i \sim \mathcal{F}_D(\mathbf{\Theta}_i)$, where $\mathbf{\Theta}_i = (\vec{\theta}_{i1}, ..., \vec{\theta}_{iK})$ is a tuple of $K$ scalar, vector or matrix-valued distribution parameters. Each of the coordinates $m$ of the distribution parameter $\vec{\theta}_k$ can again be related to its linear predictors by \begin{equation}
    g_{km}(\vec{\theta}_{km}) = \vec{\eta}_{km} = \mat{X}_{km}\vec{\beta}_{km}. 
\end{equation}
\thirdreview{Each of the distribution parameters are linked to the covariate data through a known, twice differentiable link function $g_k(\cdot)$.
Hence, we formally have: \begin{equation}
    \mat{y}_i \sim \mathcal{F}(\mathbf{\Theta}_i) \quad \text{and} \quad \theta_{kim} = g_{km}^{-1}\left(\transpose{{\vec{x}_{kim}}}\vec{\beta}_{km}\right)
\end{equation}and the probability density function $f(\vec{y}_i \mid \mathbf{\Theta}_i)$. 
The distributional regression framework therefore allows the modeling of all distribution parameters as linear regression equations of the design matrices $\mat{X}_k$, which can be a subset or all of the available the covariate data $\mat{X}$.}
\review{We note two important points here: 
\begin{itemize}
    \item \thirdreview{The distribution parameters are subject to constraints. This especially holds for the scale (respectively precision) matrix, which is often required to be positive semi-definite. To ensure this holds, unconstrained parameterizations such as the (modified) Cholesky-decomposition (\secondreview{CD, $\mat{\Sigma} = \mat{A}\transpose{\mat{A}}$ resp. MCD}, $\mat{\Sigma} = \transpose{(\mat{L}^{-1})}\mat{D}\mat{L}^{-1}$) or the low-rank approximation are often used \citep[\secondreview{LRA},][$\mat{\Omega} = \mat{A} + \mat{V}\transpose{\mat{V}}$]{pourahmadi2011covariance, muschinski2022cholesky, salinas2019high}.}
    \item In practice, the different distribution parameters $\vec{\theta}_{1}, ..., \vec{\theta}_{K}$ can have many different shapes. Take, e.g. the multivariate $t$-distribution, parameterized using the Cholesky factor of the precision matrix $\mat{\Sigma} = \transpose{\mat{L}}\mat{L}$, denoted as $t_D(\vec{\theta}_1, \vec{\theta}_2, \vec{\theta}_3) \Leftrightarrow t_D(\vec{\mu}, \mat{L}, \vec{\nu})$. Then $\vec{\mu}$ is a ${N \times D}$ matrix, $\mat{L}$ is a ${N \times D \times D}$ cube (of which each vertical slice is a triangular matrix) and $\vec{\nu}$ is a $N \times 1$ vector. Accordingly, the index set $\mathcal{M}_k$ of coordinates spans $\mathcal{M}_1 = \{1, .., D\}$, $\mathcal{M}_2 = \{(1, 1), ..., (D, D)\}$ and $\mathcal{M}_3 = \{1\}$ and its cardinality is given by the product of the parameter's dimension beyond $N$. 
\end{itemize}} The general setting introduced here includes the Gaussian multivariate distributional regression introduced by \cite{muschinski2022cholesky}, the Copula-based multivariate distributional regression by \cite{kock2023truly} and the MCD-based additive covariance models by \cite{gioia2022additive}. 
\thirdreview{In the univariate case, \cite{rigby2005generalized} introduce iteratively reweighted least squares (IRLS), maximizing the penalized likelihood, to estimate $\vec{\beta}_k$. It is important to note here that in the frequentist \secondreview{setting}, the IRLS algorithm is agnostic to the actual estimation technique \citep[see e.g. p. 113 in][]{stasinopoulos2024generalized}. Different flavors of LASSO-type regularized estimation approaches have been introduced by \cite{groll2019lasso, muniain2020probabilistic, ziel2021gamlss, o2023variable}. A regularized, incremental estimation approach using online coordinate descent has been proposed by \cite{hirsch2024online}, which will form the basis for the multivariate approach proposed in this paper.} 
The general estimation principle of repeatedly iterating through the distribution parameters until convergence translates to the multivariate case. 

\subsection{Online Estimation Algorithm}\label{sec:online_estimation}

This section introduces the online estimation algorithm for multivariate distributional regression. The algorithm is based on the combination of the iteratively reweighted least squares (IRLS) algorithm for the estimation of distributional regression models and online coordinate descent (OCD) for regularized estimation. \citet[\secondreview{RS}]{rigby2005generalized} introduce iteratively reweighted least squares for \review{the estimation of GAMLSS models.} The RS algorithm consists of two nested loops, in which we cycle repeatedly through the distribution parameters and run a weighted fit of the \review{working vector $\vec{z}$} on the design matrix $\mat{X}$ using the diagonal weight matrix $\mat{W}$. The following paragraphs introduce the scoring vector and weights, the algorithm and the necessary modifications to move from a univariate case to the multivariate case. The score vector is defined as 
\begin{equation}\label{eq:rs_score_vector}
    \vec{u} = \frac{\partial \ell}{\partial \eta},
\end{equation}
where $\ell$ is the log-likelihood $\ell = \log(\mathcal{L})$ and $\eta = g(\vec{\theta})$ is the linked predictor. The working vector for the Newton-Raphson or Fisher-Scoring algorithm is defined as
\begin{equation}\label{eq:rs_working_vector}
    \vec{z} = g\big(\hat{\vec{\theta}}\big) + \frac{\partial \ell}{\partial \eta} \mat{W}^{-1} \Leftrightarrow \vec{z} = \vec{\eta} + \frac{\partial \ell}{\partial \eta} \mat{W}^{-1},
\end{equation}
where the weights are defined as:
\begin{align}\label{eq:rs_weights}
    \mat{W} = - \frac{\partial^2 \ell}{\partial \eta^2} & & \text{or} & & \mat{W} = - \mathbb{E}\left[\frac{\partial^2 \ell}{\partial \eta^2}  \right]
\end{align} 
for Newton-Raphson and Fisher's scoring respectively. In the IRLS, estimation is done by repeatedly fitting the working vector $\vec{z}$ on the design matrix $\mat{X}$ using the weights $\mat{W}$ until convergence. \secondreview{
    \cite{hirsch2024online} use online LASSO estimation for linear models as building block to allow for online estimation of distributional regression models. The online LASSO is based on the Gramian matrices $$
        \mat{G} = \transpose{\mat{X}}\mat{W}\mat{\Gamma}\mat{X} \quad \text{and} \quad \vec{h} = \transpose{\mat{X}}\mat{W}\mat{\Gamma}\vec{z},
    $$ where $\mat{W}$ and $\mat{\Gamma}$ are estimation or sample and discounting weight matrices and $\mat{G}$ and $\vec{h}$ can be updated for the $n+1$-th observation in an online fashion as follows:
    $$       \mat{G}^{[n+1]} = \gamma\mat{G}^{[n]} + \transpose{\vec{x}_{n+1}}\vec{x}_{n+1} \quad \text{and} \quad \vec{h}^{[n+1]} = \gamma\vec{h}^{[n]} + \transpose{\vec{x}_{n+1}}z_{n+1},
    $$ where $\gamma \in (0, 1]$ is a forgetting factor, which allows to put more weight on recent data. The online LASSO estimation can then be performed by running coordinate descent on the updated $\mat{G}$ and $\vec{h}$. We denote this as $$
        \mat{\beta} = \textsc{OnlineLASSO}(\mat{G}, \vec{h}, \vec{\lambda}).
    $$ and the details can be found in \thirdreview{the online supplement} and \cite{angelosante2009online, messner2019online, hirsch2024online}.
}
\secondreview{
    In the multivariate case, we extend the two nested loops of the IRLS algorithm with a third loop along all elements of the distribution parameter $k$. In the outer loop, we iterate through all distribution parameters. In the inner loop, we repeatedly run a weighted fit of the working vector $\vec{z}$ on the design matrix $\mat{X}$ using the weights $\mat{W}$ until convergence. Algorithm \ref{alg:online_mv_gamlss} gives an overview on the online estimation of multivariate distributional regression models. Note that in the inner loop, we run the weighted fit sequentially for all elements of the distribution parameter (see Lines 4 and 6). As discussed in Section \ref{sec:distributional_regression_setting}, we define the index sets $\mathcal{K} = \{1, ..., p\}$ for the number of parameters and $\mathcal{M}_k = \{1, .., M_k\}$ for the number of elements of each parameter. 
    \begin{algorithm}[htb]
    \raggedright
    \caption{Online regularized multivariate distributional regression.}
    \label{alg:online_mv_gamlss}
    \DontPrintSemicolon
    \KwIn{$\vec{y}^{[n+1]}, \mat{X}_{k,m}^{[n+1]}$ and the stored Gramian matrices $\mat{G}_{km}^{[n]}, \vec{h}_{km}^{[n]}$.} 
    Initialize the fitted values $\widehat{\vec{\theta}}_{km}^{[n+1]} = \hat{\vec{\beta}}_{km}^{[n]}\transpose{(\mat{X}_{km}^{[n+1]})}$ for $k,m  \in \mathcal{K} \times \mathcal{M}$.  \;
    Evaluate the linear predictors ${\widehat{\mat{\eta}}^{[n+1]}_{km} = g_{km}(\hat{\theta}_{km}^{[n+1]})}$ for $k,m  \in \mathcal{K} \times \mathcal{M}$. \;
    \For{\normalfont outer iteration $i = 0, ...$ \normalfont\textbf{until} convergence}{
        \ForAll{\normalfont distribution parameter $k \in \mathcal{K}$}{         
            \For{\normalfont inner iteration $r = 0, 1, ...$ \normalfont\textbf{until} convergence}{
                \ForAll{\normalfont distribution element $m \in \mathcal{M}_k$}{         
                    Evaluate $u_{km}^{[n+1]}$, $w_{km}^{[n+1]}$ and $z_{km}^{[n+1]}$ using Equations \eqref{eq:rs_score_vector}, \eqref{eq:rs_working_vector} and \eqref{eq:rs_weights}. \;
                    Update $\mat{G}_{km}^{[n+1]}$ and $\vec{h}_{km}^{[n+1]}$ based on $u_{km}^{[n+1]}$, $w_{km}^{[n+1]}$ and $z_{km}^{[n+1]}$. \;
                    Update $\hat{\vec{\beta}}_{km}^{[n+1, \vec{\lambda}]} \gets \textsc{OnlineLASSO}(\mat{G}_{km}^{[n+1]}, \vec{h}_{km}^{[n+1]}, \vec{\lambda})$. \;
                    Model selection using IC for $\hat{\vec{\beta}}_{km}^{[n+1]}$ and calculate the updated $\hat{\vec{\eta}}_{km}^{[n+1]}$ \;
                }
            }
        }   
    }
    \KwOut{$\widehat{\vec{\beta}}_{k, n+1}$ and $\widehat{\vec{\Theta}}^{[n+1]} = (\widehat{\vec{\theta}}_0^{[n+1]}, ..., \widehat{\vec{\theta}}_p^{[n+1]})$ and the updated $\mat{G}_{km}^{[n+1]}$ and $\vec{h}_{km}^{[n+1]}$.}
    \end{algorithm}
}%
\secondreview{In the following paragraphs, we will discuss the details of the algorithm and the necessary modifications to move from the univariate case to the multivariate case.}

In the original GAMLSS, \cite{rigby2005generalized} use Fisher's scoring \secondreview{in Equation \ref{eq:rs_weights}}. Our approach generally uses Newton-Raphson scoring for the multivariate case, since the derivation of the expected value of second derivatives can be intractable, especially for more complex parameterizations of the precision matrix. Newton-Raphson scoring requires the partial derivatives of the log-likelihood function \emph{with respect to the predictors}. While many previous works on distributional regression employ Newton-Raphson scoring, each derive the partial derivatives for specific combinations of distribution function and link function only \cite[see e.g.][]{o2023variable, muschinski2022cholesky}. \secondreview{To facilitate the computational implementation in a mix-and-match fashion, we propose to use the first and second derivative of the log-likelihood \emph{with respect to the parameter} and the first and second derivative of the link function and relate both to the necessary derivatives for Newton-Raphson scoring using the equalities
\begin{align}
    \frac{\partial \ell}{\partial \eta} &= \frac{\partial \ell}{\partial \theta} \left( \frac{\partial g(\theta)}{\partial\theta}\right)^{-1} \quad\label{eq:newton_raphson_generic_first}  \quad \text{and}\\ 
    \frac{\partial^2 \ell}{\partial \eta^2} &= \left(
        \frac{\partial^2 \ell}{\partial \theta^2} \frac{\partial g(\theta)}{\partial\theta} - \frac{\partial \ell}{\partial \theta} \frac{\partial^2 g(\theta)}{\partial\theta^2}
    \right) \left( \frac{\partial g(\theta)}{\partial\theta}\right)^{-3}. \label{eq:newton_raphson_generic_second}
\end{align}
The proof is straight-forward and utilizes the chain and quotient rules and can be found in the online supplement}.  We provide the necessary first and second partial derivatives of the log-likelihood with respect to the distribution parameter's coordinates, ${\partial \ell}/{\partial \theta}$ and ${\partial^2 \ell}/{\partial \theta^2}$, for all parameters for the multivariate normal and multivariate $t$-distribution given in Table \ref{tab:implemented_dists}. The derivations can be found in \secondreview{the online supplement}.

\begin{table}[htb]
    \centering
    \resizebox{\textwidth}{!}{%
    \begin{tabular}{lllllll}
        \toprule
        Distribution            &   \multicolumn{2}{c}{Location}   &    \multicolumn{2}{c}{Scale $\boldsymbol{\Sigma}$ resp. Precision $\boldsymbol{\Omega}$} & \multicolumn{2}{c}{Shape} \\ 
        \cmidrule(lr){2-3}%
        \cmidrule(lr){4-5}%
        \cmidrule(lr){6-7}%
                                &   Param.          & Dim.         &   Param.   & Dim.                                                      & Param. & Dim. \\
        \midrule
        Multivariate Gaussian   &   $\vec{\mu}$     & $N \times D$  &   $\mat{\Omega} = \transpose{(\mat{A}^{-1})}(\mat{A}^{-1})$       & $N \times  \text{triangular}(D \times D)$              & - & - \\
        Multivariate Gaussian   &   $\vec{\mu}$     & $N \times D$  &   $\mat{\Omega} = \transpose{\mat{L}}(\mat{D}^{-1})\mat{L}$ & $N \times  \text{triangular}(D \times D)$, $N \times \text{diag}(D)$ & - & -\\
        Multivariate Gaussian   &   $\vec{\mu}$     & $N \times D$  &   $\mat{\Omega} = \mat{A} + \mat{V}\transpose{\mat{V}}$           & $N \times \text{diag}(D)$, $D \times r $ & - & -\\
        Multivariate-$t$        &   $\vec{\mu}$     & $N \times D$  &   $\mat{\Omega} = \transpose{(\mat{A}^{-1})}(\mat{A}^{-1})$       & $N \times  \text{triangular}(D \times D)$    & $\nu$ & $N \times 1$\\
        Multivariate-$t$        &   $\vec{\mu}$     & $N \times D$  &   $\mat{\Omega} = \transpose{\mat{L}}(\mat{D}^{-1})\mat{L}$ & $N \times  \text{triangular}(D \times D)$, $N \times \text{diag}(D)$ & $\nu$ & $N\times 1$ \\
        Multivariate-$t$        &   $\vec{\mu}$     & $N \times D$  &   $\mat{\Omega} = \mat{A} + \mat{V}\transpose{\mat{V}}$           & $N \times \text{diag}(D)$, $D \times r $ & $\nu$ & $N\times 1$ \\  
        \bottomrule
    \end{tabular}}
    \caption{Overview of multivariate distributions and scale matrix parameterization (Param.) implemented in the paper and the respective dimensions (Dim.) for input data $\mat{Y}$ of shape $N \times D$. Note that the number of parameters for the CD-based parameterization grows quadratically in $D$, but the LRA-based parameterizations grow linear in $D$ for fixed $r$. }
    \label{tab:implemented_dists}
\end{table}

\secondreview{
    For each inner iteration $i$, the update of the Gramian matrices starts at the Gramian matrices of $\mat{G}_{km}^{[n]}$ and $\vec{h}_{km}^{[n]}$ and the new information enters through the update \review{of the weights $\mat{W}$ and the working vector $\vec{z}$}. However, the weights are also updated iteratively along each inner and outer iteration $i$ and $r$ due to the Newton-Raphson step toward the optimal coefficients. The weights can only be updated for the current update step $n+1$, while previous weights remain fixed. In a pure batch case, all weights are updated within each Newton-Raphson step. This introduces an approximation error for the online case, which can be controlled by the forget parameter $\gamma$ as shown in \cite{hirsch2024online}.
}

\secondreview{For each element of the distribution parameter, we estimate a regularization path $\vec{\beta}_{km}$ for a decreasing $\vec{\lambda}_{km}$.} This raises the issue of model selection, i.e. the selection of the optimal regularization $\lambda_{mk}^\text{opt}$. We propose to use information criteria (IC), as it is well-aligned to the likelihood-based framework of distributional regression. Define a generalized IC \begin{equation}
    \operatorname{IC} = 
        -2\ell\left(\mat{Y} \mid \widehat{\mat{\Theta}}\right) 
        + \nu_0 K 
        + \nu_1 K \log(N) 
        + \nu_2 K \log\left(\log(N)\right) 
\end{equation}
where $\ell$ is the log-likelihood under the model, $K$ is the number of parameters in the model and $N$ the number of seen observations. We can recover Akaikes Information Criterion (AIC), the Bayesian Information Criterion (BIC), and the Hannan-Quinn Information Criterion (HQC) by setting $\nu_0, \nu_1, \nu_2$ accordingly. The optimal regularization parameter is then selected as $\lambda_{mk}^\text{opt} = \operatorname{argmin}_{\lambda} \operatorname{IC}$. Since the evaluation of the likelihood can be costly for high-dimensional data, we propose to employ the first derivative of the log-likelihood, i.e. calculate 
\begin{equation}
    \ell\left(\mat{Y} \mid \widehat{\mat{\Theta}}^{[\lambda_i]} \right) \approx 
        \ell\left(\mat{Y} \mid \widehat{\mat{\Theta}}^{[\lambda_0]}\right) 
            + \frac{\partial \ell}{\partial \theta} \left( \widehat{\theta}_{km}^{[\lambda_0]} - \widehat{\theta}_{km}^{[\lambda_i]}\right) 
\end{equation}
where the superscript ${[\lambda_i]}$ denotes the model with the regularization parameter $\lambda_i$. The approximation is valid for small changes in the regularization parameter and avoids the costly re-evaluation of the likelihood.

The algorithm runs iteratively along all coordinates of the distribution parameter \secondreview{(line 6). Since the partial derivatives are not information orthogonal, }the coordinates of the distribution parameters might impact each other, e.g. in the matrix multiplication of the CD-based scale matrix \secondreview{(see also the definition of the derivatives in Table \ref{tab:implemented_dists} and the online supplement)}. To stabilize the estimation, we propose to update the values in the very first iteration $i$ by a ``dampened" version, i.e. taking \begin{equation}\label{eq:dampening}
    \hat{\eta}_m^{[0, i]} \leftarrow g_m^{-1}\left((i+1)\hat{\theta}_m^{[0, i]} + \hat{\theta}_m^{[0, i-1]}) / (i+1)\right)
\end{equation} Hence, the predictions from the first iteration will be the average of the first fitted values and the initialization. This feature is mainly important for the scale matrix, whose coordinates are usually not orthogonal and less so for the location and (scalar) tail parameters. \secondreview{For the same reason, the options for parallelization remain limited unfortunately.} For the multivariate normal and $t$-distribution used in this paper, only the estimation of the location parameter can be parallelized, as well as the estimation of the coordinates of the LRA matrix {$\mat{A} = \operatorname{diag}(a_1, ..., a_D)$} for the normal distribution. \secondreview{This would allow to parallelize lines 6-10.} For the $t$-distribution, the estimation of $\mat{A}$ can only be parallelized for sufficiently high degrees of freedom. 
\secondreview{%
    A potential strategy for parallelization with non-orthogonal parameters could be using a step size smaller than one, i.e. using a convex combination of the previous and the newly estimated coefficients (see Equation \ref{eq:dampening}), which, in combination with a Cole-Green-algorithm \citep[CG, see][]{green1984iteratively, cole1992smoothing}, could allow to parallelize the estimation of all elements of all distribution parameters (lines 5-10). 
    However, this strategy introduces an additional hyperparameter and we incur further questions with respect to individual or joint regularization and model selection, as we have to select the optimal regularization parameter for each coordinate of each distribution parameter.
    Hence, we leave the question of parallelization for future research.
} 

\subsection{Path-based Regularized Estimation for the Scale Matrix}\label{sec:path_based_estimation}

\review{Using LASSO-type regularization, the algorithm can handle high-dimensional covariate spaces for each coordinate of the distribution parameter. However, for high-dimensional response variables, i.e. large $D$, the number of parameters in the scale matrix grows quadratically in $D$ \secondreview{for most unconstrained parameterizations}. To alleviate this issue and allow for parsimonious modeling for large~$D$, we propose a path-based estimation approach that starts with a simple (highly regularized) structure of the scale matrix and gradually increases its complexity. 

\begin{figure}[htb]
    \centering
    \resizebox{\linewidth}{!}{\begin{tikzpicture}[
    align=center,
    squarednodegreen/.style={rectangle, draw=Gray, fill=Gray!01, very thick, minimum size=5mm},
    squarednodeblue/.style={rectangle, draw=Gray, fill=Gray!01, very thick, minimum size=5mm},
]
\node[squarednodegreen]   (cholfac_00)    {
    1st iteration $\mat{L}^{[\textcolor{Red}{0}]}$ \\[1em] 
    $\begin{pmatrix}
        l_{1,1} &           &  &  & 0\\
        0        & l_{2,2}   &  & \\
        ...      &  0        & l_{3, 3} &  & \\
        & ...& ... & ... & \\
        0 &  & ... &0 & l_{d, d} 
    \end{pmatrix}$};
\node[squarednodegreen]      (cholfac_01)    [right=of cholfac_00]   {
    2nd iteration  $\mat{L}^{[\textcolor{Red}{1}]}$ \\[1em] 
    $\begin{pmatrix}
        l_{1,1} &          &  &  & 0\\
        \color{Red}{l_{2,1}}         & l_{2,2}   &  & \\
        & \color{Red}{l_{3,2}}          & l_{3, 3} &  & \\
        && \color{Red}{...} & ... & \\
        0&&& \color{Red}{l_{d,d-1}}& l_{d, d}
    \end{pmatrix}$};
\node[squarednodegreen]      (cholfac_02)    [right=of cholfac_01]    {
    3rd iteration  $\mat{L}^{[\textcolor{Red}{2}]}$ \\[1em] 
    $\begin{pmatrix}
        l_{1,1} &          &  &  & 0\\
        \color{Green}{l_{2,1}}         & l_{2,2}   &  & \\
        \color{Red}{l_{3,1}}         & \color{Green}{l_{3,2}}          & l_{3, 3} &  & \\
        & \color{Red}{...} & \color{Green}{...} & ... & \\
        0&& \color{Red}{l_{d, d-2}} & \color{Green}{l_{d,d-1}}& l_{d, d} 
    \end{pmatrix}$};
\node[squarednodegreen]      (cholfac_03)    [right=of cholfac_02]    {
    Last iteration $\mat{L}^{[\textcolor{Red}{d-1}]}$\\[1em] 
    $\begin{pmatrix}
        l_{1,1} &          &  &  & 0\\
        \color{Green}{l_{2,1}}   & l_{2,2}   &  & \\
        \color{Green}{l_{3,1}}   & \color{Green}{l_{3,2}}          & l_{3, 3} &  & \\
        \color{Green}{...}       & \color{Green}{...} & \color{Green}{...} & ... & \\
         \color{Red}{l_{d,d}}   & \color{Green}{...} & \color{Green}{l_{d, d-2}} & \color{Green}{l_{d,d-1}}& l_{d, d} 
    \end{pmatrix}$};
\node[squarednodeblue]    (lowrank0) [below=2cm of cholfac_00]    {
    1st iteration $\mat{V}^{[\textcolor{Red}{0}]}$ \\[1em] 
    $\begin{pmatrix}
        0 & ... & 0   \\
        0 & ... & 0   \\
        ... & ... & ... \\
        0 & ... & 0 \\
    \end{pmatrix}$};
\node[squarednodeblue]      (lowrank1)    [below=2cm of cholfac_01]   {
    2nd iteration $\mat{V}^{[\textcolor{Red}{1}]}$ \\[1em] 
    $\begin{pmatrix}
        \color{Red}{v_{1,1}} & 0    & ... & 0   \\
        \color{Red}{v_{2,1}} & 0    & ... & 0   \\
        \color{Red}{...}     & ...  & ... & ... \\
        \color{Red}{v_{d,1}} & 0    & ... & 0 \\
    \end{pmatrix}$};
\node[squarednodeblue]      (lowrank2)    [below=2cm of cholfac_02]   {
    3rd iteration $\mat{V}^{[\textcolor{Red}{2}]}$ \\[1em] 
    $\begin{pmatrix}
        \color{Green}{v_{1,1}}  & \color{Red}{v_{2,1}} & 0 &... & 0   \\
        \color{Green}{v_{2,1}}  & \color{Red}{v_{2,2}} & 0 &... & 0   \\
        \color{Green}{...}      & \color{green}{...} & .. &... & ... \\
        \color{Green}{v_{d,1}}  & \color{Red}{v_{d,2}} & 0 &... & 0 \\
    \end{pmatrix}$};
\node[squarednodeblue]      (lowrank3)    [below=2cm of cholfac_03]    {
    Last iteration $\mat{V}^{[\textcolor{Red}{d}]}$\\[1em] 
    $\begin{pmatrix}
        \color{Green}{v_{1,1}}  & \color{Green}{v_{2,1}} & ... & \color{Green}{v_{1,r-1}} & \color{Red}{v_{2,r}}   \\
        \color{Green}{v_{2,1}}  & \color{Green}{v_{2,2}}  & ... & \color{Green}{v_{2,r-1}} & \color{Red}{v_{2,r}}   \\
        \color{Green}{...}      & \color{Green}{...}  & ... & ... & \color{Red}{...} \\
        \color{Green}{v_{d,1}}  & \color{Green}{v_{d,2}}  & ... & \color{Green}{v_{d,r-1}}  & \color{Red}{v_{2,r}}\\
    \end{pmatrix}$};


\draw[very thick][->] (cholfac_00.east) -- (cholfac_01.west);
\draw[very thick][->] (cholfac_01.east) -- (cholfac_02.west);
\draw[very thick, dotted][->] (cholfac_02.east) -- (cholfac_03.west);
\node[draw, very thick, draw=gray!25, fit=(cholfac_00) (cholfac_01) (cholfac_02) (cholfac_03), inner sep=10pt, minimum width=3cm, minimum height=2cm, label=above:{\large Cholesky-based scale matrix parameterization $\mat{\Omega}^{[\alpha]} = \transpose{(L^{[\alpha]})} (L^{[\alpha]})$}] (box1) {};

\draw[very thick][->] (lowrank0.east) -- (lowrank1.west);
\draw[very thick][->] (lowrank1.east) -- (lowrank2.west);
\draw[very thick, dotted][->] (lowrank2.east) -- (lowrank3.west);
\node[draw, very thick, draw=gray!25, fit=(lowrank0) (lowrank1) (lowrank2) (lowrank3), inner sep=10pt, minimum width=3cm, minimum height=2cm, label=above:{\large Low-rank approximation-based scale matrix parameterization, keeping $\mat{A} = \operatorname{diag}(a_0, ..., a_d)$}] (box2) {};

\end{tikzpicture}} 
    \caption{
        Path-based estimation along increasingly complex scale matrix parameterizations. 
        \review{
            The top panel shows the AD-$r$ regression for a Cholesky-based parameterization $\transpose{((\mat{L}^{[\alpha]})^{-1})}(\mat{L}^{[\alpha]})^{-1}$. 
            The superscript $[\alpha]$ denotes the iteration regularization. 
            Black elements are the state after the initial fit assuming independence. 
            \secondreview{Red elements of the vectors are added in iteration $\alpha$.
            Every arrow denotes that we add one off-diagonal respectively column to the parameterization of the scale matrix.
            Green elements have been added in previous iterations ($\alpha-1$ and before) and are warm-started for the current iteration.}
            The lower panel shows the estimation along the LRA-based parameterization $\mat{A} + (\mat{V}^{[\alpha]})\transpose{(\mat{V}^{[\alpha]})}$,
        }%
        where $\mat{A} = \operatorname{diag}(a_1, ..., a_d)$ is not regularized and the $D \times r$ matrix $\mat{V}$ is filled column-wise with non-zero elements. 
    }
    \label{fig:path_based_estimation}
\end{figure}

\secondreview{This approach is inspired by path-based estimation methods in high-dimensional statistics \citep[such as the graphical LASSO,][]{friedman2008sparse}}.} We exploit that in many cases, some structure can be imposed on the scale matrix, i.e. in spatial or temporal data, which has a clear dependence pattern along the diagonal. In these cases, the scale matrix can be regularized by systematically setting off-diagonal elements to zero \citep{gabriel1962ante,zimmerman1997structured,zimmerman1998computational}. While both, the CD-based and the LRA-based scale matrix parameterization lend themselves to this type of regularization, the approach is mainly popular with the Cholesky-based parameterizations due to the relationship between the elements of the CD and the temporal correlation for longitudinal data under the name \secondreview{antedependence-$r$ (AD-$r$)} regularization. However, such regularization is commonly applied a-priori and not in a data-driven fashion, see e.g. \cite{muschinski2022cholesky,zimmerman1997structured}. On the other side, in coordinate descent estimation of regularized problems such as (graphical) LASSO, path-based estimation starting from a strongly regularized solution toward an (almost) not regularized solution has proven itself as an efficient solution approach. In this section, we aim to combine these two principles by introducing path-based estimation for the regularized scale matrix.

\secondreview{On a high level, our algorithm starts with an ``independence-parameterization" of the scale matrix and subsequently adds more non-zero elements and thus complexity to the parameterization of the scale matrix. Figure \ref{fig:path_based_estimation} illustrates how the path-based estimation uses increasingly complex specifications for the scale matrix $\mat{\Sigma}$ respectively $\mat{\Omega}$. Formally, for the regularization parameter $\alpha$, \begin{enumerate}
    \item Fit the distributional regression model using Algorithm \ref{alg:online_mv_gamlss} for the current regularization level $\alpha$, where we set the elements of the scale to zero for \begin{itemize}
        \item the \review{Cholesky}-based parameterizations if the indices \review{of the diagonal matrices~$\mat{A}$ or~$\mat{L}$,} $i,j$ are such that $|i-j| > \alpha$.
        \item the LRA-based parameterization if the indices $d,r$ of $\mat{V}$ are such that $r \geq \alpha$.
    \end{itemize}
    \item Evaluate the log-likelihood for the current regularization level $\alpha$.
    \item Stop early if the log-likelihood (or information criteria) does not increase sufficiently.
\end{enumerate} 
We can use warm-starting for all previously fitted elements of the scale matrix, however, due to the non-orthogonality of the elements, we need to re-estimate all elements of the scale matrix in each iteration. We discuss some details in the following:} \begin{itemize}
    \item For a (small) fixed maximum regularization size, the number of parameters in the CD \review{and MCD}-based parameterization grows (almost) linearly in $D$, alleviating the disadvantage of quadratic complexity.
    \item For the multivariate $t$-distribution, independence is only achieved as $\nu \rightarrow \infty$. We therefore set a high initial guess ($\nu = 10^6$) for the first outer iteration of $\mat{\mu}$ and $\mat{\Omega}$ to ensure numerical stability for the first iteration and subsequently choose a lower initial guess for the first iteration of $\nu$, since the Newton-Raphson algorithm relies on appropriate start values and tends to alternate between extrema otherwise \cite[see e.g.][on the impact of initial values for Newton-Raphson algorithms.]{casella2021choice,kornerup2006choosing}\footnote{We have found the algorithm to iterate between $\nu = 2$ and $\nu > 10^{10}$ for too large start values for the degrees of freedom. The proposed approach however has proved stable through the full simulation study with highly volatile electricity prices.}
    \item We can employ the path-based estimation to early stop the estimation, if the log-likelihood \review{or} an information criterion does not increase sufficiently by adding more non-zero elements. This allows for both, implicit regularization and decreased estimation time. However, \secondreview{once the algorithm stops early}, we cannot increase the complexity of the parameterization in the online estimation, but need to treat it as fixed.
\end{itemize}
Currently, the Algorithm will add only full off-diagonals \review{(for Cholesky-based approaches)} respectively columns (LRA). The implementation however could also work for block-wise schemes \cite[see e.g. the adaptive block structure in][]{cai2012adaptive} or user-defined regularization patterns. \secondreview{The development of smart selection schemes for choosing the next elements to include in the scale matrix} would be beneficial for the speed of the algorithm. \review{We provide an analysis of the in-sample selection of $\alpha$ and the out-of-sample performance for various $\alpha$ in the forecasting study in \ref{sec:overfitting} and leave the development of more advanced selection schemes for future research.}

\section{Forecasting Study}\label{sec:forecasting_study}

\subsection{Day-ahead Electricity Market and Data}

For electricity produced on day $t$ and in hour $h$, the short-term electricity market in Germany is split in three major parts: The daily day-ahead auction on $t-1$ at 12:00 hours for 24 hourly delivery periods $h \in \{0, ..., 23\}$, the afternoon auction with quarter-hourly delivery periods on $t-1$, at 15:00 hours and the continuous intraday market. The market is organized by EPEX SPOT and Nordpool in the joint single day-ahead coupling (SDAC) as a pay-as-cleared auction, resembling the merit-order model for the electricity market \citep{bille2023forecasting,hirsch2024online,viehmann2017state}. At each day $t$, before the day-ahead auction at 12:00, we aim to generate forecasts for day $t+1$. Prior to forecasting, we update our models by taking into account the realized prices and forecasts for delivery day $t$. \review{Figure \ref{fig:online_forecasting_study} contrasts online learning with the expanding and rolling window batch learning, two popular schemes for forecasting studies. Note that in the online learning scheme, we only use the new observation for updating the model, while in the batch learning schemes, we re-use all or a fixed window of previous observations. The rolling window scheme is popular in the EPF community \citep[see e.g.][]{lago2021forecasting,nowotarski2018recent} while online learning schemes are a rather recent development in EPF \citep[e.g.][]{berrisch2024multivariate, zaffran2022adaptive,hirsch2024online}.}

\begin{figure}[htb]
    \centering
    \resizebox{\linewidth}{!}{\begin{tikzpicture}[    
    node_initial_batch/.style={rectangle, draw=Blue, fill=Blue!10, very thick, minimum size=10mm},
    node_initial_batch_last/.style={rectangle, draw=White, fill=Blue!5, very thick, minimum size=10mm},
    node_step/.style={rectangle, draw=Green, fill=Green!10, very thick, minimum size=10mm},
    node_step_last/.style={rectangle, draw=White, fill=Green!5, very thick, minimum size=10mm},
]

\node[node_initial_batch, minimum width=5cm] (batch_0) [
    label={[label distance=0.25cm, align=left] above:\raggedright\text{\Large Batch Expanding Window}}
] {Initial set $i = 0, ..., N$};
\node[node_initial_batch, minimum width=6cm] (batch_1) [below=0.25cm of batch_0.south west, anchor=north west] {Dataset $i = 0, ..., N+1$};
\node[node_initial_batch, minimum width=7cm] (batch_2) [below=0.25cm of batch_1.south west, anchor=north west] {Dataset $i = 0, ..., N+2$};
\node[node_initial_batch, minimum width=8cm] (batch_3) [below=0.25cm of batch_2.south west, anchor=north west] {Dataset $i = 0, ..., N+3$};
\node[node_initial_batch, minimum width=9cm] (batch_4) [below=0.25cm of batch_3.south west, anchor=north west] {Dataset $i = 0, ..., N+4$};
\node[node_initial_batch_last, minimum width=10cm] (batch_5) [below=0.25cm of batch_4.south west, anchor=north west] {...};

\draw[-] (9, 1) to (9, -7.5);

\node[node_initial_batch, minimum width=5cm] (slide_0) [
    label={[label distance=0.25cm, align=left] above:\raggedright\text{\Large Batch Rolling Window}},
    right=8cm of batch_0
] {Initial set $i = 0, ..., N$};
\node[node_initial_batch, minimum width=5cm] (slide_1) [below right=0.25cm and 1cm of slide_0.south west, anchor=north west] {Dataset $i = 1, ..., N+1$};
\node[node_initial_batch, minimum width=5cm] (slide_2) [below right=0.25cm and 1cm of slide_1.south west, anchor=north west] {Dataset $i = 2, ..., N+2$};
\node[node_initial_batch, minimum width=5cm] (slide_3) [below right=0.25cm and 1cm of slide_2.south west, anchor=north west] {Dataset $i = 3, ..., N+3$};
\node[node_initial_batch, minimum width=5cm] (slide_4) [below right=0.25cm and 1cm of slide_3.south west, anchor=north west] {Dataset $i = 4, ..., N+4$};
\node[node_initial_batch_last, minimum width=5cm] (slide_5) [below right=0.25cm and 1cm of slide_4.south west, anchor=north west] {...};

\draw[-] (22, 1) to (22, -7.5);

\node[node_initial_batch, minimum width=5cm] (online_0) [
    label={[label distance=0.25cm, align=left]above:\text{\Large Online Learning}}, 
    right=8cm of slide_0
    ] {Initial set $i = 0, ..., N$};
\node[node_step] (online_1) [below=0.25cm of online_0.south east, anchor=north west] {$N$+1};
\node[node_step] (online_2) [
    label={[label distance=0.25cm, align=left]left:{New observations \\ added row-by-row.}}, 
    below=0.25cm of online_1.south east, anchor=north west] {$N$+2};
\node[node_step] (online_3) [below=0.25cm of online_2.south east, anchor=north west] {$N$+3};
\node[node_step] (online_4) [below=0.25cm of online_3.south east, anchor=north west] {$N$+4};
\node[node_step_last] (online_5) [below=0.25cm of online_4.south east, anchor=north west] {...};

\end{tikzpicture}} 
    \caption{Repeated Batch Learning vs. Online Learning for the forecasting study.}
    \label{fig:online_forecasting_study}
\end{figure}

\review{We use the same data set as in \cite{lipiecki2024postprocessing}, which consists of electricity prices for the German day-ahead market from 2015-01-01 \thirdreview{to 2020-12-31}. In line with previous works, we use the data until 2018-12-26 as initial training set, leaving 1831 observations and therefore more than 4 years, as it is best practice \citep{lago2021forecasting}, for out-of-sample testing. This dataset has been employed in various studies and to enable comparability, we further split the dataset in two sub-samples, that is 2018-12-27 to 2020-12-31 (736 days) and 2021-01-01 to 2023-12-31 (1095 days).} Additionally, the data set contains day-ahead renewable production forecasts, load forecasts and prices for fundamental commodities. All features are briefly described in Table \ref{tab:fundamental_variables}. \review{We use an incremental mean-variance scaling to \secondreview{pre}-process the electricity prices. We denote the electricity price for day $t$ and hour $h \in 0, ..., 23$ as $P_{th}$. We have $y_{th} = (P_{th} - \tilde{\mu}_{th}) / \tilde{\sigma}_{th}$, where $\tilde{\mu}_{th} = 1/t \sum_i^t P_{ih}$ and $\tilde{\sigma}_{th} = \sqrt{1/t \sum_i^t (P_{ih} - \tilde{\mu}_{ih})^2}$ are the mean and standard deviation up to observation $t,h$ and therefore have $\mat{Y} = (\vec{y}_0, ..., \vec{y}_{23})$ as the 24-dimensional ($D=H=24$) response matrix. We find that this stabilizes the estimation of covariance matrices and the normalization can be re-applied after the estimation, i.e. the covariance of $\vec{P}_t$ corresponds to $\operatorname{diag}(\tilde{\vec{\sigma}}_t) \mat{\Sigma} \operatorname{diag}(\tilde{\vec{\sigma}}_t)$, where $\mat{\Sigma}$ is estimated based on $\vec{y}_t$. Incremental updates of the mean and variance are straight-forward in an online learning setting using Welford's method \citep{welford1962note}.} Note that, in this application study, we have $T$ corresponding to $N$ and $H$ corresponding to $D$ in the general notation. 

\begin{table}[thb]
    \centering
    \resizebox{\textwidth}{!}{%
    \begin{tabular}{llll}
        \toprule
            Variable                                    &   Description                                         &  Resolution       &   Source       \\
            \midrule
            $\operatorname{ResLoad}_{t,h}$              &   Day-ahead residual load forecast      &   Hourly          &   ENTSO-E \\
            $\overline{\operatorname{ResLoad}}_{t}$     &   Day-ahead baseload residual load forecast $\frac{1}{H}\sum_{h=1}^H\operatorname{ResLoad}_{t,h}$ & Daily & ENTSO-E  \\
            $\operatorname{EUA}_{t}$                    &   Log European Union Emission Allowance (EUA) prices          &   Daily           &   \url{investing.com} \\
            $\operatorname{Gas}_{t}$                    &   Log natural gas prices              &   Daily           &   \url{investing.com} \\
            $\operatorname{Coal}_{t}$                   &   Log coal prices                     &   Daily           &   \url{investing.com} \\
            $\operatorname{Oil}_{t}$                    &   Log oil prices                      &   Daily           &   \url{investing.com} \\
            $\operatorname{WD}_{t}$                     &   Weekday dummies                     &   Daily           &   Calendar \\
        \bottomrule
    \end{tabular}%
    }%
    \caption{Variables from the data set of \cite{marcjasz2023distributional} and \cite{lipiecki2024postprocessing}. \secondreview{Wind power production, solar power production and system load forecasts are summarized in the residual load and are retrieved from the European Network of Transmission System Operators for Electricity (ENTSO-E).}}
    \label{tab:fundamental_variables}
\end{table}

\subsection{Model Definition and Benchmarks}

We propose modeling the multivariate distribution of the day-ahead electricity prices in increasing complexity. \review{If possible}, models are updated online, i.e. using only the new data for each day. We differentiate between an \emph{adaptive} estimation, which is updating a single, unconditional distributional parameter and the full \emph{conditional} estimation linking the distribution parameter to explanatory variables. \secondreview{Table~\ref{tab:models}} shows the increasing model complexity. \secondreview{For all three complexity levels, we include a reference model that assumes independence to showcase the value-added of modeling the dependence structure.} We start with the established LEAR models and naively estimate the unconditional residual distribution (denoted as LEAR-N(0, $\sigma$) and LEAR-N(0, $\Sigma$)). \review{Additionally, we employ two univariate benchmarks. First, we use a conformal prediction (CP) approach, which is based on the LEAR model and uses a combination of the split-conformal prediction (SCP, \secondreview{during initial training}) and adaptive conformal prediction \citep[ACI, see][\secondreview{during the online training}]{gibbs2021adaptive,gibbs2024conformal}. We use absolute residual scores and a calibration set of \secondreview{200 days per delivery hour}. \secondreview{This model is denoted as LEAR-CP.} Conformal prediction is generally a univariate post-processing method, since it is centered on issuing prediction intervals and therefore not suitable for the generation of ensemble forecasts. Multivariate conformal prediction approaches exist, but are in their infancy and focussed on a notion of multivariate quantiles.\footnote{As a simple workaround, we have tried to combine CP with a copula-based approach using a \secondreview{probability integral transform (PIT).} However, the conformal predictive density approximated using 199 quantiles was not sufficient to generate useful ensembles in the inverse transformation, i.e. the step from the simulated copula on $\mathcal{U}(0, 1)$ to the original domain failed.} Second, we use a GARCH(1,1) model with a Gaussian distributional assumption (denoted as LEAR-GARCH). We estimate the GARCH model on the residuals of the LEAR model using the \texttt{arch} package in Python \citep{kevin_sheppard_2024_14035889}. \secondreview{This is the only model that we cannot run online in the study.}} 

\begin{table}
    \resizebox{\textwidth}{!}{%
    \begin{tabular}{p{0.4\textwidth}p{0.95\textwidth}}
\toprule
    Abbreviation                        &   Short Description \\ 
\midrule
    LEAR-\{structure\}                  &   LEAR Model (Equation \ref{eq:mean_model}) for the mean and probabilistic model according to {structure}, either $N(0, \sigma$), $N(0,\Sigma)$, conformal prediction (CP) or GARCH. 
                                            Estimated online for all but the LEAR-GARCH. \\
    ODR-\{dep\}                         &   (Univariate) distributional regression model (Equation \ref{eq:mean_model}, \ref{eq:model-scale}, \ref{eq:model-dof}) and independence or (sparse) Gaussian copula (dep=GC and dep=SPGC. Estimated online. \\
    MODR-\{method\}-\{param\}-\{dep\}   &   Multivariate distributional regression models (Equation \ref{eq:mean_model}, \ref{eq:model-scale}, \ref{eq:model-dof}) estimated with method (OLS, LASSO) and the parameterization of $\boldsymbol{\Omega}$. 
                                            Assuming independence (dep=IND) or modelling the dependence structure (dep=$\Sigma$). 
                                            Estimated online. \\
\bottomrule 
\end{tabular}
}%
    \caption{Overview and brief descriptions of the models included in the forecasting study.}\label{tab:models}
\end{table}

We increase complexity by moving to a full distributional model for the marginals and adding an adaptive estimation of the dependence structure using the Gaussian copula \secondreview{(denoted as ODR-Copula), where the Copula is either the independence Copula (ODR-IND), or a (sparse) Gaussian copula (ODR-GC and ODR-SPGC).}  Lastly, we estimate the full multivariate distribution in a conditional way using the proposed multivariate online distributional regression approach \secondreview{(denoted as MODR-Method-Parameterization-Dependence)}. We describe the full model in the following and note that we additionally describe the hyperparameters \thirdreview{in the online supplement}. 

We model the mean/location for all regression models, \review{also for the LEAR models,} by \begin{equation}\label{eq:mean_model}
    \review{
    \begin{aligned}
        g_\mu(\mu_{t, h}) &= \beta_{\mu, 0, h} 
            + \sum_{l=1}^{L=7} \beta_{\mu, l, h} y_{t-l, h}
            + \sum_{i \in \{0, ..., 23\} \setminus h} \beta_{\mu, 8+i, h} y_{t-1, i} 
            + \sum_{w=1}^{W=6} \beta_{\mu, 30+w, h} \operatorname{WD}_{t, h}    \\
            &+ \beta_{\mu, 37, h} \operatorname{min}(\vec{y}_{t-1}) 
            + \beta_{\mu, 38, h} \operatorname{max}(\vec{y}_{t-1}) 
            + \beta_{\mu, 39, h} \operatorname{Q}_{10}(\vec{y}_{t-1})
            + \beta_{\mu, 40, h} \operatorname{Q}_{90}(\vec{y}_{t-1}) \\
           &+ \beta_{\mu, 41, h} \operatorname{ResLoad}_{t, h} 
            + \beta_{\mu, 42, h} \operatorname{EUA}_{t} 
            + \beta_{\mu, 43, h} \operatorname{Gas}_{t} 
            + \beta_{\mu, 44, h} \operatorname{Coal}_{t} 
            + \beta_{\mu, 45, h} \operatorname{Oil}_{t}.                     
      \end{aligned}}
\end{equation} 
We model the scale parameters for univariate distributional models, as well as the elements of the Cholesky-factor $\mat{\Omega} = \transpose{(\mat{A}^{-1})}(\mat{A}^{-1})$, and the elements of the diagonal matrix $\mat{A}$ in the LRA-based scale matrices by \begin{equation}
    \review{\begin{aligned}\label{eq:model-scale}
        g_\theta(\theta_{t, h, h}) 
            &= \beta_{\theta, 0, h} 
            + \beta_{\theta, 1, h} \operatorname{SignedSquare}\left(\mat{\Sigma}_{h,h}^{[t-1:t-7]}\right)^{-1}
            + \beta_{\theta, 2, h} \operatorname{ResLoad}_{t, h} \\ 
            &+ \beta_{\theta, 4, h} \operatorname{EUA}_{t} 
            + \beta_{\theta, 5, h} \operatorname{Gas}_{t} 
            + \beta_{\theta, 6, h} \operatorname{Coal}_{t} 
            + \beta_{\theta, 7, h} \operatorname{Oil}_{t},
    \end{aligned}}
\end{equation}
where $\operatorname{SignedSquare}(a) = \operatorname{sign}(a)\sqrt{|a|}$ is the signed square root and $\mat{\Sigma}^{[t-1:t-7]}$ is the rolling empirical covariance matrix of $\mat{y}_t$ for the last 7 days. \review{Note that we use the inverse of the signed square, since we are working on the precision matrix. For the univariate models, we replace this accordingly with the empirical rolling standard deviation.} For the LRA-based parameterization, we choose $r=2$ and model the elements of $\mat{V}$ as 
\begin{equation}
    \review{\begin{aligned}\label{eq:model-scale-v}
        g_v(v_{t, h, 0}) 
            &= \beta_{v, 0, h} 
            + \beta_{v, 1, h} \operatorname{SignedSquare}\left(\mat{\Sigma}_{h,h}^{[t-1:t-7]}\right)
            + \beta_{v, 2, h} \overline{\operatorname{ResLoad}}_{t}   \\
            &+ \beta_{v, 3, h} \operatorname{EUA}_{t} 
            + \beta_{v, 4, h} \operatorname{Gas}_{t} 
            + \beta_{v, 5, h} \operatorname{Coal}_{t} 
            + \beta_{v, 6, h} \operatorname{Oil}_{t},
    \end{aligned}}
\end{equation}
\begin{equation}
    \review{\begin{aligned}\label{eq:model-v-2}
        g_v(v_{t, h, 1}) = \sum_w^{W=6} \beta_{v, 14+w, h} \operatorname{WD}_{t, h}
    \end{aligned}}
\end{equation}
that is, the first rank takes most of the fundamental variables, while the second rank contains the weekday binary variables. The degrees of freedom are modeled as\begin{equation}
    \begin{aligned}\label{eq:model-dof}
        g_\nu(\nu_{t}) 
            &= \beta_{\nu, 0} 
            + \beta_{\nu, 1} \operatorname{mean}(\vec{y}_{t-1})  
            + \sum_w^{W=6} \beta_{\nu, 1+w, h} \operatorname{WD}_{t, h}
            + \beta_{\nu, 8} \overline{\operatorname{ResLoad}}_{t} \\
            &+ \beta_{\nu, 9} \operatorname{EUA}_{t} 
            + \beta_{\nu, 10} \operatorname{Gas}_{t} 
            + \beta_{\nu, 11} \operatorname{Coal}_{t} 
            + \beta_{\nu, 12} \operatorname{Oil}_{t}.
    \end{aligned}
\end{equation} The univariate models are therefore a slight simplification compared to the models used in \cite{hirsch2024online}, thereby the multivariate distributional regression models and the Copula-based approaches are better comparable. \thirdreview{We fit the Copula by transforming the in-sample data to the uniform space $\mat{u}$ by the probability integral transform (PIT) and subsequently transforming to the $\mathcal{N}(0, 1)$ space $\vec{n}$, where we can track the covariance matrix online \citep[see][and the online supplement]{dasgupta2007line, arbenz2013bayesian}.}
We employ a second model, where we sparsify the estimated dependence matrix of the Gaussian copula \secondreview{using} the graphical LASSO \citep{friedman2008sparse}.

\subsection{Forecast Evaluation and Scoring Rules}

\review{Forecast evaluation should follow the well-known principle of sharpness subject to calibration \citep{gneiting2007probabilistic}}.
\review{We check the calibration of the forecasts by calculating joint prediction bands (JPB) from the simulations. \secondreview{JPB aim to cover the true price vector with probability $1-\alpha$ and therefore generalize marginal prediction intervals to the multivariate case \citep{staszewskabystrova2011bootstrap, luetkepohl2015comparison}. They have been used in energy market forecasting by \cite{serafin2022trading, chen2025probabilistic} before and proven effective in supporting trading decision-making.} As it is standard in probabilistic forecasting, we evaluate both marginal and multivariate quality of the forecasts. Therefore, we} employ the root mean square error (RMSE), mean absolute error (MAE) and the continuous ranked probability score (CRPS), \review{which focus on point prediction and the marginal distribution.} We employ four well-established multivariate probabilistic scoring rules: The Energy Score (ES), the Dawid-Sebastiani Score (DSS), the Variogram Score (VS) and the Log-Score (LS). Additionally, we test for statistically significant score differences using the well-established Diebold-Mariano test. 

\review{Let us briefly review some recent results with respect to the four multivariate scores: \begin{itemize}
    \item The ES, DSS and LS are all able to reliably detect mis-specifications in the mean structure of the multivariate distribution \citep{marcotte2023regions}, while the VS (by design) is not sensitive. However, \cite{pinson2013discrimination,alexander2024evaluating} discuss the discrimination ability of the ES. While widely used for multivariate forecast evaluation, the ES has been shown to have low discrimination ability with respect to misspecified covariance structures, \secondreview{especially in dimensions  $D > 10$ and \cite{alexander2024evaluating} recommend to use the VS in addition to the ES.}
    \item In a similar vein, \cite{marcotte2023regions} show that the ES has low reliability, i.e. statistical power to discriminate between a correctly and incorrectly specified dependence model for multivariate forecasts compared to the DSS and the LS, with the VS being somewhat in the middle. Their analysis also reflects the role of the number of test samples (which, with more than 1800 out-of-sample days, is not a concern in our study) and the number of sample paths $M$. Interestingly, they find that the ES and the VS are complimentary, i.e. the VS is more reliable in cases where the ES is not and vice versa, and they find the reliability of the VS to increase non-monotonically with respect to the dimension $D$ and the number of samples $M$.
    \item In contrast, \cite{ziel2019multivariate} find that the ES, in combination with the Diebold-Mariano test, is able to detect the correct model specification in a simulation study. Nevertheless, they still recommend the use of multiple scores.
\end{itemize}%
\thirdreview{In practice, scoring rules are chosen based on the application. In this paper, we evaluate multiple scores as suggested by \cite{ziel2019multivariate,alexander2024evaluating} and contrast the results. However, since the ODR, Copula and MODR approaches are likelihood-based, we have a focus on the LS and the related DSS to align evaluation and estimation metric.}
\thirdreview{For the formal definition of the scoring rules we refer the reader to the original references \citep{gneiting2007strictly, scheuerer2015variogram, dawid1999coherent} and the online supplement. We use implementations provided in the Python package \texttt{scoringrules} \citep{zanetta2024scoringrules}}.
}

\review{A joint prediction band for the $1-\alpha$ coverage is defined by the lower bound $\vec{l}_t = (l_{t,0}, ..., l_{t,23})$ and upper bound $\vec{u}_t = (u_{t,0}, ..., u_{t,23})$, such that \begin{equation*}
    \mathbb{P}_t\left(\vec{l}_{t} \leq \vec{y}_{t} \leq \vec{u}_{t}  \; \forall \; h \in \{0, ..., 23\} \right) = 1 - \alpha,
\end{equation*}
that is, the true price trajectory~$\vec{y}_t$ is \emph{fully} covered by the JPB with probability $1-\alpha$. This is in difference to marginal prediction intervals based on predicted quantiles, which consider element-wise coverage. There are multiple algorithms to construct such bands and following \cite{serafin2022trading, chen2025probabilistic}, we use the neighboring paths method described by \cite{staszewskabystrova2011bootstrap, luetkepohl2015comparison}. The method is based on iteratively removing paths from the ensemble forecast $\mat{F}$ \thirdreview{of shape $T \times H \times M$ of $M = 2500$ samples}, such that the envelope of the remaining paths covers the true price trajectory~$\vec{y}_t$ with probability~$1-\alpha$. It should be noted that JPB are in general not unique, i.e. there are multiple sets of lower and upper bounds that cover the true trajectory with the desired probability. We compare the mis-coverage of the \secondreview{joint prediction band}, which is defined as \begin{equation}
    \operatorname{MC}_{1-\alpha} = \frac{1}{T}\sum^T_{t=0}\boldsymbol{1}\left(\vec{l}_{t} \leq \vec{y}_{t} \leq \vec{u}_{t} \; \forall \; h \in \{0, ..., 23\}\right) - (1-\alpha), 
\end{equation} 
and gives a measure of the multivariate calibration of the joint prediction. Additionally, we evaluate the mean width of the \secondreview{JPB}, which is defined as \begin{equation}
    \operatorname{JPBW}_{1-\alpha} = \frac{1}{TH}\sum^T_{t=0}\sum^H_{h=0}\left(\vec{u}_{t,h} - \vec{l}_{t,h}\right),
\end{equation} which gives a measure of the efficiency of the prediction bands and can be interpreted as the area of the prediction band divided by the dimensions.} 

Conclusions on the performance of forecasting models cannot be \review{derived by} looking at aggregate scores alone, but need to be drawn by evaluating whether the differential between the loss series of two models is statistically significantly from zero \citep{diebold2002comparing, diebold2015comparing}. For the DM-test, we evaluate the differential of two score series $\Delta\vec{s}^{\mathcal{A},\mathcal{B}} = \vec{s}^\mathcal{A} - \vec{s}^\mathcal{B}$, where $\vec{s}^\mathcal{A} = (s_0^\mathcal{A}, ..., s_T^\mathcal{A})$ are the scores for each scoring rule at $t$ for model $\mathcal{A}$ respectively $\mathcal{B}$. We provide two one-sided and hence complementary tests. \review{To ensure validity of the DM-test, we check stationarity of the differential series $\Delta\vec{s}^{\mathcal{A},\mathcal{B}}$ by the augmented Dickey-Fuller test \citep[ADF,][]{dickey1979distribution, cheung1995lag}.}
\section{Results}\label{sec:results}

\secondreview{
    We present the results of our forecasting study in three parts. First, we analyze the predicted time-varying dependence structure and discuss the economic interpretation. Second, we present the forecasting accuracy of all models using the scoring rules presented above. Lastly, we discuss the computational costs of the different models and compare the online and batch estimation.
}

\begin{figure}[htb]
    \centering
    \includegraphics[width=\textwidth]{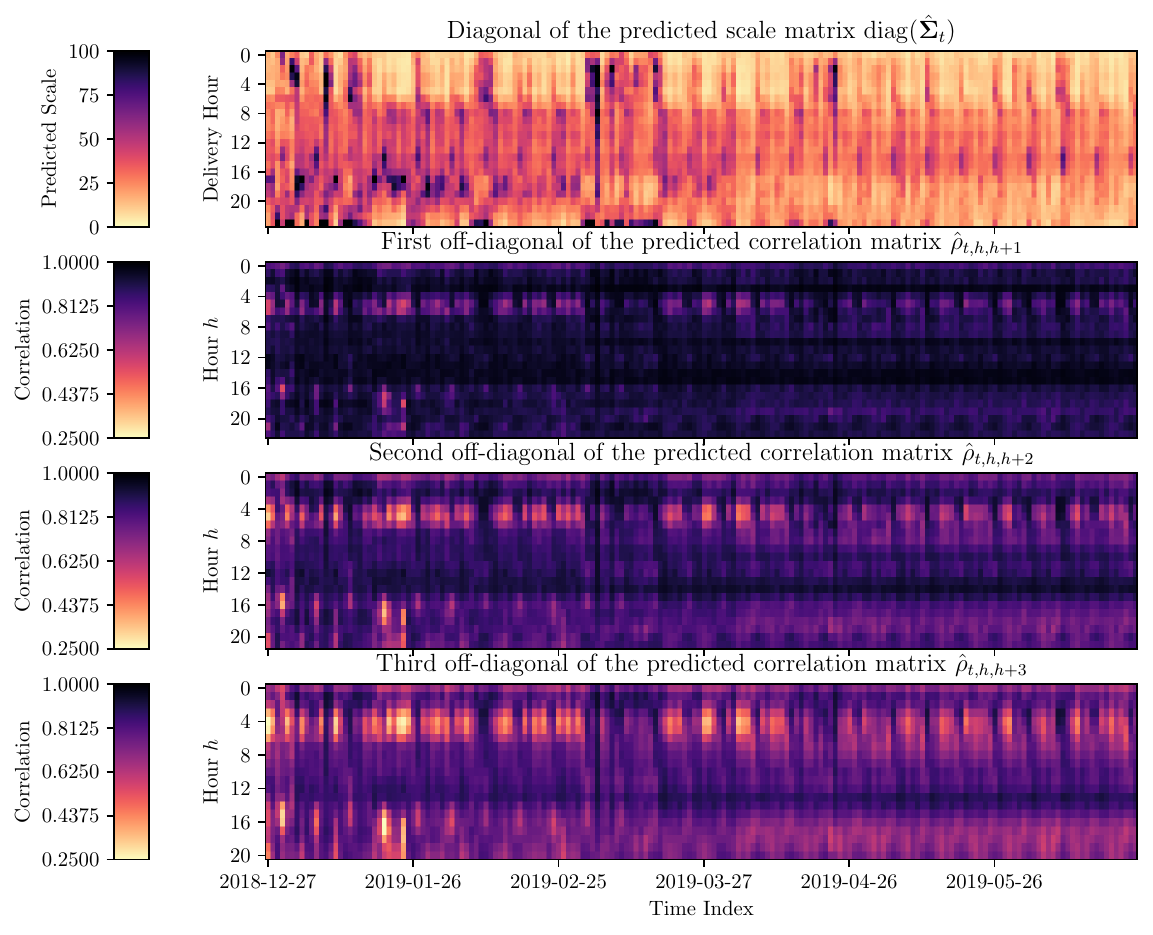}
    \caption{Evolution of the predicted volatility (standard deviation) and correlation matrix for the {MODR-OLS-MCD-$\Sigma$} model over time. The lower three panels show the 1st, 2nd and 3rd off-diagonal of the correlation matrix over time. We see that the correlation is the lowest around the hours 5-6, which corresponds to the start of the morning ramp. The correlation between hours $h$ and $h+i$ decays stronger for days with lower levels of volatility and vice versa. There are some weekly patterns visible.}
    \label{fig:insample_analysis}
\end{figure}

\review{\subsection{Forecasting the Time-varying Dependence Structure}}
\secondreview{
    Figure~\ref{fig:insample_analysis} plots the evolution of the predicted volatility (standard deviation) and correlation matrix for the {MODR-OLS-MCD-$\Sigma$} model over time. The lower three panels show the 1st, 2nd and 3rd off-diagonal of the correlation matrix over time. We see that the correlation is the lowest around the hours 5-7, which correspond to start of the morning ramp. The correlation between hours $h$ and $h+i$ decays stronger for days with lower levels of volatility and vice versa. There are some weekly patterns visible: The dependence between the hours 5-7 and 16-19 tapers off more strongly on working days. This is likely driven by stronger shapes in prices and residual demand. Overall, we see that the dependence structure is not constant over time, which underlines the importance of including a time-varying dependence structure in multivariate electricity price forecasting.
}

\FloatBarrier
\review{
    \subsection{Forecasting Accuracy}
} 

\begin{figure}[htb]
    \centering
    \includegraphics[width=\textwidth]{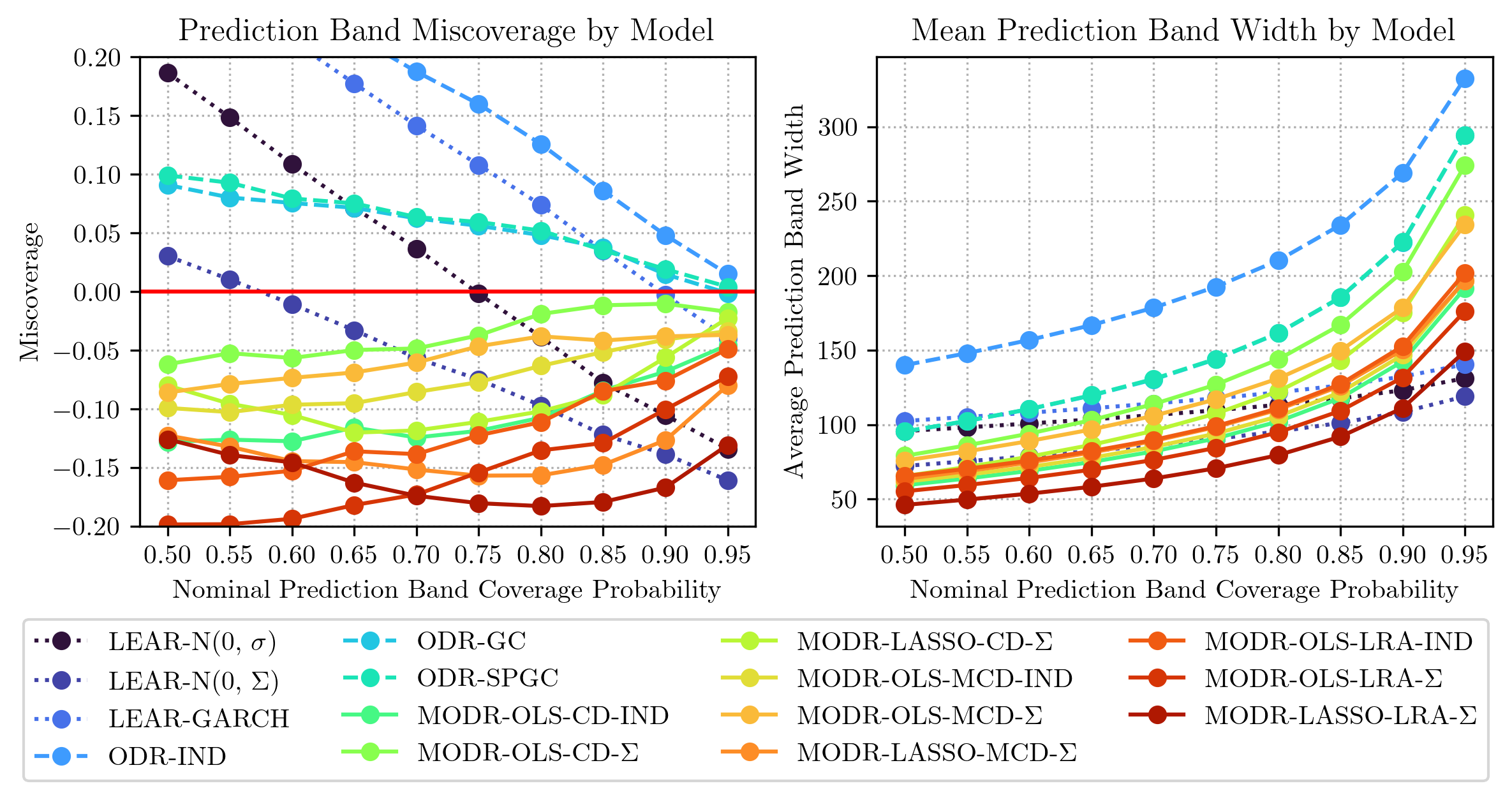}
    \caption{Calibration of the $50\%, 55\%, ...,  95\%$-joint prediction bands. A positive miscoverage indicates over-coverage, while a negative value indicates under-coverage. The second panel shows the mean width of the $50\%, 55\%, ...,  95\%$-joint prediction bands.}
    \label{fig:calibration_bands}
\end{figure}

\begin{figure}[htb]
    \centering
    \includegraphics[width=\linewidth]{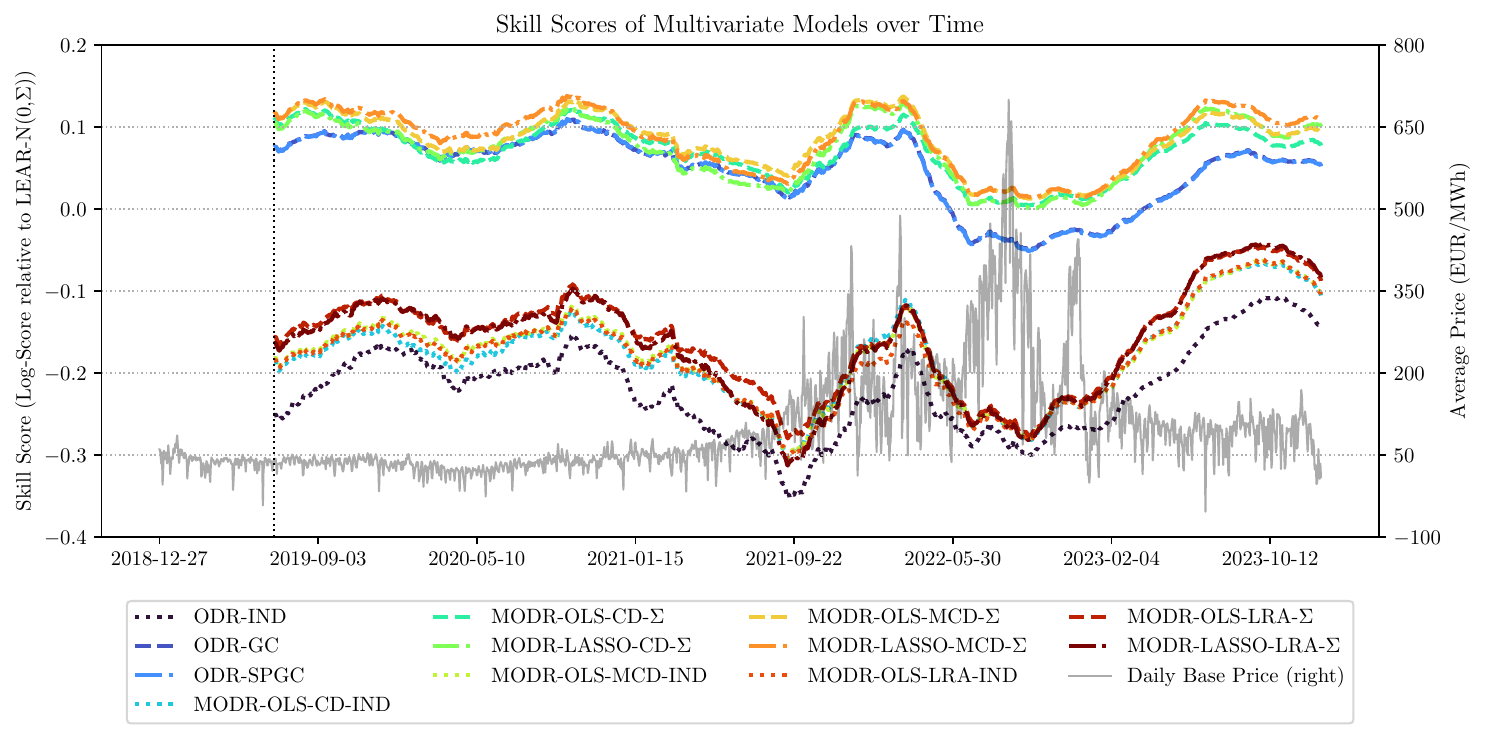}
    \caption{Rolling 182-day average of skill scores. The skill score is defined as $\operatorname{SS} = 1 - \operatorname{LS}_\text{model} / \operatorname{LS}_\text{baseline}$, where the baseline is the LEAR+N(0,$\Sigma$) model. A positive skill score indicates an improvement over the baseline. The secondary $y$-axis shows the daily average day-ahead price.}
    \label{fig:skill_scores}
\end{figure}

\secondreview{
We first examine the calibration of the joint prediction bands in Figure~\ref{fig:calibration_bands}. The figure shows miscoverage and mean JPB width for the $50\%, 55\%, ...,  95\%$ nominal coverage joint prediction bands. Table \ref{tab:results_scores} gives the results for the scoring rules for each model. Figure~\ref{fig:skill_scores} gives daily skills scores of the multivariate distributional regression models over the benchmarks {LEAR-N(0,$\Sigma$)}. As a mental guidance for the increasing complexity of the online regression models, remember Section \ref{sec:forecasting_study} and Table \ref{tab:models}: The LEAR-based models have an adaptive, but unconditional estimation for the probabilistic forecast. The univariate online distributional models (ODR) employ conditional estimation for all marginal distributional parameters and an adaptive, but unconditional estimation for the Copula, while the multivariate online distributional regression (MODR) models yield an online estimated conditional multivariate distribution. We summarize the main results before we discuss the results in more detail.
}%

In terms of the calibration of the joint prediction bands, the multivariate distributional regression models yield \thirdreview{good} coverage across all widths. The MODR models for the CD and MCD parameterization yield prediction intervals closest to the nominal coverage, while the LRA-based models generally yield too small JPB. \thirdreview{The LASSO-estimated MODR models generally have slightly worse coverage than the OLS-estimated counterparts.} For the lower nominal coverage levels, the copula-based models yield too conservative (too wide) JPB. \thirdreview{The LEAR-based models exhibit over-coverage for low nominal coverage levels and under-coverage for high nominal coverage. Here, the Gaussian distribution does not account for heavy tails and assigns to much mass to the center of the distribution. The effect is notably stronger for the model assuming independence.} \fourthreview{For the LEAR-GARCH, we see that the model provides better joint coverage during 2021 to 2023, and, in the latter spike analysis, especially during periods with high min-to-max spreads, although at the cost of over-coverage in calm periods (see Figures 2 and 3 in the online supplement). Generally, the ODR-IND and LEAR-GARCH show similar joint coverage, as both model time-varying and conditional volatility.} The JPB width of the multivariate distributional regression models is also lower than the univariate distributional regression models, however, the under-covering LEAR-based models yield the narrowest bands.

 \thirdreview{For the scoring rules,} the LEAR models yield the best performance in terms of RMSE, MAE and CRPS. These models yield sharp predictions, but at the cost of poor coverage of the joint prediction bands. This is also visible in the very small JPBW \thirdreview{and consistent with the notion that the CRPS rewards underestimation of variance over overestimation of variance \citep{buchweitz2025asymmetric}.}
The Copula-based models yield the best performance in terms of the ES. The MCD-based multivariate distributional regression model yields the best overall performance in terms of the LS and DSS as well as a \secondreview{second-best performance in terms of the VS}.

\begin{table}[h!]
\begin{center}
    \centering \small Panel A: Full sample (2018-12-27 to 2023-12-31) 
    \resizebox{\textwidth}{!}{\begin{tabular}{R{6cm}L{1.9cm}L{1.9cm}L{1.9cm}L{1.9cm}L{1.9cm}L{1.9cm}L{1.9cm}L{1.9cm}L{1.9cm}L{1.9cm}}
\toprule
 & RMSE & MAE & CRPS & $\text{VS}_{p=0.5}$ & $\text{VS}_{p=1}$ & ES & DSS & LS & $\text{MC}_{0.95}$ & $\text{JPBW}_{0.95}$ \\
\midrule
LEAR-N(0, $\sigma$) & {\cellcolor[HTML]{053061}} \color[HTML]{F1F1F1} \bfseries 31.177 & {\cellcolor[HTML]{053061}} \color[HTML]{F1F1F1} \bfseries 18.214 & {\cellcolor[HTML]{266CAF}} \color[HTML]{F1F1F1} 13.677 & {\cellcolor[HTML]{E17860}} \color[HTML]{F1F1F1} 4.709 & {\cellcolor[HTML]{327CB7}} \color[HTML]{F1F1F1} 887.202 & {\cellcolor[HTML]{2065AB}} \color[HTML]{F1F1F1} 81.201 & {\cellcolor[HTML]{EB9172}} \color[HTML]{F1F1F1} 176.367 & {\cellcolor[HTML]{67001F}} \color[HTML]{F1F1F1} 110.038 & {\cellcolor[HTML]{4695C4}} \color[HTML]{F1F1F1} -0.134 & {\cellcolor[HTML]{F2F5F6}} \color[HTML]{000000} 131.129 \\
LEAR-N(0, $\Sigma$) & {\cellcolor[HTML]{053061}} \color[HTML]{F1F1F1} 31.182 & {\cellcolor[HTML]{053061}} \color[HTML]{F1F1F1} 18.219 & {\cellcolor[HTML]{266CAF}} \color[HTML]{F1F1F1} 13.678 & {\cellcolor[HTML]{053061}} \color[HTML]{F1F1F1} \bfseries 3.396 & {\cellcolor[HTML]{053061}} \color[HTML]{F1F1F1} \bfseries 654.953 & {\cellcolor[HTML]{073467}} \color[HTML]{F1F1F1} 78.603 & {\cellcolor[HTML]{185493}} \color[HTML]{F1F1F1} 124.345 & {\cellcolor[HTML]{96C7DF}} \color[HTML]{000000} 84.057 & {\cellcolor[HTML]{65A9CF}} \color[HTML]{F1F1F1} -0.161 & {\cellcolor[HTML]{DEEBF2}} \color[HTML]{000000} 118.914 \\
LEAR-CP & {\cellcolor[HTML]{063264}} \color[HTML]{F1F1F1} 31.268 & {\cellcolor[HTML]{0A3B70}} \color[HTML]{F1F1F1} 18.374 & {\cellcolor[HTML]{5CA3CB}} \color[HTML]{F1F1F1} 14.402 & {\cellcolor[HTML]{FFFFFF}} \color[HTML]{000000} \color{white}  & {\cellcolor[HTML]{FFFFFF}} \color[HTML]{000000} \color{white}  & {\cellcolor[HTML]{FFFFFF}} \color[HTML]{000000} \color{white}  & {\cellcolor[HTML]{FFFFFF}} \color[HTML]{000000} \color{white}  & {\cellcolor[HTML]{FFFFFF}} \color[HTML]{000000} \color{white}  & {\cellcolor[HTML]{67001F}} \color[HTML]{F1F1F1} -0.655 & {\cellcolor[HTML]{053061}} \color[HTML]{F1F1F1} \bfseries 53.751 \\
LEAR-GARCH & {\cellcolor[HTML]{053061}} \color[HTML]{F1F1F1} 31.181 & {\cellcolor[HTML]{053061}} \color[HTML]{F1F1F1} 18.218 & {\cellcolor[HTML]{053061}} \color[HTML]{F1F1F1} \bfseries 12.992 & {\cellcolor[HTML]{F6B394}} \color[HTML]{000000} 4.536 & {\cellcolor[HTML]{2A71B2}} \color[HTML]{F1F1F1} 850.636 & {\cellcolor[HTML]{053061}} \color[HTML]{F1F1F1} \bfseries 78.312 & {\cellcolor[HTML]{F9EBE3}} \color[HTML]{000000} 159.545 & {\cellcolor[HTML]{D6604D}} \color[HTML]{F1F1F1} 101.790 & {\cellcolor[HTML]{15508D}} \color[HTML]{F1F1F1} -0.040 & {\cellcolor[HTML]{F9F0EB}} \color[HTML]{000000} 140.459 \\
ODR-IND & {\cellcolor[HTML]{7EB8D7}} \color[HTML]{000000} 36.040 & {\cellcolor[HTML]{2A71B2}} \color[HTML]{F1F1F1} 19.104 & {\cellcolor[HTML]{3783BB}} \color[HTML]{F1F1F1} 13.980 & {\cellcolor[HTML]{DE735C}} \color[HTML]{F1F1F1} 4.722 & {\cellcolor[HTML]{59A1CA}} \color[HTML]{F1F1F1} 1046.549 & {\cellcolor[HTML]{4997C5}} \color[HTML]{F1F1F1} 84.445 & {\cellcolor[HTML]{FAE9DF}} \color[HTML]{000000} 160.422 & {\cellcolor[HTML]{EA8E70}} \color[HTML]{F1F1F1} 99.266 & {\cellcolor[HTML]{0A3B70}} \color[HTML]{F1F1F1} 0.015 & {\cellcolor[HTML]{67001F}} \color[HTML]{F1F1F1} 332.441 \\
ODR-GC & {\cellcolor[HTML]{3480B9}} \color[HTML]{F1F1F1} 33.901 & {\cellcolor[HTML]{10457E}} \color[HTML]{F1F1F1} 18.511 & {\cellcolor[HTML]{1D5FA2}} \color[HTML]{F1F1F1} 13.501 & {\cellcolor[HTML]{2A71B2}} \color[HTML]{F1F1F1} 3.591 & {\cellcolor[HTML]{1F63A8}} \color[HTML]{F1F1F1} 793.156 & {\cellcolor[HTML]{053061}} \color[HTML]{F1F1F1} 78.372 & {\cellcolor[HTML]{1A5899}} \color[HTML]{F1F1F1} 124.869 & {\cellcolor[HTML]{246AAE}} \color[HTML]{F1F1F1} 77.934 & {\cellcolor[HTML]{053061}} \color[HTML]{F1F1F1} \bfseries -0.001 & {\cellcolor[HTML]{991027}} \color[HTML]{F1F1F1} 294.100 \\
ODR-SPGC & {\cellcolor[HTML]{3681BA}} \color[HTML]{F1F1F1} 33.964 & {\cellcolor[HTML]{114781}} \color[HTML]{F1F1F1} 18.522 & {\cellcolor[HTML]{1D5FA2}} \color[HTML]{F1F1F1} 13.508 & {\cellcolor[HTML]{2B73B3}} \color[HTML]{F1F1F1} 3.597 & {\cellcolor[HTML]{1F63A8}} \color[HTML]{F1F1F1} 793.049 & {\cellcolor[HTML]{053061}} \color[HTML]{F1F1F1} 78.417 & {\cellcolor[HTML]{195696}} \color[HTML]{F1F1F1} 124.646 & {\cellcolor[HTML]{246AAE}} \color[HTML]{F1F1F1} 77.964 & {\cellcolor[HTML]{063264}} \color[HTML]{F1F1F1} 0.004 & {\cellcolor[HTML]{991027}} \color[HTML]{F1F1F1} 294.163 \\
MODR-OLS-CD-IND & {\cellcolor[HTML]{D8E9F1}} \color[HTML]{000000} 38.898 & {\cellcolor[HTML]{7EB8D7}} \color[HTML]{000000} 20.207 & {\cellcolor[HTML]{CAE1EE}} \color[HTML]{000000} 15.410 & {\cellcolor[HTML]{68ABD0}} \color[HTML]{F1F1F1} 3.787 & {\cellcolor[HTML]{1B5A9C}} \color[HTML]{F1F1F1} 771.899 & {\cellcolor[HTML]{B6D7E8}} \color[HTML]{000000} 89.011 & {\cellcolor[HTML]{F2A17F}} \color[HTML]{000000} 174.092 & {\cellcolor[HTML]{FAC8AF}} \color[HTML]{000000} 95.446 & {\cellcolor[HTML]{185493}} \color[HTML]{F1F1F1} -0.046 & {\cellcolor[HTML]{F4A683}} \color[HTML]{000000} 191.574 \\
MODR-OLS-CD-$\Sigma$ & {\cellcolor[HTML]{DF765E}} \color[HTML]{F1F1F1} 46.742 & {\cellcolor[HTML]{F6B191}} \color[HTML]{000000} 23.493 & {\cellcolor[HTML]{F6B394}} \color[HTML]{000000} 17.463 & {\cellcolor[HTML]{569FC9}} \color[HTML]{F1F1F1} 3.747 & {\cellcolor[HTML]{1F63A8}} \color[HTML]{F1F1F1} 797.964 & {\cellcolor[HTML]{FBCCB4}} \color[HTML]{000000} 97.948 & {\cellcolor[HTML]{2065AB}} \color[HTML]{F1F1F1} 126.357 & {\cellcolor[HTML]{114781}} \color[HTML]{F1F1F1} 76.013 & {\cellcolor[HTML]{0C3D73}} \color[HTML]{F1F1F1} -0.017 & {\cellcolor[HTML]{B41C2D}} \color[HTML]{F1F1F1} 273.946 \\
MODR-LASSO-CD-$\Sigma$ & {\cellcolor[HTML]{67001F}} \color[HTML]{F1F1F1} 52.791 & {\cellcolor[HTML]{67001F}} \color[HTML]{F1F1F1} 26.483 & {\cellcolor[HTML]{67001F}} \color[HTML]{F1F1F1} 20.094 & {\cellcolor[HTML]{D7E8F1}} \color[HTML]{000000} 4.070 & {\cellcolor[HTML]{2C75B4}} \color[HTML]{F1F1F1} 860.208 & {\cellcolor[HTML]{67001F}} \color[HTML]{F1F1F1} 111.976 & {\cellcolor[HTML]{266CAF}} \color[HTML]{F1F1F1} 127.573 & {\cellcolor[HTML]{0D3F76}} \color[HTML]{F1F1F1} 75.543 & {\cellcolor[HTML]{0E4179}} \color[HTML]{F1F1F1} -0.023 & {\cellcolor[HTML]{CE4F45}} \color[HTML]{F1F1F1} 240.903 \\
MODR-OLS-MCD-IND & {\cellcolor[HTML]{D8E9F1}} \color[HTML]{000000} 38.937 & {\cellcolor[HTML]{7EB8D7}} \color[HTML]{000000} 20.196 & {\cellcolor[HTML]{C7E0ED}} \color[HTML]{000000} 15.379 & {\cellcolor[HTML]{8DC2DC}} \color[HTML]{000000} 3.861 & {\cellcolor[HTML]{1E61A5}} \color[HTML]{F1F1F1} 792.223 & {\cellcolor[HTML]{B1D5E7}} \color[HTML]{000000} 88.771 & {\cellcolor[HTML]{EE9677}} \color[HTML]{000000} 175.635 & {\cellcolor[HTML]{FBCCB4}} \color[HTML]{000000} 95.274 & {\cellcolor[HTML]{124984}} \color[HTML]{F1F1F1} -0.034 & {\cellcolor[HTML]{F19E7D}} \color[HTML]{000000} 196.792 \\
MODR-OLS-MCD-$\Sigma$ & {\cellcolor[HTML]{E37E64}} \color[HTML]{F1F1F1} 46.450 & {\cellcolor[HTML]{F7B799}} \color[HTML]{000000} 23.376 & {\cellcolor[HTML]{F9C2A7}} \color[HTML]{000000} 17.255 & {\cellcolor[HTML]{2B73B3}} \color[HTML]{F1F1F1} 3.596 & {\cellcolor[HTML]{175290}} \color[HTML]{F1F1F1} 747.954 & {\cellcolor[HTML]{FDDDCB}} \color[HTML]{000000} 96.739 & {\cellcolor[HTML]{0A3B70}} \color[HTML]{F1F1F1} 121.365 & {\cellcolor[HTML]{073467}} \color[HTML]{F1F1F1} 74.907 & {\cellcolor[HTML]{134C87}} \color[HTML]{F1F1F1} -0.037 & {\cellcolor[HTML]{D35A4A}} \color[HTML]{F1F1F1} 234.422 \\
MODR-LASSO-MCD-$\Sigma$ & {\cellcolor[HTML]{B3192C}} \color[HTML]{F1F1F1} 50.034 & {\cellcolor[HTML]{C2383A}} \color[HTML]{F1F1F1} 25.078 & {\cellcolor[HTML]{BF3338}} \color[HTML]{F1F1F1} 18.923 & {\cellcolor[HTML]{93C6DE}} \color[HTML]{000000} 3.872 & {\cellcolor[HTML]{2065AB}} \color[HTML]{F1F1F1} 804.906 & {\cellcolor[HTML]{C6413E}} \color[HTML]{F1F1F1} 105.786 & {\cellcolor[HTML]{053061}} \color[HTML]{F1F1F1} \bfseries 119.963 & {\cellcolor[HTML]{053061}} \color[HTML]{F1F1F1} \bfseries 74.646 & {\cellcolor[HTML]{276EB0}} \color[HTML]{F1F1F1} -0.080 & {\cellcolor[HTML]{F19E7D}} \color[HTML]{000000} 196.544 \\
MODR-OLS-LRA-IND & {\cellcolor[HTML]{DDEBF2}} \color[HTML]{000000} 39.197 & {\cellcolor[HTML]{87BEDA}} \color[HTML]{000000} 20.292 & {\cellcolor[HTML]{DDEBF2}} \color[HTML]{000000} 15.710 & {\cellcolor[HTML]{67001F}} \color[HTML]{F1F1F1} 5.189 & {\cellcolor[HTML]{67001F}} \color[HTML]{F1F1F1} $>5,000$ & {\cellcolor[HTML]{F7F6F6}} \color[HTML]{000000} 93.714 & {\cellcolor[HTML]{EC9374}} \color[HTML]{F1F1F1} 175.884 & {\cellcolor[HTML]{F9C6AC}} \color[HTML]{000000} 95.602 & {\cellcolor[HTML]{195696}} \color[HTML]{F1F1F1} -0.049 & {\cellcolor[HTML]{EC9374}} \color[HTML]{F1F1F1} 201.897 \\
MODR-OLS-LRA-$\Sigma$ & {\cellcolor[HTML]{FAE9DF}} \color[HTML]{000000} 41.642 & {\cellcolor[HTML]{D7E8F1}} \color[HTML]{000000} 21.292 & {\cellcolor[HTML]{FAE9DF}} \color[HTML]{000000} 16.520 & {\cellcolor[HTML]{ACD2E5}} \color[HTML]{000000} 3.940 & {\cellcolor[HTML]{84BCD9}} \color[HTML]{000000} 1161.734 & {\cellcolor[HTML]{F8F1ED}} \color[HTML]{000000} 94.398 & {\cellcolor[HTML]{F09C7B}} \color[HTML]{000000} 174.901 & {\cellcolor[HTML]{FDDDCB}} \color[HTML]{000000} 93.993 & {\cellcolor[HTML]{2369AD}} \color[HTML]{F1F1F1} -0.072 & {\cellcolor[HTML]{F8BFA4}} \color[HTML]{000000} 176.183 \\
MODR-LASSO-LRA-$\Sigma$ & {\cellcolor[HTML]{E6EFF4}} \color[HTML]{000000} 39.664 & {\cellcolor[HTML]{A0CCE2}} \color[HTML]{000000} 20.550 & {\cellcolor[HTML]{FBE3D4}} \color[HTML]{000000} 16.699 & {\cellcolor[HTML]{529DC8}} \color[HTML]{F1F1F1} 3.741 & {\cellcolor[HTML]{1A5899}} \color[HTML]{F1F1F1} 764.797 & {\cellcolor[HTML]{F9EEE7}} \color[HTML]{000000} 94.777 & {\cellcolor[HTML]{67001F}} \color[HTML]{F1F1F1} 203.316 & {\cellcolor[HTML]{FDD9C4}} \color[HTML]{000000} 94.322 & {\cellcolor[HTML]{4291C2}} \color[HTML]{F1F1F1} -0.131 & {\cellcolor[HTML]{FAE7DC}} \color[HTML]{000000} 149.184 \\
\bottomrule
\end{tabular}
} \\
    \vspace{0.25cm} 
    \centering \small Panel B: First sub-sample (2018-12-27 to 2020-12-31, $n = 736$)
    \resizebox{\textwidth}{!}{\begin{tabular}{R{6cm}L{1.9cm}L{1.9cm}L{1.9cm}L{1.9cm}L{1.9cm}L{1.9cm}L{1.9cm}L{1.9cm}L{1.9cm}L{1.9cm}}
\toprule
 & RMSE & MAE & CRPS & $\text{VS}_{p=0.5}$ & $\text{VS}_{p=1}$ & ES & DSS & LS & $\text{MC}_{0.95}$ & $\text{JPBW}_{0.95}$ \\
\midrule
LEAR-N(0, $\sigma$) & {\cellcolor[HTML]{053061}} \color[HTML]{F1F1F1} 7.568 & {\cellcolor[HTML]{F0F4F6}} \color[HTML]{000000} 4.790 & {\cellcolor[HTML]{E7F0F4}} \color[HTML]{000000} 3.588 & {\cellcolor[HTML]{DF765E}} \color[HTML]{F1F1F1} 1.163 & {\cellcolor[HTML]{F9EEE7}} \color[HTML]{000000} 59.095 & {\cellcolor[HTML]{FCE2D2}} \color[HTML]{000000} 21.940 & {\cellcolor[HTML]{D05548}} \color[HTML]{F1F1F1} 120.417 & {\cellcolor[HTML]{67001F}} \color[HTML]{F1F1F1} 82.117 & {\cellcolor[HTML]{246AAE}} \color[HTML]{F1F1F1} -0.063 & {\cellcolor[HTML]{FCDFCF}} \color[HTML]{000000} 39.127 \\
LEAR-N(0, $\Sigma$) & {\cellcolor[HTML]{053061}} \color[HTML]{F1F1F1} 7.568 & {\cellcolor[HTML]{F0F4F6}} \color[HTML]{000000} 4.791 & {\cellcolor[HTML]{E7F0F4}} \color[HTML]{000000} 3.588 & {\cellcolor[HTML]{2065AB}} \color[HTML]{F1F1F1} 0.912 & {\cellcolor[HTML]{053061}} \color[HTML]{F1F1F1} \bfseries 52.607 & {\cellcolor[HTML]{ACD2E5}} \color[HTML]{000000} 21.368 & {\cellcolor[HTML]{4695C4}} \color[HTML]{F1F1F1} 85.142 & {\cellcolor[HTML]{CFE4EF}} \color[HTML]{000000} 64.467 & {\cellcolor[HTML]{307AB6}} \color[HTML]{F1F1F1} -0.080 & {\cellcolor[HTML]{F8F1ED}} \color[HTML]{000000} 36.139 \\
LEAR-CP & {\cellcolor[HTML]{3F8EC0}} \color[HTML]{F1F1F1} 7.751 & {\cellcolor[HTML]{D35A4A}} \color[HTML]{F1F1F1} 4.959 & {\cellcolor[HTML]{67001F}} \color[HTML]{F1F1F1} 3.831 & {\cellcolor[HTML]{FFFFFF}} \color[HTML]{000000} \color{white}  & {\cellcolor[HTML]{FFFFFF}} \color[HTML]{000000} \color{white}  & {\cellcolor[HTML]{FFFFFF}} \color[HTML]{000000} \color{white}  & {\cellcolor[HTML]{FFFFFF}} \color[HTML]{000000} \color{white}  & {\cellcolor[HTML]{FFFFFF}} \color[HTML]{000000} \color{white}  & {\cellcolor[HTML]{67001F}} \color[HTML]{F1F1F1} -0.525 & {\cellcolor[HTML]{053061}} \color[HTML]{F1F1F1} \bfseries 18.469 \\
LEAR-GARCH & {\cellcolor[HTML]{053061}} \color[HTML]{F1F1F1} \bfseries 7.567 & {\cellcolor[HTML]{F0F4F6}} \color[HTML]{000000} 4.791 & {\cellcolor[HTML]{F3F5F6}} \color[HTML]{000000} 3.601 & {\cellcolor[HTML]{67001F}} \color[HTML]{F1F1F1} 1.258 & {\cellcolor[HTML]{67001F}} \color[HTML]{F1F1F1} 65.391 & {\cellcolor[HTML]{F9C4A9}} \color[HTML]{000000} 22.099 & {\cellcolor[HTML]{DA6853}} \color[HTML]{F1F1F1} 118.522 & {\cellcolor[HTML]{760521}} \color[HTML]{F1F1F1} 81.369 & {\cellcolor[HTML]{1C5C9F}} \color[HTML]{F1F1F1} -0.049 & {\cellcolor[HTML]{F9C6AC}} \color[HTML]{000000} 42.002 \\
ODR-IND & {\cellcolor[HTML]{8AC0DB}} \color[HTML]{000000} 7.849 & {\cellcolor[HTML]{073467}} \color[HTML]{F1F1F1} 4.552 & {\cellcolor[HTML]{09386D}} \color[HTML]{F1F1F1} 3.399 & {\cellcolor[HTML]{A2CDE3}} \color[HTML]{000000} 0.994 & {\cellcolor[HTML]{93C6DE}} \color[HTML]{000000} 56.194 & {\cellcolor[HTML]{4393C3}} \color[HTML]{F1F1F1} 21.024 & {\cellcolor[HTML]{F4A683}} \color[HTML]{000000} 112.649 & {\cellcolor[HTML]{E58368}} \color[HTML]{F1F1F1} 74.195 & {\cellcolor[HTML]{053061}} \color[HTML]{F1F1F1} -0.006 & {\cellcolor[HTML]{DB6B55}} \color[HTML]{F1F1F1} 50.476 \\
ODR-GC & {\cellcolor[HTML]{78B4D5}} \color[HTML]{000000} 7.828 & {\cellcolor[HTML]{063264}} \color[HTML]{F1F1F1} 4.551 & {\cellcolor[HTML]{073467}} \color[HTML]{F1F1F1} 3.394 & {\cellcolor[HTML]{053061}} \color[HTML]{F1F1F1} 0.877 & {\cellcolor[HTML]{1E61A5}} \color[HTML]{F1F1F1} 53.687 & {\cellcolor[HTML]{073467}} \color[HTML]{F1F1F1} 20.579 & {\cellcolor[HTML]{073467}} \color[HTML]{F1F1F1} 76.162 & {\cellcolor[HTML]{195696}} \color[HTML]{F1F1F1} 56.599 & {\cellcolor[HTML]{0F437B}} \color[HTML]{F1F1F1} -0.025 & {\cellcolor[HTML]{F8BB9E}} \color[HTML]{000000} 43.055 \\
ODR-SPGC & {\cellcolor[HTML]{71B0D3}} \color[HTML]{F1F1F1} 7.819 & {\cellcolor[HTML]{053061}} \color[HTML]{F1F1F1} \bfseries 4.548 & {\cellcolor[HTML]{053061}} \color[HTML]{F1F1F1} \bfseries 3.390 & {\cellcolor[HTML]{053061}} \color[HTML]{F1F1F1} \bfseries 0.877 & {\cellcolor[HTML]{1F63A8}} \color[HTML]{F1F1F1} 53.706 & {\cellcolor[HTML]{053061}} \color[HTML]{F1F1F1} \bfseries 20.557 & {\cellcolor[HTML]{053061}} \color[HTML]{F1F1F1} \bfseries 75.768 & {\cellcolor[HTML]{185493}} \color[HTML]{F1F1F1} 56.561 & {\cellcolor[HTML]{0D3F76}} \color[HTML]{F1F1F1} -0.021 & {\cellcolor[HTML]{F8BB9E}} \color[HTML]{000000} 43.102 \\
MODR-OLS-CD-IND & {\cellcolor[HTML]{EDF2F5}} \color[HTML]{000000} 8.034 & {\cellcolor[HTML]{5FA5CD}} \color[HTML]{F1F1F1} 4.664 & {\cellcolor[HTML]{CAE1EE}} \color[HTML]{000000} 3.557 & {\cellcolor[HTML]{4C99C6}} \color[HTML]{F1F1F1} 0.952 & {\cellcolor[HTML]{A7D0E4}} \color[HTML]{000000} 56.556 & {\cellcolor[HTML]{F8F4F2}} \color[HTML]{000000} 21.777 & {\cellcolor[HTML]{67001F}} \color[HTML]{F1F1F1} 133.695 & {\cellcolor[HTML]{F8BDA1}} \color[HTML]{000000} 71.549 & {\cellcolor[HTML]{2369AD}} \color[HTML]{F1F1F1} -0.061 & {\cellcolor[HTML]{F5AA89}} \color[HTML]{000000} 44.810 \\
MODR-OLS-CD-$\Sigma$ & {\cellcolor[HTML]{F3F5F6}} \color[HTML]{000000} 8.050 & {\cellcolor[HTML]{A5CEE3}} \color[HTML]{000000} 4.713 & {\cellcolor[HTML]{ACD2E5}} \color[HTML]{000000} 3.536 & {\cellcolor[HTML]{246AAE}} \color[HTML]{F1F1F1} 0.916 & {\cellcolor[HTML]{65A9CF}} \color[HTML]{F1F1F1} 55.459 & {\cellcolor[HTML]{AED3E6}} \color[HTML]{000000} 21.373 & {\cellcolor[HTML]{408FC1}} \color[HTML]{F1F1F1} 84.530 & {\cellcolor[HTML]{124984}} \color[HTML]{F1F1F1} 56.039 & {\cellcolor[HTML]{1A5899}} \color[HTML]{F1F1F1} -0.045 & {\cellcolor[HTML]{E17860}} \color[HTML]{F1F1F1} 49.244 \\
MODR-LASSO-CD-$\Sigma$ & {\cellcolor[HTML]{67001F}} \color[HTML]{F1F1F1} 8.579 & {\cellcolor[HTML]{67001F}} \color[HTML]{F1F1F1} 5.062 & {\cellcolor[HTML]{6D0220}} \color[HTML]{F1F1F1} 3.827 & {\cellcolor[HTML]{C2DDEC}} \color[HTML]{000000} 1.012 & {\cellcolor[HTML]{F2A17F}} \color[HTML]{000000} 61.332 & {\cellcolor[HTML]{67001F}} \color[HTML]{F1F1F1} 23.007 & {\cellcolor[HTML]{75B2D4}} \color[HTML]{F1F1F1} 88.095 & {\cellcolor[HTML]{114781}} \color[HTML]{F1F1F1} 55.962 & {\cellcolor[HTML]{053061}} \color[HTML]{F1F1F1} \bfseries 0.005 & {\cellcolor[HTML]{67001F}} \color[HTML]{F1F1F1} 66.428 \\
MODR-OLS-MCD-IND & {\cellcolor[HTML]{F6F7F7}} \color[HTML]{000000} 8.056 & {\cellcolor[HTML]{569FC9}} \color[HTML]{F1F1F1} 4.659 & {\cellcolor[HTML]{81BAD8}} \color[HTML]{000000} 3.509 & {\cellcolor[HTML]{5FA5CD}} \color[HTML]{F1F1F1} 0.961 & {\cellcolor[HTML]{9DCBE1}} \color[HTML]{000000} 56.386 & {\cellcolor[HTML]{D8E9F1}} \color[HTML]{000000} 21.557 & {\cellcolor[HTML]{C6413E}} \color[HTML]{F1F1F1} 122.110 & {\cellcolor[HTML]{FAC8AF}} \color[HTML]{000000} 70.944 & {\cellcolor[HTML]{0A3B70}} \color[HTML]{F1F1F1} -0.017 & {\cellcolor[HTML]{DF765E}} \color[HTML]{F1F1F1} 49.627 \\
MODR-OLS-MCD-$\Sigma$ & {\cellcolor[HTML]{F7F6F6}} \color[HTML]{000000} 8.059 & {\cellcolor[HTML]{93C6DE}} \color[HTML]{000000} 4.699 & {\cellcolor[HTML]{7BB6D6}} \color[HTML]{000000} 3.505 & {\cellcolor[HTML]{185493}} \color[HTML]{F1F1F1} 0.900 & {\cellcolor[HTML]{3C8ABE}} \color[HTML]{F1F1F1} 54.724 & {\cellcolor[HTML]{96C7DF}} \color[HTML]{000000} 21.277 & {\cellcolor[HTML]{2369AD}} \color[HTML]{F1F1F1} 80.581 & {\cellcolor[HTML]{08366A}} \color[HTML]{F1F1F1} 55.291 & {\cellcolor[HTML]{144E8A}} \color[HTML]{F1F1F1} -0.036 & {\cellcolor[HTML]{F2A17F}} \color[HTML]{000000} 45.742 \\
MODR-LASSO-MCD-$\Sigma$ & {\cellcolor[HTML]{D25849}} \color[HTML]{F1F1F1} 8.380 & {\cellcolor[HTML]{F3A481}} \color[HTML]{000000} 4.903 & {\cellcolor[HTML]{F7B99C}} \color[HTML]{000000} 3.678 & {\cellcolor[HTML]{68ABD0}} \color[HTML]{F1F1F1} 0.965 & {\cellcolor[HTML]{FBD0B9}} \color[HTML]{000000} 60.209 & {\cellcolor[HTML]{EE9677}} \color[HTML]{000000} 22.292 & {\cellcolor[HTML]{15508D}} \color[HTML]{F1F1F1} 78.513 & {\cellcolor[HTML]{053061}} \color[HTML]{F1F1F1} \bfseries 54.989 & {\cellcolor[HTML]{2369AD}} \color[HTML]{F1F1F1} -0.061 & {\cellcolor[HTML]{FAE8DE}} \color[HTML]{000000} 37.680 \\
MODR-OLS-LRA-IND & {\cellcolor[HTML]{F2F5F6}} \color[HTML]{000000} 8.047 & {\cellcolor[HTML]{5CA3CB}} \color[HTML]{F1F1F1} 4.662 & {\cellcolor[HTML]{D2E6F0}} \color[HTML]{000000} 3.564 & {\cellcolor[HTML]{3C8ABE}} \color[HTML]{F1F1F1} 0.941 & {\cellcolor[HTML]{BBDAEA}} \color[HTML]{000000} 56.963 & {\cellcolor[HTML]{F9EEE7}} \color[HTML]{000000} 21.830 & {\cellcolor[HTML]{B1182B}} \color[HTML]{F1F1F1} 126.458 & {\cellcolor[HTML]{F9C4A9}} \color[HTML]{000000} 71.156 & {\cellcolor[HTML]{185493}} \color[HTML]{F1F1F1} -0.041 & {\cellcolor[HTML]{F5A886}} \color[HTML]{000000} 44.971 \\
MODR-OLS-LRA-$\Sigma$ & {\cellcolor[HTML]{F6F7F7}} \color[HTML]{000000} 8.057 & {\cellcolor[HTML]{78B4D5}} \color[HTML]{000000} 4.681 & {\cellcolor[HTML]{D4E6F1}} \color[HTML]{000000} 3.566 & {\cellcolor[HTML]{4C99C6}} \color[HTML]{F1F1F1} 0.952 & {\cellcolor[HTML]{AED3E6}} \color[HTML]{000000} 56.712 & {\cellcolor[HTML]{F9EDE5}} \color[HTML]{000000} 21.840 & {\cellcolor[HTML]{D05548}} \color[HTML]{F1F1F1} 120.469 & {\cellcolor[HTML]{FCDFCF}} \color[HTML]{000000} 69.491 & {\cellcolor[HTML]{144E8A}} \color[HTML]{F1F1F1} -0.034 & {\cellcolor[HTML]{F19E7D}} \color[HTML]{000000} 45.885 \\
MODR-LASSO-LRA-$\Sigma$ & {\cellcolor[HTML]{F8F2EF}} \color[HTML]{000000} 8.078 & {\cellcolor[HTML]{4695C4}} \color[HTML]{F1F1F1} 4.649 & {\cellcolor[HTML]{93C6DE}} \color[HTML]{000000} 3.519 & {\cellcolor[HTML]{A2CDE3}} \color[HTML]{000000} 0.993 & {\cellcolor[HTML]{E6EFF4}} \color[HTML]{000000} 58.064 & {\cellcolor[HTML]{DEEBF2}} \color[HTML]{000000} 21.594 & {\cellcolor[HTML]{D35A4A}} \color[HTML]{F1F1F1} 119.889 & {\cellcolor[HTML]{FDDDCB}} \color[HTML]{000000} 69.719 & {\cellcolor[HTML]{053061}} \color[HTML]{F1F1F1} -0.006 & {\cellcolor[HTML]{C94741}} \color[HTML]{F1F1F1} 53.847 \\
\bottomrule
\end{tabular}
} \\
    \vspace{0.25cm}
    \centering \small Panel C: Second sub-sample (2021-01-01 to 2023-12-31, $n = 1095$)
    \resizebox{\textwidth}{!}{\begin{tabular}{R{6cm}L{1.9cm}L{1.9cm}L{1.9cm}L{1.9cm}L{1.9cm}L{1.9cm}L{1.9cm}L{1.9cm}L{1.9cm}L{1.9cm}}
\toprule
 & RMSE & MAE & CRPS & $\text{VS}_{p=0.5}$ & $\text{VS}_{p=1}$ & ES & DSS & LS & $\text{MC}_{0.95}$ & $\text{JPBW}_{0.95}$ \\
\midrule
LEAR-N(0, $\sigma$) & {\cellcolor[HTML]{053061}} \color[HTML]{F1F1F1} \bfseries 39.835 & {\cellcolor[HTML]{053061}} \color[HTML]{F1F1F1} \bfseries 27.237 & {\cellcolor[HTML]{2870B1}} \color[HTML]{F1F1F1} 20.459 & {\cellcolor[HTML]{EE9677}} \color[HTML]{000000} 7.093 & {\cellcolor[HTML]{4393C3}} \color[HTML]{F1F1F1} 1443.811 & {\cellcolor[HTML]{2369AD}} \color[HTML]{F1F1F1} 121.033 & {\cellcolor[HTML]{F8BB9E}} \color[HTML]{000000} 213.973 & {\cellcolor[HTML]{67001F}} \color[HTML]{F1F1F1} 128.804 & {\cellcolor[HTML]{65A9CF}} \color[HTML]{F1F1F1} -0.182 & {\cellcolor[HTML]{EFF3F5}} \color[HTML]{000000} 192.967 \\
LEAR-N(0, $\Sigma$) & {\cellcolor[HTML]{053061}} \color[HTML]{F1F1F1} 39.841 & {\cellcolor[HTML]{053061}} \color[HTML]{F1F1F1} 27.245 & {\cellcolor[HTML]{2870B1}} \color[HTML]{F1F1F1} 20.460 & {\cellcolor[HTML]{053061}} \color[HTML]{F1F1F1} \bfseries 5.065 & {\cellcolor[HTML]{053061}} \color[HTML]{F1F1F1} \bfseries 1059.817 & {\cellcolor[HTML]{0A3B70}} \color[HTML]{F1F1F1} 117.073 & {\cellcolor[HTML]{0E4179}} \color[HTML]{F1F1F1} 150.695 & {\cellcolor[HTML]{75B2D4}} \color[HTML]{F1F1F1} 97.225 & {\cellcolor[HTML]{87BEDA}} \color[HTML]{000000} -0.215 & {\cellcolor[HTML]{DAE9F2}} \color[HTML]{000000} 174.550 \\
LEAR-CP & {\cellcolor[HTML]{063264}} \color[HTML]{F1F1F1} 39.931 & {\cellcolor[HTML]{08366A}} \color[HTML]{F1F1F1} 27.391 & {\cellcolor[HTML]{569FC9}} \color[HTML]{F1F1F1} 21.507 & {\cellcolor[HTML]{FFFFFF}} \color[HTML]{000000} \color{white}  & {\cellcolor[HTML]{FFFFFF}} \color[HTML]{000000} \color{white}  & {\cellcolor[HTML]{FFFFFF}} \color[HTML]{000000} \color{white}  & {\cellcolor[HTML]{FFFFFF}} \color[HTML]{000000} \color{white}  & {\cellcolor[HTML]{FFFFFF}} \color[HTML]{000000} \color{white}  & {\cellcolor[HTML]{67001F}} \color[HTML]{F1F1F1} -0.742 & {\cellcolor[HTML]{053061}} \color[HTML]{F1F1F1} \bfseries 77.465 \\
LEAR-GARCH & {\cellcolor[HTML]{053061}} \color[HTML]{F1F1F1} 39.840 & {\cellcolor[HTML]{053061}} \color[HTML]{F1F1F1} 27.243 & {\cellcolor[HTML]{053061}} \color[HTML]{F1F1F1} \bfseries 19.304 & {\cellcolor[HTML]{FCD5BF}} \color[HTML]{000000} 6.739 & {\cellcolor[HTML]{3885BC}} \color[HTML]{F1F1F1} 1378.435 & {\cellcolor[HTML]{053061}} \color[HTML]{F1F1F1} \bfseries 116.095 & {\cellcolor[HTML]{D8E9F1}} \color[HTML]{000000} 187.118 & {\cellcolor[HTML]{EF9979}} \color[HTML]{000000} 115.516 & {\cellcolor[HTML]{114781}} \color[HTML]{F1F1F1} -0.034 & {\cellcolor[HTML]{F8F3F0}} \color[HTML]{000000} 206.636 \\
ODR-IND & {\cellcolor[HTML]{7EB8D7}} \color[HTML]{000000} 46.157 & {\cellcolor[HTML]{2F79B5}} \color[HTML]{F1F1F1} 28.885 & {\cellcolor[HTML]{3E8CBF}} \color[HTML]{F1F1F1} 21.092 & {\cellcolor[HTML]{E17860}} \color[HTML]{F1F1F1} 7.228 & {\cellcolor[HTML]{98C8E0}} \color[HTML]{000000} 1712.213 & {\cellcolor[HTML]{59A1CA}} \color[HTML]{F1F1F1} 127.073 & {\cellcolor[HTML]{ECF2F5}} \color[HTML]{000000} 192.533 & {\cellcolor[HTML]{EB9172}} \color[HTML]{F1F1F1} 116.117 & {\cellcolor[HTML]{0F437B}} \color[HTML]{F1F1F1} 0.029 & {\cellcolor[HTML]{67001F}} \color[HTML]{F1F1F1} 521.962 \\
ODR-GC & {\cellcolor[HTML]{3480B9}} \color[HTML]{F1F1F1} 43.365 & {\cellcolor[HTML]{144E8A}} \color[HTML]{F1F1F1} 27.894 & {\cellcolor[HTML]{2267AC}} \color[HTML]{F1F1F1} 20.295 & {\cellcolor[HTML]{2E77B5}} \color[HTML]{F1F1F1} 5.415 & {\cellcolor[HTML]{2A71B2}} \color[HTML]{F1F1F1} 1290.187 & {\cellcolor[HTML]{0C3D73}} \color[HTML]{F1F1F1} 117.218 & {\cellcolor[HTML]{266CAF}} \color[HTML]{F1F1F1} 157.607 & {\cellcolor[HTML]{2A71B2}} \color[HTML]{F1F1F1} 92.275 & {\cellcolor[HTML]{09386D}} \color[HTML]{F1F1F1} 0.014 & {\cellcolor[HTML]{960F27}} \color[HTML]{F1F1F1} 462.839 \\
ODR-SPGC & {\cellcolor[HTML]{3681BA}} \color[HTML]{F1F1F1} 43.450 & {\cellcolor[HTML]{15508D}} \color[HTML]{F1F1F1} 27.915 & {\cellcolor[HTML]{2369AD}} \color[HTML]{F1F1F1} 20.309 & {\cellcolor[HTML]{2F79B5}} \color[HTML]{F1F1F1} 5.425 & {\cellcolor[HTML]{2A71B2}} \color[HTML]{F1F1F1} 1289.995 & {\cellcolor[HTML]{0C3D73}} \color[HTML]{F1F1F1} 117.308 & {\cellcolor[HTML]{246AAE}} \color[HTML]{F1F1F1} 157.499 & {\cellcolor[HTML]{2B73B3}} \color[HTML]{F1F1F1} 92.349 & {\cellcolor[HTML]{0C3D73}} \color[HTML]{F1F1F1} 0.021 & {\cellcolor[HTML]{960F27}} \color[HTML]{F1F1F1} 462.912 \\
MODR-OLS-CD-IND & {\cellcolor[HTML]{D8E9F1}} \color[HTML]{000000} 49.867 & {\cellcolor[HTML]{8AC0DB}} \color[HTML]{000000} 30.654 & {\cellcolor[HTML]{D1E5F0}} \color[HTML]{000000} 23.377 & {\cellcolor[HTML]{68ABD0}} \color[HTML]{F1F1F1} 5.693 & {\cellcolor[HTML]{2369AD}} \color[HTML]{F1F1F1} 1252.714 & {\cellcolor[HTML]{BDDBEA}} \color[HTML]{000000} 134.201 & {\cellcolor[HTML]{FAE9DF}} \color[HTML]{000000} 201.244 & {\cellcolor[HTML]{FBCEB7}} \color[HTML]{000000} 111.508 & {\cellcolor[HTML]{114781}} \color[HTML]{F1F1F1} -0.035 & {\cellcolor[HTML]{F5AA89}} \color[HTML]{000000} 290.222 \\
MODR-OLS-CD-$\Sigma$ & {\cellcolor[HTML]{DE735C}} \color[HTML]{F1F1F1} 60.081 & {\cellcolor[HTML]{F5A886}} \color[HTML]{000000} 36.115 & {\cellcolor[HTML]{F5AA89}} \color[HTML]{000000} 26.825 & {\cellcolor[HTML]{59A1CA}} \color[HTML]{F1F1F1} 5.650 & {\cellcolor[HTML]{2B73B3}} \color[HTML]{F1F1F1} 1297.036 & {\cellcolor[HTML]{F9C4A9}} \color[HTML]{000000} 149.418 & {\cellcolor[HTML]{1A5899}} \color[HTML]{F1F1F1} 154.471 & {\cellcolor[HTML]{114781}} \color[HTML]{F1F1F1} 89.439 & {\cellcolor[HTML]{053061}} \color[HTML]{F1F1F1} \bfseries 0.002 & {\cellcolor[HTML]{B41C2D}} \color[HTML]{F1F1F1} 424.979 \\
MODR-LASSO-CD-$\Sigma$ & {\cellcolor[HTML]{67001F}} \color[HTML]{F1F1F1} 67.902 & {\cellcolor[HTML]{67001F}} \color[HTML]{F1F1F1} 40.882 & {\cellcolor[HTML]{67001F}} \color[HTML]{F1F1F1} 31.029 & {\cellcolor[HTML]{D2E6F0}} \color[HTML]{000000} 6.126 & {\cellcolor[HTML]{3B88BE}} \color[HTML]{F1F1F1} 1397.169 & {\cellcolor[HTML]{67001F}} \color[HTML]{F1F1F1} 171.775 & {\cellcolor[HTML]{195696}} \color[HTML]{F1F1F1} 154.108 & {\cellcolor[HTML]{0C3D73}} \color[HTML]{F1F1F1} 88.704 & {\cellcolor[HTML]{134C87}} \color[HTML]{F1F1F1} -0.041 & {\cellcolor[HTML]{D55D4C}} \color[HTML]{F1F1F1} 358.176 \\
MODR-OLS-MCD-IND & {\cellcolor[HTML]{D8E9F1}} \color[HTML]{000000} 49.915 & {\cellcolor[HTML]{87BEDA}} \color[HTML]{000000} 30.639 & {\cellcolor[HTML]{D1E5F0}} \color[HTML]{000000} 23.357 & {\cellcolor[HTML]{8AC0DB}} \color[HTML]{000000} 5.810 & {\cellcolor[HTML]{2A71B2}} \color[HTML]{F1F1F1} 1286.812 & {\cellcolor[HTML]{BBDAEA}} \color[HTML]{000000} 133.948 & {\cellcolor[HTML]{F9C6AC}} \color[HTML]{000000} 211.612 & {\cellcolor[HTML]{FBCCB4}} \color[HTML]{000000} 111.628 & {\cellcolor[HTML]{15508D}} \color[HTML]{F1F1F1} -0.045 & {\cellcolor[HTML]{F3A481}} \color[HTML]{000000} 295.708 \\
MODR-OLS-MCD-$\Sigma$ & {\cellcolor[HTML]{E37E64}} \color[HTML]{F1F1F1} 59.700 & {\cellcolor[HTML]{F6AF8E}} \color[HTML]{000000} 35.929 & {\cellcolor[HTML]{F7B799}} \color[HTML]{000000} 26.497 & {\cellcolor[HTML]{2E77B5}} \color[HTML]{F1F1F1} 5.407 & {\cellcolor[HTML]{1D5FA2}} \color[HTML]{F1F1F1} 1213.906 & {\cellcolor[HTML]{FCD5BF}} \color[HTML]{000000} 147.461 & {\cellcolor[HTML]{073467}} \color[HTML]{F1F1F1} 148.778 & {\cellcolor[HTML]{063264}} \color[HTML]{F1F1F1} 88.092 & {\cellcolor[HTML]{124984}} \color[HTML]{F1F1F1} -0.038 & {\cellcolor[HTML]{D35A4A}} \color[HTML]{F1F1F1} 361.243 \\
MODR-LASSO-MCD-$\Sigma$ & {\cellcolor[HTML]{B3192C}} \color[HTML]{F1F1F1} 64.334 & {\cellcolor[HTML]{BF3338}} \color[HTML]{F1F1F1} 38.639 & {\cellcolor[HTML]{BD2D35}} \color[HTML]{F1F1F1} 29.170 & {\cellcolor[HTML]{8DC2DC}} \color[HTML]{000000} 5.827 & {\cellcolor[HTML]{2C75B4}} \color[HTML]{F1F1F1} 1305.452 & {\cellcolor[HTML]{C43B3C}} \color[HTML]{F1F1F1} 161.906 & {\cellcolor[HTML]{053061}} \color[HTML]{F1F1F1} \bfseries 147.824 & {\cellcolor[HTML]{053061}} \color[HTML]{F1F1F1} \bfseries 87.859 & {\cellcolor[HTML]{2870B1}} \color[HTML]{F1F1F1} -0.092 & {\cellcolor[HTML]{F09C7B}} \color[HTML]{000000} 303.324 \\
MODR-OLS-LRA-IND & {\cellcolor[HTML]{DEEBF2}} \color[HTML]{000000} 50.255 & {\cellcolor[HTML]{93C6DE}} \color[HTML]{000000} 30.798 & {\cellcolor[HTML]{E3EDF3}} \color[HTML]{000000} 23.874 & {\cellcolor[HTML]{67001F}} \color[HTML]{F1F1F1} 8.044 & {\cellcolor[HTML]{67001F}} \color[HTML]{F1F1F1} $>5,000$ & {\cellcolor[HTML]{F8F3F0}} \color[HTML]{000000} 142.030 & {\cellcolor[HTML]{FCD3BC}} \color[HTML]{000000} 209.106 & {\cellcolor[HTML]{FAC8AF}} \color[HTML]{000000} 112.033 & {\cellcolor[HTML]{195696}} \color[HTML]{F1F1F1} -0.054 & {\cellcolor[HTML]{EE9677}} \color[HTML]{000000} 307.374 \\
MODR-OLS-LRA-$\Sigma$ & {\cellcolor[HTML]{FAE9DF}} \color[HTML]{000000} 53.441 & {\cellcolor[HTML]{DDEBF2}} \color[HTML]{000000} 32.456 & {\cellcolor[HTML]{FBE6DA}} \color[HTML]{000000} 25.226 & {\cellcolor[HTML]{ACD2E5}} \color[HTML]{000000} 5.949 & {\cellcolor[HTML]{C2DDEC}} \color[HTML]{000000} 1904.470 & {\cellcolor[HTML]{F9EEE7}} \color[HTML]{000000} 143.167 & {\cellcolor[HTML]{FAC8AF}} \color[HTML]{000000} 211.488 & {\cellcolor[HTML]{FDDBC7}} \color[HTML]{000000} 110.463 & {\cellcolor[HTML]{2B73B3}} \color[HTML]{F1F1F1} -0.098 & {\cellcolor[HTML]{F9C4A9}} \color[HTML]{000000} 263.763 \\
MODR-LASSO-LRA-$\Sigma$ & {\cellcolor[HTML]{E6EFF4}} \color[HTML]{000000} 50.860 & {\cellcolor[HTML]{A9D1E5}} \color[HTML]{000000} 31.239 & {\cellcolor[HTML]{FCDECD}} \color[HTML]{000000} 25.558 & {\cellcolor[HTML]{4695C4}} \color[HTML]{F1F1F1} 5.588 & {\cellcolor[HTML]{2065AB}} \color[HTML]{F1F1F1} 1239.825 & {\cellcolor[HTML]{FAE9DF}} \color[HTML]{000000} 143.967 & {\cellcolor[HTML]{67001F}} \color[HTML]{F1F1F1} 259.391 & {\cellcolor[HTML]{FCD7C2}} \color[HTML]{000000} 110.858 & {\cellcolor[HTML]{87BEDA}} \color[HTML]{000000} -0.215 & {\cellcolor[HTML]{F9EFE9}} \color[HTML]{000000} 213.264 \\
\bottomrule
\end{tabular}
}
    \caption{Scoring Rules. The best score in each column is marked \textbf{bold}. Note that the $LEAR-\mathcal{N}(0, \sigma)$, the ODR-IND and the MODR-IND models do not model the dependence structure.}
    \end{center} \label{tab:results_scores}
\end{table}

The $p$-values of the DM-test \secondreview{(see Figure \ref{fig:dm_tests} in \ref{app:figures})} largely confirm the statistical significance of the aforementioned results. We note the strong performance of the Copula-based models for the ES and the statistically significant superior performance of the multivariate distributional regression for the DSS, LS, and the VS. \review{We note the disconnect between the results of the ES on the one hand and the VS, DSS and LS on the other hand, which is in line with the findings of the asymmetry of scoring rules  by \cite{buchweitz2025asymmetric} and the different behaviors observed by \cite{pinson2013discrimination, marcotte2023regions} and \cite{alexander2024evaluating}.}

\review{
    Looking at Figure \ref{fig:skill_scores}, we see that skill scores are also well correlated with the current market regime. Skill scores are defined as $$
    \operatorname{SS}_\text{model} = 1 - \frac{\operatorname{LS}_\text{model}}{\operatorname{LS}_\text{baseline}},
    $$ where we use the LEAR-N(0,$\Sigma$) model as baseline. \secondreview{The MODR-LASSO models achieve up to 10\% improvement to the baseline LEAR-N(0,$\Sigma$) and even provide small gains during the high-volatility period of 2022. Especially during this period, the MODR models outperform the univariate ODR-models.} Additionally, we see that the models assuming independence have an increasingly better performance in recent years. This can be interpreted as a sign that the dependence structure has both weakened and changed with the \secondreview{turbulent} market conditions.
}%

\review{
    A common theme for distributional regression models is the trade-off between accuracy in the point predictions and the distributional error metrics. \cite{marcjasz2023distributional} and \cite{hirsch2024online} note this observation in the univariate case, trading RMSE for CRPS. This behavior is rooted in the fact that observations with predicted high variance are weighted down in the estimation of the mean parameters, which leads to slightly worse point predictions, but better distributional forecasts. At the same time, these observations with high variance are especially costly in terms of the RMSE. We see that the behavior extends in the multivariate case as well, trading off marginal accuracy for somewhat better modeling of the dependence structure in otherwise equal models. So far, this issue has not been discussed for multivariate distributional regression and yet deserves further attention in future research \citep[see e.g.][\secondreview{who focus on the DSS and LS as evaluation metrics}]{gioia2022additive,muschinski2022cholesky}.
}%

\review{
    For the multivariate online distributional regression models, we see that the models using the Cholesky-based parameterizations \secondreview{(CD and MCD)} provide better performance. This is likely due to the fact that these parameterizations are closer to the natural, time-based structure of the (conditional) dependence than the LRA and therefore the path-based regularization allows to select a parsimonious model, which still reflects the shape of the dependence structure. \secondreview{We further discuss the scale matrix regularization in~\ref{sec:overfitting}. For regularization with respect to the feature space,} using LASSO yields slight improvements in the scoring rules . \secondreview{The rather marginal gain for aggregate scores might be explained by the fact that all fundamental variables are known to be relevant for electricity price formation, and hence the regularization does not remove many variables, but helps to stabilize the estimates in the online learning setting.} 
}%

\review{Additionally, we analyze the performance of the models for extreme price spikes and for days with large spreads during the day. These days are interesting from a risk-management respectively from a battery optimization perspective. We define a price spike as a day with a minimum or maximum price exceeding the 5\%{} resp. 95\%-quantile of all minimum resp. maximum prices. A large spread event is defined as the min-to-max spread exceeding the 90\%-quantile of all min-to-max spreads. Formally:  \begin{align*}
    \operatorname{Spike} &= 
        \{t \mid \min(\vec{y}_t) < \operatorname{Q}_{0.05}(\min(\vec{y}_t)) \lor \max(\vec{y}_t) > \operatorname{Q}_{0.05}(\max(\vec{y}_t))\}, \quad \text{and}\\
    \operatorname{Spread} &= \{t \mid \max(\vec{y}_t) - \min(\vec{y}_t) > \operatorname{Q}_{0.90}(\max(\vec{y}_t) - \min(\vec{y}_t))\}.
\end{align*}
The results for the CRPS, LS and the miscoverage are given in Table \ref{tab:results_spike}. For the CRPS, we see that the ordering of the models is similar to the overall results, and remains largely unchanged between spike and no-spike events. For the miscoverage, we see that the multivariate distributional regression models yield the best results for both regimes for large spread events, and the best results for the price spike events, while the Copula-based models yield the best results for the no-spike events. Overall, these results highlight the robustness of the distributional regression models also for extreme events. \thirdreview{The results for the ES, DSS and VS can be found in the online supplement and mirror roughly the results for the CRPS and LS. We mirror Figure~\ref{fig:calibration_bands} for both regimes in the Figures 2 and 3 in the online supplement, giving the miscoverage of the $50\%, 55\%, ..., 95\%$-joint prediction bands for the spike and large spread events.} \secondreview{The disaggregated analysis reveals undercoverage during spikes for almost all models for smaller nominal coverage levels, where the LEAR and the ODR-SPGC deliver robust performance, while for larger nominal coverage levels, the undercoverage is less severe.
}}

\begin{table}[htb!]
\begin{center}
    \resizebox{\textwidth}{!}{\begin{tabular}{lrrrrrrrrrrrrrrrrrrrrrrrr}
\toprule
 & \multicolumn{4}{c}{CRPS} & \multicolumn{4}{c}{LS} & \multicolumn{4}{c}{$\text{MC}_{0.95}$} \\
 & \multicolumn{2}{c}{Spread} & \multicolumn{2}{c}{Spike} & \multicolumn{2}{c}{Spread} & \multicolumn{2}{c}{Spike} & \multicolumn{2}{c}{Spread} & \multicolumn{2}{c}{Spike} \\
 & Low & High & Low & High & Low & High & Low & High & Low & High & Low & High \\
\midrule
LEAR-N(0, $\sigma$) & {\cellcolor[HTML]{C2DDEC}} \color[HTML]{000000} 10.99 (10) & {\cellcolor[HTML]{1D5FA2}} \color[HTML]{F1F1F1} 37.85 (02) & {\cellcolor[HTML]{7BB6D6}} \color[HTML]{000000} 11.96 (07) & {\cellcolor[HTML]{185493}} \color[HTML]{F1F1F1} 29.02 (03) & {\cellcolor[HTML]{67001F}} \color[HTML]{F1F1F1} 103.42 (15) & {\cellcolor[HTML]{67001F}} \color[HTML]{F1F1F1} 169.63 (15) & {\cellcolor[HTML]{67001F}} \color[HTML]{F1F1F1} 104.74 (15) & {\cellcolor[HTML]{67001F}} \color[HTML]{F1F1F1} 157.46 (15) & {\cellcolor[HTML]{3885BC}} \color[HTML]{F1F1F1} -0.11 (14) & {\cellcolor[HTML]{E4EEF4}} \color[HTML]{000000} -0.37 (13) & {\cellcolor[HTML]{337EB8}} \color[HTML]{F1F1F1} -0.10 (14) & {\cellcolor[HTML]{E7F0F4}} \color[HTML]{000000} -0.45 (12) \\
LEAR-N(0, $\Sigma$) & {\cellcolor[HTML]{C2DDEC}} \color[HTML]{000000} 10.99 (09) & {\cellcolor[HTML]{1D5FA2}} \color[HTML]{F1F1F1} 37.85 (03) & {\cellcolor[HTML]{7BB6D6}} \color[HTML]{000000} 11.96 (08) & {\cellcolor[HTML]{185493}} \color[HTML]{F1F1F1} 29.02 (02) & {\cellcolor[HTML]{9BC9E0}} \color[HTML]{000000} 79.69 (07) & {\cellcolor[HTML]{6EAED2}} \color[HTML]{F1F1F1} 123.39 (07) & {\cellcolor[HTML]{87BEDA}} \color[HTML]{000000} 80.07 (07) & {\cellcolor[HTML]{D4E6F1}} \color[HTML]{000000} 119.74 (07) & {\cellcolor[HTML]{4695C4}} \color[HTML]{F1F1F1} -0.13 (15) & {\cellcolor[HTML]{F9EDE5}} \color[HTML]{000000} -0.43 (14) & {\cellcolor[HTML]{3F8EC0}} \color[HTML]{F1F1F1} -0.12 (15) & {\cellcolor[HTML]{F9EBE3}} \color[HTML]{000000} -0.52 (14) \\
LEAR-CP & {\cellcolor[HTML]{FCD3BC}} \color[HTML]{000000} 11.75 (13) & {\cellcolor[HTML]{2065AB}} \color[HTML]{F1F1F1} 38.27 (04) & {\cellcolor[HTML]{E4EEF4}} \color[HTML]{000000} 12.64 (12) & {\cellcolor[HTML]{246AAE}} \color[HTML]{F1F1F1} 30.16 (04) & {\cellcolor[HTML]{FFFFFF}} \color[HTML]{000000}  (17) & {\cellcolor[HTML]{FFFFFF}} \color[HTML]{000000}  (17) & {\cellcolor[HTML]{FFFFFF}} \color[HTML]{000000}  (17) & {\cellcolor[HTML]{FFFFFF}} \color[HTML]{000000}  (17) & {\cellcolor[HTML]{67001F}} \color[HTML]{F1F1F1} -0.64 (16) & {\cellcolor[HTML]{67001F}} \color[HTML]{F1F1F1} -0.79 (16) & {\cellcolor[HTML]{67001F}} \color[HTML]{F1F1F1} -0.63 (16) & {\cellcolor[HTML]{67001F}} \color[HTML]{F1F1F1} -0.86 (16) \\
LEAR-GARCH & {\cellcolor[HTML]{59A1CA}} \color[HTML]{F1F1F1} 10.54 (06) & {\cellcolor[HTML]{053061}} \color[HTML]{F1F1F1} 35.11 (01) & {\cellcolor[HTML]{266CAF}} \color[HTML]{F1F1F1} 11.40 (04) & {\cellcolor[HTML]{053061}} \color[HTML]{F1F1F1} 27.27 (01) & {\cellcolor[HTML]{BF3338}} \color[HTML]{F1F1F1} 98.06 (14) & {\cellcolor[HTML]{EDF2F5}} \color[HTML]{000000} 135.37 (08) & {\cellcolor[HTML]{D7634F}} \color[HTML]{F1F1F1} 97.02 (14) & {\cellcolor[HTML]{D05548}} \color[HTML]{F1F1F1} 144.52 (14) & {\cellcolor[HTML]{124984}} \color[HTML]{F1F1F1} -0.03 (08) & {\cellcolor[HTML]{246AAE}} \color[HTML]{F1F1F1} -0.10 (05) & {\cellcolor[HTML]{0A3B70}} \color[HTML]{F1F1F1} -0.01 (05) & {\cellcolor[HTML]{5FA5CD}} \color[HTML]{F1F1F1} -0.28 (07) \\
ODR-IND & {\cellcolor[HTML]{144E8A}} \color[HTML]{F1F1F1} 10.03 (03) & {\cellcolor[HTML]{CFE4EF}} \color[HTML]{000000} 49.54 (07) & {\cellcolor[HTML]{124984}} \color[HTML]{F1F1F1} 11.16 (03) & {\cellcolor[HTML]{D4E6F1}} \color[HTML]{000000} 39.18 (07) & {\cellcolor[HTML]{E27B62}} \color[HTML]{F1F1F1} 94.39 (13) & {\cellcolor[HTML]{FDD9C4}} \color[HTML]{000000} 143.14 (14) & {\cellcolor[HTML]{E37E64}} \color[HTML]{F1F1F1} 95.57 (13) & {\cellcolor[HTML]{FBCEB7}} \color[HTML]{000000} 132.34 (13) & {\cellcolor[HTML]{09386D}} \color[HTML]{F1F1F1} 0.01 (04) & {\cellcolor[HTML]{0A3B70}} \color[HTML]{F1F1F1} 0.03 (03) & {\cellcolor[HTML]{114781}} \color[HTML]{F1F1F1} 0.03 (10) & {\cellcolor[HTML]{053061}} \color[HTML]{F1F1F1} -0.10 (01) \\
ODR-GC & {\cellcolor[HTML]{053061}} \color[HTML]{F1F1F1} 9.87 (01) & {\cellcolor[HTML]{9BC9E0}} \color[HTML]{000000} 46.18 (05) & {\cellcolor[HTML]{053061}} \color[HTML]{F1F1F1} 10.99 (01) & {\cellcolor[HTML]{98C8E0}} \color[HTML]{000000} 35.96 (05) & {\cellcolor[HTML]{276EB0}} \color[HTML]{F1F1F1} 73.87 (05) & {\cellcolor[HTML]{1C5C9F}} \color[HTML]{F1F1F1} 114.50 (05) & {\cellcolor[HTML]{276EB0}} \color[HTML]{F1F1F1} 75.17 (05) & {\cellcolor[HTML]{195696}} \color[HTML]{F1F1F1} 102.66 (06) & {\cellcolor[HTML]{063264}} \color[HTML]{F1F1F1} -0.00 (03) & {\cellcolor[HTML]{053061}} \color[HTML]{F1F1F1} 0.02 (01) & {\cellcolor[HTML]{0A3B70}} \color[HTML]{F1F1F1} 0.01 (06) & {\cellcolor[HTML]{144E8A}} \color[HTML]{F1F1F1} -0.15 (03) \\
ODR-SPGC & {\cellcolor[HTML]{053061}} \color[HTML]{F1F1F1} 9.88 (02) & {\cellcolor[HTML]{9DCBE1}} \color[HTML]{000000} 46.22 (06) & {\cellcolor[HTML]{053061}} \color[HTML]{F1F1F1} 10.99 (02) & {\cellcolor[HTML]{98C8E0}} \color[HTML]{000000} 36.01 (06) & {\cellcolor[HTML]{276EB0}} \color[HTML]{F1F1F1} 73.90 (06) & {\cellcolor[HTML]{1C5C9F}} \color[HTML]{F1F1F1} 114.55 (06) & {\cellcolor[HTML]{276EB0}} \color[HTML]{F1F1F1} 75.21 (06) & {\cellcolor[HTML]{195696}} \color[HTML]{F1F1F1} 102.65 (05) & {\cellcolor[HTML]{053061}} \color[HTML]{F1F1F1} 0.00 (01) & {\cellcolor[HTML]{0A3B70}} \color[HTML]{F1F1F1} 0.03 (03) & {\cellcolor[HTML]{0D3F76}} \color[HTML]{F1F1F1} 0.02 (09) & {\cellcolor[HTML]{10457E}} \color[HTML]{F1F1F1} -0.13 (02) \\
MODR-OLS-CD-IND & {\cellcolor[HTML]{4291C2}} \color[HTML]{F1F1F1} 10.45 (05) & {\cellcolor[HTML]{FBCEB7}} \color[HTML]{000000} 60.09 (09) & {\cellcolor[HTML]{4695C4}} \color[HTML]{F1F1F1} 11.71 (06) & {\cellcolor[HTML]{F9C2A7}} \color[HTML]{000000} 48.49 (09) & {\cellcolor[HTML]{F8BDA1}} \color[HTML]{000000} 90.59 (11) & {\cellcolor[HTML]{F9EDE5}} \color[HTML]{000000} 139.14 (10) & {\cellcolor[HTML]{F8BDA1}} \color[HTML]{000000} 91.93 (11) & {\cellcolor[HTML]{F9EEE7}} \color[HTML]{000000} 126.91 (11) & {\cellcolor[HTML]{134C87}} \color[HTML]{F1F1F1} -0.03 (09) & {\cellcolor[HTML]{3885BC}} \color[HTML]{F1F1F1} -0.15 (06) & {\cellcolor[HTML]{0A3B70}} \color[HTML]{F1F1F1} -0.01 (07) & {\cellcolor[HTML]{87BEDA}} \color[HTML]{000000} -0.32 (10) \\
MODR-OLS-CD-$\Sigma$ & {\cellcolor[HTML]{F8BDA1}} \color[HTML]{000000} 11.88 (14) & {\cellcolor[HTML]{E27B62}} \color[HTML]{F1F1F1} 67.75 (12) & {\cellcolor[HTML]{FCD3BC}} \color[HTML]{000000} 13.29 (14) & {\cellcolor[HTML]{DA6853}} \color[HTML]{F1F1F1} 54.82 (13) & {\cellcolor[HTML]{134C87}} \color[HTML]{F1F1F1} 72.06 (04) & {\cellcolor[HTML]{0C3D73}} \color[HTML]{F1F1F1} 111.61 (03) & {\cellcolor[HTML]{134C87}} \color[HTML]{F1F1F1} 73.26 (04) & {\cellcolor[HTML]{0D3F76}} \color[HTML]{F1F1F1} 100.64 (03) & {\cellcolor[HTML]{0A3B70}} \color[HTML]{F1F1F1} -0.02 (05) & {\cellcolor[HTML]{09386D}} \color[HTML]{F1F1F1} -0.03 (02) & {\cellcolor[HTML]{053061}} \color[HTML]{F1F1F1} 0.00 (01) & {\cellcolor[HTML]{2065AB}} \color[HTML]{F1F1F1} -0.18 (04) \\
MODR-LASSO-CD-$\Sigma$ & {\cellcolor[HTML]{67001F}} \color[HTML]{F1F1F1} 13.09 (16) & {\cellcolor[HTML]{67001F}} \color[HTML]{F1F1F1} 83.18 (16) & {\cellcolor[HTML]{67001F}} \color[HTML]{F1F1F1} 14.95 (16) & {\cellcolor[HTML]{67001F}} \color[HTML]{F1F1F1} 66.14 (16) & {\cellcolor[HTML]{0E4179}} \color[HTML]{F1F1F1} 71.49 (03) & {\cellcolor[HTML]{0E4179}} \color[HTML]{F1F1F1} 112.08 (04) & {\cellcolor[HTML]{0E4179}} \color[HTML]{F1F1F1} 72.72 (03) & {\cellcolor[HTML]{0E4179}} \color[HTML]{F1F1F1} 100.79 (04) & {\cellcolor[HTML]{053061}} \color[HTML]{F1F1F1} -0.00 (02) & {\cellcolor[HTML]{71B0D3}} \color[HTML]{F1F1F1} -0.22 (10) & {\cellcolor[HTML]{08366A}} \color[HTML]{F1F1F1} 0.01 (03) & {\cellcolor[HTML]{7BB6D6}} \color[HTML]{000000} -0.31 (08) \\
MODR-OLS-MCD-IND & {\cellcolor[HTML]{3F8EC0}} \color[HTML]{F1F1F1} 10.42 (04) & {\cellcolor[HTML]{FBD0B9}} \color[HTML]{000000} 60.06 (08) & {\cellcolor[HTML]{4393C3}} \color[HTML]{F1F1F1} 11.69 (05) & {\cellcolor[HTML]{F9C4A9}} \color[HTML]{000000} 48.35 (08) & {\cellcolor[HTML]{F9C2A7}} \color[HTML]{000000} 90.37 (10) & {\cellcolor[HTML]{F9EBE3}} \color[HTML]{000000} 139.46 (11) & {\cellcolor[HTML]{F8BFA4}} \color[HTML]{000000} 91.79 (10) & {\cellcolor[HTML]{F9F0EB}} \color[HTML]{000000} 126.46 (09) & {\cellcolor[HTML]{0C3D73}} \color[HTML]{F1F1F1} -0.02 (06) & {\cellcolor[HTML]{4F9BC7}} \color[HTML]{F1F1F1} -0.18 (09) & {\cellcolor[HTML]{073467}} \color[HTML]{F1F1F1} -0.01 (02) & {\cellcolor[HTML]{529DC8}} \color[HTML]{F1F1F1} -0.27 (06) \\
MODR-OLS-MCD-$\Sigma$ & {\cellcolor[HTML]{FAE8DE}} \color[HTML]{000000} 11.55 (12) & {\cellcolor[HTML]{DD7059}} \color[HTML]{F1F1F1} 68.64 (14) & {\cellcolor[HTML]{F9F0EB}} \color[HTML]{000000} 12.93 (13) & {\cellcolor[HTML]{D25849}} \color[HTML]{F1F1F1} 56.01 (14) & {\cellcolor[HTML]{08366A}} \color[HTML]{F1F1F1} 70.96 (02) & {\cellcolor[HTML]{053061}} \color[HTML]{F1F1F1} 110.49 (01) & {\cellcolor[HTML]{08366A}} \color[HTML]{F1F1F1} 72.19 (02) & {\cellcolor[HTML]{053061}} \color[HTML]{F1F1F1} 99.20 (01) & {\cellcolor[HTML]{0F437B}} \color[HTML]{F1F1F1} -0.02 (07) & {\cellcolor[HTML]{3A87BD}} \color[HTML]{F1F1F1} -0.15 (07) & {\cellcolor[HTML]{09386D}} \color[HTML]{F1F1F1} -0.01 (04) & {\cellcolor[HTML]{4997C5}} \color[HTML]{F1F1F1} -0.26 (05) \\
MODR-LASSO-MCD-$\Sigma$ & {\cellcolor[HTML]{DE735C}} \color[HTML]{F1F1F1} 12.28 (15) & {\cellcolor[HTML]{960F27}} \color[HTML]{F1F1F1} 78.74 (15) & {\cellcolor[HTML]{DA6853}} \color[HTML]{F1F1F1} 14.01 (15) & {\cellcolor[HTML]{900D26}} \color[HTML]{F1F1F1} 62.86 (15) & {\cellcolor[HTML]{053061}} \color[HTML]{F1F1F1} 70.62 (01) & {\cellcolor[HTML]{073467}} \color[HTML]{F1F1F1} 110.91 (02) & {\cellcolor[HTML]{053061}} \color[HTML]{F1F1F1} 71.85 (01) & {\cellcolor[HTML]{073467}} \color[HTML]{F1F1F1} 99.66 (02) & {\cellcolor[HTML]{1F63A8}} \color[HTML]{F1F1F1} -0.06 (12) & {\cellcolor[HTML]{8AC0DB}} \color[HTML]{000000} -0.24 (11) & {\cellcolor[HTML]{185493}} \color[HTML]{F1F1F1} -0.04 (12) & {\cellcolor[HTML]{D4E6F1}} \color[HTML]{000000} -0.41 (11) \\
MODR-OLS-LRA-IND & {\cellcolor[HTML]{93C6DE}} \color[HTML]{000000} 10.76 (07) & {\cellcolor[HTML]{FBCEB7}} \color[HTML]{000000} 60.27 (10) & {\cellcolor[HTML]{87BEDA}} \color[HTML]{000000} 12.01 (09) & {\cellcolor[HTML]{F8BDA1}} \color[HTML]{000000} 48.83 (10) & {\cellcolor[HTML]{F8BDA1}} \color[HTML]{000000} 90.66 (12) & {\cellcolor[HTML]{FAE8DE}} \color[HTML]{000000} 140.08 (13) & {\cellcolor[HTML]{F8BB9E}} \color[HTML]{000000} 92.10 (12) & {\cellcolor[HTML]{F9EEE7}} \color[HTML]{000000} 126.99 (12) & {\cellcolor[HTML]{134C87}} \color[HTML]{F1F1F1} -0.03 (10) & {\cellcolor[HTML]{4695C4}} \color[HTML]{F1F1F1} -0.17 (08) & {\cellcolor[HTML]{0D3F76}} \color[HTML]{F1F1F1} -0.02 (08) & {\cellcolor[HTML]{81BAD8}} \color[HTML]{000000} -0.31 (09) \\
MODR-OLS-LRA-$\Sigma$ & {\cellcolor[HTML]{BBDAEA}} \color[HTML]{000000} 10.95 (08) & {\cellcolor[HTML]{E8896C}} \color[HTML]{F1F1F1} 66.68 (11) & {\cellcolor[HTML]{B3D6E8}} \color[HTML]{000000} 12.26 (10) & {\cellcolor[HTML]{DB6B55}} \color[HTML]{F1F1F1} 54.65 (12) & {\cellcolor[HTML]{FCD7C2}} \color[HTML]{000000} 89.00 (08) & {\cellcolor[HTML]{F9EEE7}} \color[HTML]{000000} 138.95 (09) & {\cellcolor[HTML]{FCD5BF}} \color[HTML]{000000} 90.48 (08) & {\cellcolor[HTML]{F7F5F4}} \color[HTML]{000000} 125.42 (08) & {\cellcolor[HTML]{185493}} \color[HTML]{F1F1F1} -0.04 (11) & {\cellcolor[HTML]{D2E6F0}} \color[HTML]{000000} -0.33 (12) & {\cellcolor[HTML]{114781}} \color[HTML]{F1F1F1} -0.03 (11) & {\cellcolor[HTML]{E7F0F4}} \color[HTML]{000000} -0.45 (12) \\
MODR-LASSO-LRA-$\Sigma$ & {\cellcolor[HTML]{C5DFEC}} \color[HTML]{000000} 11.00 (11) & {\cellcolor[HTML]{E17860}} \color[HTML]{F1F1F1} 68.00 (13) & {\cellcolor[HTML]{D7E8F1}} \color[HTML]{000000} 12.51 (11) & {\cellcolor[HTML]{DD7059}} \color[HTML]{F1F1F1} 54.21 (11) & {\cellcolor[HTML]{FCD3BC}} \color[HTML]{000000} 89.29 (09) & {\cellcolor[HTML]{FAEAE1}} \color[HTML]{000000} 139.65 (12) & {\cellcolor[HTML]{FBD0B9}} \color[HTML]{000000} 90.72 (09) & {\cellcolor[HTML]{F9F0EB}} \color[HTML]{000000} 126.57 (10) & {\cellcolor[HTML]{2A71B2}} \color[HTML]{F1F1F1} -0.08 (13) & {\cellcolor[HTML]{EF9979}} \color[HTML]{000000} -0.57 (15) & {\cellcolor[HTML]{2B73B3}} \color[HTML]{F1F1F1} -0.08 (13) & {\cellcolor[HTML]{FCD7C2}} \color[HTML]{000000} -0.56 (15) \\
\bottomrule
\end{tabular}
}
    \caption{\review{Price spike and spread analysis for the CRPS and the miscoverage $\text{MC}_{0.95}$. A spike event is defined if the min/max price of a day exceeds the 5\%{} resp. 95\%-quantile of all min/max prices. A large spread event is defined as the min-to-max spread exceeding the 90\%-quantile of all min-to-max spreads. Numbers in brackets give the model ranking.}}\label{tab:results_spike}
    \end{center} 
\end{table}
\review{
    Before concluding the results, we want to emphasize some limitations of this study. \begin{itemize}
        \item We acknowledge that the proposed models yield a trade-off between accuracy in the marginal distributions and correct modeling of the dependence structure. This trade-off is reflected also in the disconnect between marginally dominated scoring rules (CRPS and ES) and the LS, DSS, and VS, and therefore, we yield somewhat ambiguous results.   
        \item While the results of the spike analysis are encouraging, the parametric assumption on the distributional family might be too restrictive in some cases and could be complemented with non-parametric approaches, or with multivariate extreme value models to capture the tail behavior. This could also help to improve the forecasting performance in the marginal distributions and hence CRPS and ES.
    \end{itemize} 
    Overall, our results highlight that neglecting the dependence structure by relying solely on marginal, univariate models yields subpar probabilistic forecasting performance. We note that for the truly multivariate approaches, using both, copula-based combinations of univariate models and the fully multivariate distributional regression yield statistically significant performance improvements. 
}

\FloatBarrier
\review{
    \subsection{Computation Time}
}\label{sec:results_comptime}

\begin{table}[htb]
    \begin{minipage}[t]{.7\linewidth}
    \vspace{0pt}
    \resizebox{\linewidth}{!}{%
        \begin{tabular}{lrrrrr}
\toprule
Model & Initial Fit & Avg. Update & Total Time & Est. Speedup \\
\midrule
LEAR-N(0, $\sigma$) & {\cellcolor[HTML]{F7FBFF}} \color[HTML]{000000} 1.23 & {\cellcolor[HTML]{E7F1FA}} \color[HTML]{000000} 0.06 & {\cellcolor[HTML]{F2F8FD}} \color[HTML]{000000} 103.80 & {\cellcolor[HTML]{B5D4E9}} \color[HTML]{000000} $\times$ 21 \\
LEAR-N(0, $\Sigma$) & {\cellcolor[HTML]{F7FBFF}} \color[HTML]{000000} 1.23 & {\cellcolor[HTML]{E7F1FA}} \color[HTML]{000000} 0.06 & {\cellcolor[HTML]{F2F8FD}} \color[HTML]{000000} 103.80 & {\cellcolor[HTML]{B5D4E9}} \color[HTML]{000000} $\times$ 21 \\
LEAR-CP & {\cellcolor[HTML]{F7FBFF}} \color[HTML]{000000} 1.23 & {\cellcolor[HTML]{E7F1FA}} \color[HTML]{000000} 0.06 & {\cellcolor[HTML]{F2F8FD}} \color[HTML]{000000} 103.80 & {\cellcolor[HTML]{B5D4E9}} \color[HTML]{000000} $\times$ 21 \\
LEAR-GARCH & {\cellcolor[HTML]{F3F8FE}} \color[HTML]{000000} 1.44 & {\cellcolor[HTML]{89BEDC}} \color[HTML]{000000} 0.27 & {\cellcolor[HTML]{A1CBE2}} \color[HTML]{000000} 501.31 & {\cellcolor[HTML]{F7FBFF}} \color[HTML]{000000} $\times$ 5 \\
ODR-IND & {\cellcolor[HTML]{6AAED6}} \color[HTML]{F1F1F1} 55.31 & {\cellcolor[HTML]{82BBDB}} \color[HTML]{000000} 0.30 & {\cellcolor[HTML]{92C4DE}} \color[HTML]{000000} 598.78 & {\cellcolor[HTML]{2272B6}} \color[HTML]{F1F1F1} $\times$ 169 \\
ODR-GC & {\cellcolor[HTML]{6AAED6}} \color[HTML]{F1F1F1} 55.32 & {\cellcolor[HTML]{82BBDB}} \color[HTML]{000000} 0.30 & {\cellcolor[HTML]{92C4DE}} \color[HTML]{000000} 599.61 & {\cellcolor[HTML]{2272B6}} \color[HTML]{F1F1F1} $\times$ 168 \\
ODR-SPGC & {\cellcolor[HTML]{6AAED6}} \color[HTML]{F1F1F1} 55.46 & {\cellcolor[HTML]{60A7D2}} \color[HTML]{F1F1F1} 0.45 & {\cellcolor[HTML]{71B1D7}} \color[HTML]{F1F1F1} 870.45 & {\cellcolor[HTML]{3787C0}} \color[HTML]{F1F1F1} $\times$ 116 \\
MODR-OLS-CD-IND & {\cellcolor[HTML]{7DB8DA}} \color[HTML]{000000} 39.08 & {\cellcolor[HTML]{D6E6F4}} \color[HTML]{000000} 0.08 & {\cellcolor[HTML]{D8E7F5}} \color[HTML]{000000} 190.40 & {\cellcolor[HTML]{08478D}} \color[HTML]{F1F1F1} $\times$ 375 \\
MODR-OLS-CD-$\Sigma$ & {\cellcolor[HTML]{4594C7}} \color[HTML]{F1F1F1} 131.51 & {\cellcolor[HTML]{A4CCE3}} \color[HTML]{000000} 0.20 & {\cellcolor[HTML]{A3CCE3}} \color[HTML]{000000} 494.41 & {\cellcolor[HTML]{083877}} \color[HTML]{F1F1F1} $\times$ 487 \\
MODR-LASSO-CD-$\Sigma$ & {\cellcolor[HTML]{0E59A2}} \color[HTML]{F1F1F1} 752.57 & {\cellcolor[HTML]{084A91}} \color[HTML]{F1F1F1} 2.41 & {\cellcolor[HTML]{09529D}} \color[HTML]{F1F1F1} 5156.91 & {\cellcolor[HTML]{0E59A2}} \color[HTML]{F1F1F1} $\times$ 267 \\
MODR-OLS-MCD-IND & {\cellcolor[HTML]{77B5D9}} \color[HTML]{000000} 43.97 & {\cellcolor[HTML]{E0ECF8}} \color[HTML]{000000} 0.07 & {\cellcolor[HTML]{DFEBF7}} \color[HTML]{000000} 163.73 & {\cellcolor[HTML]{083776}} \color[HTML]{F1F1F1} $\times$ 491 \\
MODR-OLS-MCD-$\Sigma$ & {\cellcolor[HTML]{4493C7}} \color[HTML]{F1F1F1} 134.60 & {\cellcolor[HTML]{B2D2E8}} \color[HTML]{000000} 0.16 & {\cellcolor[HTML]{ABD0E6}} \color[HTML]{000000} 434.25 & {\cellcolor[HTML]{08306B}} \color[HTML]{F1F1F1} $\times$ 567 \\
MODR-LASSO-MCD-$\Sigma$ & {\cellcolor[HTML]{084184}} \color[HTML]{F1F1F1} 1530.21 & {\cellcolor[HTML]{083370}} \color[HTML]{F1F1F1} 3.57 & {\cellcolor[HTML]{083877}} \color[HTML]{F1F1F1} 8066.77 & {\cellcolor[HTML]{084B93}} \color[HTML]{F1F1F1} $\times$ 347 \\
MODR-OLS-LRA-IND & {\cellcolor[HTML]{9DCAE1}} \color[HTML]{000000} 21.62 & {\cellcolor[HTML]{F7FBFF}} \color[HTML]{000000} 0.04 & {\cellcolor[HTML]{F7FBFF}} \color[HTML]{000000} 92.11 & {\cellcolor[HTML]{084082}} \color[HTML]{F1F1F1} $\times$ 429 \\
MODR-OLS-LRA-$\Sigma$ & {\cellcolor[HTML]{2777B8}} \color[HTML]{F1F1F1} 301.48 & {\cellcolor[HTML]{3A8AC2}} \color[HTML]{F1F1F1} 0.78 & {\cellcolor[HTML]{3F8FC5}} \color[HTML]{F1F1F1} 1735.92 & {\cellcolor[HTML]{08509B}} \color[HTML]{F1F1F1} $\times$ 317 \\
MODR-LASSO-LRA-$\Sigma$ & {\cellcolor[HTML]{08306B}} \color[HTML]{F1F1F1} 2469.66 & {\cellcolor[HTML]{08306B}} \color[HTML]{F1F1F1} 3.81 & {\cellcolor[HTML]{08306B}} \color[HTML]{F1F1F1} 9440.61 & {\cellcolor[HTML]{083979}} \color[HTML]{F1F1F1} $\times$ 478 \\
\bottomrule
\end{tabular}

    }%
    \end{minipage}
    \begin{minipage}[t]{.29\linewidth}
    \vspace{0pt}
    \includegraphics[width=\linewidth]{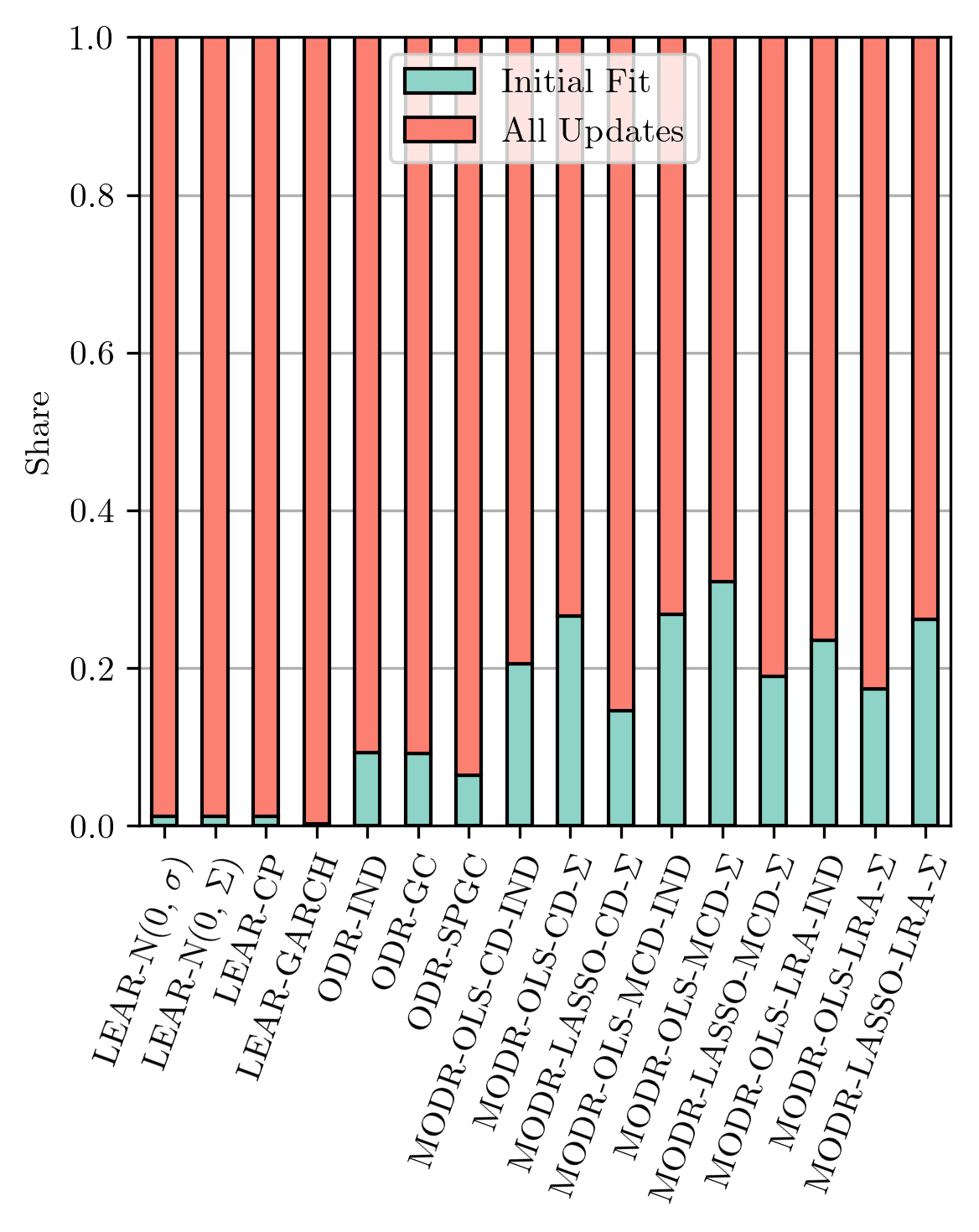}
    \end{minipage}
    \caption{Computation times. All timings are in seconds. The out-of-sample data for the forecasting study consists of 1831 days. We update the 24-dimensional distributional regression model on each day. All experiments are run on a standard laptop (Intel Core i7 (16 Threads, 4.9 GHz), 32GB RAM). Estimated speed-ups are calculated by taking $\text{Speedup} = (\text{Initial Fit} \times T) / \text{Total Time}$. \review{The figure on the right shows the share of the time spent on the initial fit and on the out-of-sample updating}.}
    \label{tab:computation_times}
\end{table}

Table \ref{tab:computation_times} gives computation times for all experiments. The initial fit for the multivariate distributional regression model takes a few minutes, the update algorithm can be executed in seconds. \review{Using online estimation methods, the experiments can be run in a few hours on a standard laptop.} An estimate for the benefit of online vs. repeated batch fitting can be achieved by multiplying the initial fit duration with 1831 days of out-of-sample and comparing this to the total time of the online study: $$
    \text{Speedup} = \frac{\text{Initial Fit} \times T}{\text{Total Time}}. 
$$ By this (albeit simple) measure, the online learning improves computation by a factor \review{of 80 to 600}. These estimates are in line with benefits reported in \cite{hirsch2024online} for the univariate online distributional regression case in \review{direct} comparison \review{between online estimation and repeated batch estimation}. 

\begin{figure}[htb]
    \centering
    \includegraphics[width=\linewidth]{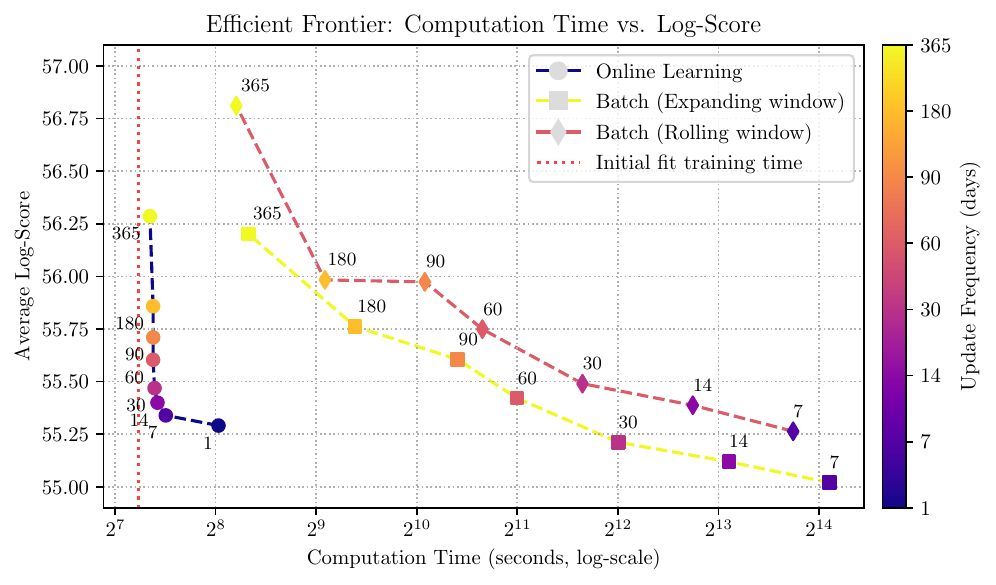}
    \caption{Efficient frontier of computation time against average log-score for the online multivariate distributional regression model for different update frequencies in the batch and online case. Note that the $x$-axis is in log-scale.}
    \label{fig:results_comptime_eff}
\end{figure}

\review{
    To further analyze the trade-off between computation time and accuracy, we take the OLS-estimated distributional regression model and estimate the model every 1, 7, 14, 30, 60, 180, 365 days online (using a mini-batch online update) and, vice versa, re-estimate the full model in a repeated batch fitting every 7, 14, ..., 365 days, using the first subsample of the test data.\footnote{Note that already estimating the batch model every 7 days takes more than 10 hours and hence we did not run the experiment for daily estimation in the batch case. \secondreview{We expect daily refitting scheme to deliver further improvements in the predictive accuracy, but at about seven times the computational time; the linear relationship between the computation time and update frequency is clearly visible in Figure \ref{fig:results_comptime_eff}.}} We compare the results for online learning, rolling and expanding window batch estimation in Figure~\ref{fig:results_comptime_eff}. \secondreview{For online and batch estimation, we see that higher update frequency improves forecasting accuracy, but also increases computation times.} Crucially, the ``efficient frontiers`` of both approaches never intersect. We see that for low computation time budgets, online learning approaches give strictly better results than repeated batch fitting approaches and only for large computation time budgets, expanding window batch estimation takes the lead. 
    A second important observation is that the online learning algorithm is almost constant in the time per update, since we always process a fixed amount of data points. On the other hand the number of data points in the expanding window batch estimation increases, which increases the computation time per fit. 
    Interestingly, the rolling window estimation scheme, which is often favored in the EPF literature, delivers worse predictive accuracy than an expanding window scheme. 
    These results further confirm the estimated speed-up in Table~\ref{tab:computation_times}.
}%
\review{
    Summarizing the results on the computational effort, our online update algorithm makes the approach practically viable for researchers and data scientists without access to specialized high-performance computation centers. \secondreview{Since one good model is usually preceeded by fitting a large amount of bad models, taking} into account the trade-off between accuracy and computational effort is of practical importance in many industrial settings, where analysts work under time pressure and data arrives at high velocity. Furthermore, time saved in raw computation can be used for better data exploration and feature engineering. 
}%

\FloatBarrier
\section{Discussion and Conclusion}\label{sec:conclusion}

Distributional learning algorithms such as GAMLSS and deep distributional networks have been used successfully for probabilistic electricity price forecasting \cite[PEPF, see e.g.][]{muniain2020probabilistic,hirsch2024online,marcjasz2023distributional}. However, even for univariate distributions, these models are computationally expensive. At the same time, the literature on probabilistic electricity price forecasting has largely focused on modeling the hourly marginal distributions only, leaving the dependence structure neglected. Against this background, we develop an online estimation algorithm for multivariate distributional regression models, \secondreview{making the use of these algorithms feasible on a standard laptop, even for high-dimensional problems such as the 24-dimensional distribution of electricity prices}. We benchmark our implementation in a forecasting study for the German day-ahead electricity market and thereby provide the first study exclusively focused on online learning for multivariate PEPF. 

\review{
    \thirdreview{
        \fourthreview{Our results provide indication that for multivariate probabilistic prediction of the price distribution the dependence structure cannot be neglected. 
        However, we see trade-offs in terms of predictive performance between and within model groups and scoring rules. First, we see that the benchmark models perform better in the point and marginal prediction.  
        For the the LEAR-based benchmark models and the univariate online distributional regression (ODR), including the dependence structure improves multivariate predictive performance across all metrics. 
        For the multivariate online distributional regression (MODR), we note an improvement in the LS, the DSS and the joint coverage, while the marginal performance, the VS and ES degrade.  
        }%
    We assess calibration by the coverage of joint prediction bands \citep[JPB, see][]{staszewskabystrova2011bootstrap}, where the proposed method provides robust performance.
    }%
    Additionally, we highlight the importance of regularization to avoid overfitting in a high-dimensional setting and conduct two experiments to validate the regularization of the scale matrix. Distributional regression models are interpretable and therefore allow us to discuss the economic interpretation of the time-varying dependence structure.

We analyze the trade-off between computation time and forecasting accuracy. Building an efficient frontier between accuracy and computation time we show that online learning, compared to repeated batch fitting, yields better results for a given computation budget -- with speed-ups of 2-3 orders of magnitude.} \thirdreview{Thereby, the algorithm not only enables streaming data applications, but also allows for faster experimentation and back testing of distributional regression models.}
\review{Our algorithm is implemented in a fairly generic manner, allowing e.g. for different distributional assumptions and keeping a familiar, \texttt{sklearn}-like API to facilitate the usage by other researchers and data scientists \citep{pedregosa2011scikit} and contributed the implementation to the \texttt{ondil} package \citep{hirsch2024online}.\footnote{See: \url{https://github.com/simon-hirsch/ondil}}}

Our research opens multiple avenues for future work. First, further research on the driving forces of the dependence structure in the German electricity market is necessary to improve the forecasting performance and guide decision-making processes in electricity trading. Modeling the dependence structure in electricity markets is a rather open field and has implications beyond forecasting, concerning also risk and portfolio management and asset optimization \citep{pena2024hedging,lohndorf2023value, beykirch2022bidding, beykirch2024value}. 
\thirdreview{From an algorithmic perspective,
further speed improvements could be achieved by distributed and parallel computing.}
\secondreview{We outline a potential strategy for future research in Section \ref{sec:online_estimation}.} 
\thirdreview{Another interesting avenue is (online) model selection, as the models are complex} and can be prone to overfitting. Lastly, due to the generic nature of our implementation, the usage for other high-dimensional forecasting problems such as probabilistic wind, solar, and load forecasting can be explored.

\FloatBarrier
\section*{Acknowledgments}

Simon Hirsch is employed as an industrial PhD student by Statkraft Trading GmbH and gratefully acknowledges the support and funding received. This work contains the author’s opinions and does not necessarily reflect Statkraft’s position. Simon Hirsch is grateful for helpful discussions with Florian Ziel, Daniel Gruhlke, Christoph Hanck and the participants of the IWSM 2025 in Limerick. 

\section*{Declaration of Interest}

The author declares that he has no known competing financial interests or personal relationships that could have appeared to influence the work reported in this paper.

\section*{Data Statement} 

The data used in this paper has been provided by \cite{lipiecki2024postprocessing} and cannot be shared further by the author. Therefore, for any requests with respect to the data, please contact \cite{lipiecki2024postprocessing} directly. A shorter data set of the same explanatory variables is available at the \texttt{GitHub} repository of \cite{marcjasz2023distributional}.\footnote{See: \url{https://github.com/gmarcjasz/distributionalnn}.} For the differences between both data sets please consult \cite[][Section 2 and Section 5.1.3]{lipiecki2024postprocessing}. 

\section*{Generative AI Statement}

During the preparation of this work the author used generative AI tools such as ChatGPT and GitHub Copilot in order to improve the quality of the language and of the code \review{(esp. automatic PR reviews, documentation, creation of figures)}. After using this tool/service, the author reviewed and edited the content as needed and takes full responsibility for the content of the publication

\bibliographystyle{elsarticle-harv} 
\bibliography{references}

@Article{browell2022covariance,
  author    = {Browell, Jethro and Gilbert, Ciaran and Fasiolo, Matteo},
  journal   = {Electric Power Systems Research},
  title     = {Covariance structures for high-dimensional energy forecasting},
  year      = {2022},
  pages     = {108446},
  volume    = {211},
  publisher = {Elsevier},
}

@article{gabriel1962ante,
  title={Ante-dependence analysis of an ordered set of variables},
  author={Gabriel, KR},
  journal={The Annals of Mathematical Statistics},
  pages={201--212},
  year={1962},
  publisher={JSTOR}
}

@article{umlauf2025scalable,
  title={Scalable estimation for structured additive distributional regression},
  author={Umlauf, Nikolaus and Seiler, Johannes and Wetscher, Mattias and Simon, Thorsten and Lang, Stefan and Klein, Nadja},
  journal={Journal of Computational and Graphical Statistics},
  volume={34},
  number={2},
  pages={601--617},
  year={2025},
  publisher={Taylor \& Francis}
}

@article{dickey1979distribution,
  title={Distribution of the estimators for autoregressive time series with a unit root},
  author={Dickey, David A and Fuller, Wayne A},
  journal={Journal of the American statistical association},
  volume={74},
  number={366a},
  pages={427--431},
  year={1979},
  publisher={Taylor \& Francis}
}

@article{cheung1995lag,
  title={Lag order and critical values of the augmented Dickey--Fuller test},
  author={Cheung, Yin-Wong and Lai, Kon S},
  journal={Journal of Business \& Economic Statistics},
  volume={13},
  number={3},
  pages={277--280},
  year={1995},
  publisher={Taylor \& Francis}
}

@Article{chen2025probabilistic,
  author  = {Chen, Jieyu and Lerch, Sebastian and Schienle, Melanie and Serafin, Tomasz and Weron, Rafa{\l}},
  journal = {arXiv preprint arXiv:2506.00044},
  title   = {Probabilistic intraday electricity price forecasting using generative machine learning},
  year    = {2025},
}

@article{buchweitz2025asymmetric,
  title={Asymmetric penalties underlie proper loss functions in probabilistic forecasting},
  author={Buchweitz, Erez and Romano, Jo{\~a}o Vitor and Tibshirani, Ryan J},
  journal={arXiv preprint arXiv:2505.00937},
  year={2025}
}

@article{welford1962note,
  title={Note on a method for calculating corrected sums of squares and products},
  author={Welford, Barry Payne},
  journal={Technometrics},
  volume={4},
  number={3},
  pages={419--420},
  year={1962},
  publisher={Taylor \& Francis}
}

@software{kevin_sheppard_2024_14035889,
  author       = {Kevin Sheppard and
                  Stanislav Khrapov and
                  Gábor Lipták and
                  Rick van Hattem and
                  mikedeltalima and
                  Joren Hammudoglu and
                  Rob Capellini and
                  alejandro-cermeno and
                  Snyk bot and
                  Hugle and
                  esvhd and
                  Alex Fortin and
                  JPN and
                  Matt Judell and
                  Ryan Russell and
                  Weiliang Li and
                  645775992 and
                  Austin Adams and
                  jbrockmendel and
                  LGTM Migrator and
                  M. Rabba and
                  Michael E. Rose and
                  Nikolay Tretyak and
                  Tom Rochette and
                  UNO Leo and
                  Xavier RENE-CORAIL and
                  Xin Du and
                  Burak Çelik},
  title        = {bashtage/arch: Release 7.2},
  month        = nov,
  year         = 2024,
  publisher    = {Zenodo},
  version      = {v7.2.0},
  doi          = {10.5281/zenodo.14035889},
  url          = {https://doi.org/10.5281/zenodo.14035889},
}

@article{ziel2018day,
  title={Day-ahead electricity price forecasting with high-dimensional structures: Univariate vs. multivariate modeling frameworks},
  author={Ziel, Florian and Weron, Rafa{\l}},
  journal={Energy Economics},
  volume={70},
  pages={396--420},
  year={2018},
  publisher={Elsevier}
}

@Article{casella2021choice,
  author    = {Casella, Francesco and Bachmann, Bernhard},
  journal   = {Applied Mathematics and Computation},
  title     = {On the choice of initial guesses for the Newton-Raphson algorithm},
  year      = {2021},
  pages     = {125991},
  volume    = {398},
  publisher = {Elsevier},
}

@Article{kornerup2006choosing,
  author    = {Kornerup, Peter and Muller, Jean-Michel},
  journal   = {Theoretical computer science},
  title     = {Choosing starting values for certain Newton--Raphson iterations},
  year      = {2006},
  number    = {1},
  pages     = {101--110},
  volume    = {351},
  publisher = {Elsevier},
}

@Article{staszewskabystrova2011bootstrap,
  author    = {Staszewska-Bystrova, Anna},
  journal   = {Journal of Forecasting},
  title     = {Bootstrap prediction bands for forecast paths from vector autoregressive models},
  year      = {2011},
  number    = {8},
  pages     = {721--735},
  volume    = {30},
  publisher = {Wiley Online Library},
}

@Article{luetkepohl2015comparison,
  author    = {L{\"u}tkepohl, Helmut and Staszewska-Bystrova, Anna and Winker, Peter},
  journal   = {International Journal of Forecasting},
  title     = {Comparison of methods for constructing joint confidence bands for impulse response functions},
  year      = {2015},
  number    = {3},
  pages     = {782--798},
  volume    = {31},
  publisher = {Elsevier},
}

@Article{serafin2022trading,
  author    = {Serafin, Tomasz and Marcjasz, Grzegorz and Weron, Rafa{\l}},
  journal   = {Energy Economics},
  title     = {Trading on short-term path forecasts of intraday electricity prices},
  year      = {2022},
  pages     = {106125},
  volume    = {112},
  publisher = {Elsevier},
}

@article{pena2024hedging,
  title={Hedging renewable power purchase agreements},
  author={Pe{\~n}a, Juan Ignacio and Rodr{\'\i}guez, Rosa and Mayoral, Silvia},
  journal={Energy Strategy Reviews},
  volume={55},
  pages={101513},
  year={2024},
  publisher={Elsevier}
}

@article{lohndorf2023value,
  title={The value of coordination in multimarket bidding of grid energy storage},
  author={L{\"o}hndorf, Nils and Wozabal, David},
  journal={Operations research},
  volume={71},
  number={1},
  pages={1--22},
  year={2023},
  publisher={INFORMS}
}

@article{friedman2008sparse,
  title={Sparse inverse covariance estimation with the graphical lasso},
  author={Friedman, Jerome and Hastie, Trevor and Tibshirani, Robert},
  journal={Biostatistics},
  volume={9},
  number={3},
  pages={432--441},
  year={2008},
  publisher={Oxford University Press}
}

@Article{viehmann2017state,
  author    = {Viehmann, Johannes},
  journal   = {Zeitschrift f{\"u}r Energiewirtschaft},
  title     = {State of the German short-term power market},
  year      = {2017},
  number    = {2},
  pages     = {87--103},
  volume    = {41},
  publisher = {Springer},
}

@article{lago2021forecasting,
  title={Forecasting day-ahead electricity prices: A review of state-of-the-art algorithms, best practices and an open-access benchmark},
  author={Lago, Jesus and Marcjasz, Grzegorz and De Schutter, Bart and Weron, Rafa{\l}},
  journal={Applied Energy},
  volume={293},
  pages={116983},
  year={2021},
  publisher={Elsevier}
}

@inproceedings{zimmerman1997structured,
  title={Structured antedependence models for longitudinal data},
  author={Zimmerman, Dale L and N{\'u}{\~n}ez-Ant{\'o}n, Vicente},
  booktitle={Modelling longitudinal and spatially correlated data},
  pages={63--76},
  year={1997},
  organization={Springer}
}

@Article{zimmerman1998computational,
  author    = {Zimmerman, Dale L and Nu{\'u}{\~n}ez-ant{\'o}n, Vicente and El-Barmi, Hammou},
  journal   = {Journal of Statistical Computation and Simulation},
  title     = {Computational aspects of likelihood-based estimation of first-order antedependence models},
  year      = {1998},
  number    = {1},
  pages     = {67--84},
  volume    = {60},
  publisher = {Taylor \& Francis},
}

@article{groll2019lasso,
  title={LASSO-type penalization in the framework of generalized additive models for location, scale and shape},
  author={Groll, Andreas and Hambuckers, Julien and Kneib, Thomas and Umlauf, Nikolaus},
  journal={Computational Statistics \& Data Analysis},
  volume={140},
  pages={59--73},
  year={2019},
  publisher={Elsevier}
}

@article{gioia2022additive,
  title={Additive covariance matrix models: modeling regional electricity net-demand in Great Britain},
  author={Gioia, Vincenzo and Fasiolo, Matteo and Browell, Jethro and Bellio, Ruggero},
  journal={Journal of the American Statistical Association},
  volume={120},
  number={549},
  pages={107--119},
  year={2025},
  publisher={Taylor \& Francis}

}

@article{gioia2025scalable,
  title={Scalable Fitting Methods for Multivariate Gaussian Additive Models with Covariate-dependent Covariance Matrices},
  author={Gioia, Vincenzo and Fasiolo, Matteo and Bellio, Ruggero and Wood, Simon N},
  journal={arXiv preprint arXiv:2504.03368},
  year={2025}
}

@article{hirsch2024online,
  title={Online Distributional Regression},
  author={Hirsch, Simon and Berrisch, Jonathan and Ziel, Florian},
  journal={arXiv preprint arXiv:2407.08750},
  year={2024}
}

@Article{pedregosa2011scikit,
  author    = {Pedregosa, Fabian and Varoquaux, Ga{\"e}l and Gramfort, Alexandre and Michel, Vincent and Thirion, Bertrand and Grisel, Olivier and Blondel, Mathieu and Prettenhofer, Peter and Weiss, Ron and Dubourg, Vincent and others},
  journal   = {the Journal of machine Learning research},
  title     = {Scikit-learn: Machine learning in Python},
  year      = {2011},
  pages     = {2825--2830},
  volume    = {12},
  publisher = {JMLR. org},
}

@article{ziel2021gamlss,
  title={gamlss. lasso: Extra Lasso-Type Additive Terms for GAMLSS},
  author={Ziel, F and Muniain, P and Stasinopoulos, M},
  journal={R package version},
  pages={1--0},
  year={2021}
}

@article{muniain2020probabilistic,
  title={Probabilistic forecasting in day-ahead electricity markets: Simulating peak and off-peak prices},
  author={Muniain, Peru and Ziel, Florian},
  journal={International Journal of Forecasting},
  volume={36},
  number={4},
  pages={1193--1210},
  year={2020},
  publisher={Elsevier}
}

@article{pourahmadi2011covariance,
  title={Covariance Estimation: The GLM and Regularization Perspectives},
  author={Pourahmadi, Mohsen},
  journal={Statistical Science},
  volume={26},
  number={3},
  pages={369--387},
  year={2011},
  publisher={Citeseer}
}

@article{rigby2005generalized,
  title={Generalized additive models for location, scale and shape},
  author={Rigby, Robert A and Stasinopoulos, D Mikis},
  journal={Journal of the Royal Statistical Society Series C: Applied Statistics},
  volume={54},
  number={3},
  pages={507--554},
  year={2005},
  publisher={Oxford University Press}
}

@InProceedings{janke2020probabilistic,
  author       = {Janke, Tim and Steinke, Florian},
  booktitle    = {2020 International Conference on Probabilistic Methods Applied to Power Systems (PMAPS)},
  title        = {Probabilistic multivariate electricity price forecasting using implicit generative ensemble post-processing},
  year         = {2020},
  organization = {IEEE},
  pages        = {1--6},
}

@Article{agakishiev2025multivariate,
  author    = {Agakishiev, Ilyas and H{\"a}rdle, Wolfgang Karl and Kopa, Milos and Kozmik, Karel and Petukhina, Alla},
  journal   = {Energy Economics},
  title     = {Multivariate probabilistic forecasting of electricity prices with trading applications},
  year      = {2025},
  pages     = {108008},
  volume    = {141},
  publisher = {Elsevier},
}

@article{maciejowska2024multiple,
  title={Multiple split approach--multidimensional probabilistic forecasting of electricity markets},
  author={Maciejowska, Katarzyna and Nitka, Weronika},
  journal={arXiv preprint arXiv:2407.07795},
  year={2024}
}

@InProceedings{beykirch2022bidding,
  author       = {Beykirch, Mario and Janke, Tim and Steinke, Florian},
  booktitle    = {2022 17th International Conference on Probabilistic Methods Applied to Power Systems (PMAPS)},
  title        = {Bidding and scheduling in energy markets: Which probabilistic forecast do we need?},
  year         = {2022},
  organization = {IEEE},
  pages        = {1--6},
}

@Article{beykirch2024value,
  author    = {Beykirch, Mario and Bott, Andreas and Janke, Tim and Steinke, Florian},
  journal   = {IEEE Transactions on Power Systems},
  title     = {The value of probabilistic forecasts for electricity market bidding and scheduling under uncertainty},
  year      = {2024},
  publisher = {IEEE},
}

@article{rugamer2024semi,
  title={Semi-structured distributional regression},
  author={R{\"u}gamer, David and Kolb, Chris and Klein, Nadja},
  journal={The American Statistician},
  volume={78},
  number={1},
  pages={88--99},
  year={2024},
  publisher={Taylor \& Francis}
}

@article{klein2023deep,
  title={Deep distributional time series models and the probabilistic forecasting of intraday electricity prices},
  author={Klein, Nadja and Smith, Michael Stanley and Nott, David J},
  journal={Journal of Applied Econometrics},
  volume={38},
  number={4},
  pages={493--511},
  year={2023},
  publisher={Wiley Online Library}
}

@article{klein2021marginally,
  title={Marginally calibrated deep distributional regression},
  author={Klein, Nadja and Nott, David J and Smith, Michael Stanley},
  journal={Journal of Computational and Graphical Statistics},
  volume={30},
  number={2},
  pages={467--483},
  year={2021},
  publisher={Taylor \& Francis}
}

@article{bille2023forecasting,
  title={Forecasting electricity prices with expert, linear, and nonlinear models},
  author={Bill{\'e}, Anna Gloria and Gianfreda, Angelica and Del Grosso, Filippo and Ravazzolo, Francesco},
  journal={International Journal of Forecasting},
  volume={39},
  number={2},
  pages={570--586},
  year={2023},
  publisher={Elsevier}
}

@article{angelosante2010online,
  title={Online adaptive estimation of sparse signals: Where RLS meets the $\ell_1$-norm},
  author={Angelosante, Daniele and Bazerque, Juan Andr{\'e}s and Giannakis, Georgios B},
  journal={IEEE Transactions on signal Processing},
  volume={58},
  number={7},
  pages={3436--3447},
  year={2010},
  publisher={IEEE}
}

@article{cai2012adaptive,
  title={Adaptive Covariance Matrix Estimation through block thresholding},
  author={Cai, T Tony and Yuan, Ming},
  journal={The Annals of Statistics},
  volume={40},
  number={4},
  pages={2014--2042},
  year={2012}
}

@InProceedings{dasgupta2007line,
  author       = {Dasgupta, Sanjoy and Hsu, Daniel},
  booktitle    = {International Conference on Computational Learning Theory},
  title        = {On-line estimation with the multivariate Gaussian distribution},
  year         = {2007},
  organization = {Springer},
  pages        = {278--292},
}

@inproceedings{landgrebe2020online,
  title={Online mixed missing value imputation using gaussian copula},
  author={Landgrebe, Eric and Udell, Madeleine and others},
  booktitle={ICML Workshop on the Art of Learning with Missing Values (Artemiss)},
  year={2020}
}

@online{dexter2024probabilistic,
  author = {{Dexter Energy}},
  title = {Probabilistic price forecasts for short-term trade optimization},
  year = {2024},
  url = {https://dexterenergy.ai/news/probabilistic-price-forecasts-for-short-term-trade-optimization/},
  urldate = {2025-02-20}
}

@article{arbenz2013bayesian,
  title={Bayesian copulae distributions, with application to operational risk management—some comments},
  author={Arbenz, Philipp},
  journal={Methodology and computing in applied probability},
  volume={15},
  pages={105--108},
  year={2013},
  publisher={Springer}
}

@inproceedings{marcotte2023regions,
  title={Regions of reliability in the evaluation of multivariate probabilistic forecasts},
  author={Marcotte, {\'E}tienne and Zantedeschi, Valentina and Drouin, Alexandre and Chapados, Nicolas},
  booktitle={International Conference on Machine Learning},
  pages={23958--24004},
  year={2023},
  organization={PMLR}
}

@Article{ziel2019multivariate,
  author  = {Ziel, Florian and Berk, Kevin},
  journal = {arXiv preprint arXiv:1910.07325},
  title   = {Multivariate forecasting evaluation: On sensitive and strictly proper scoring rules},
  year    = {2019},
}

@Article{gneiting2007strictly,
  author    = {Gneiting, Tilmann and Raftery, Adrian E},
  journal   = {Journal of the American statistical Association},
  title     = {Strictly proper scoring rules, prediction, and estimation},
  year      = {2007},
  number    = {477},
  pages     = {359--378},
  volume    = {102},
  publisher = {Taylor \& Francis},
}

@Book{stasinopoulos2024generalized,
  author    = {Stasinopoulos, Mikis D and Kneib, Thomas and Klein, Nadja and Mayr, Andreas and Heller, Gillian Z},
  publisher = {Cambridge University Press},
  title     = {Generalized additive models for location, scale and shape: a distributional regression approach, with applications},
  year      = {2024},
  volume    = {56},
}

@Article{klein2024distributional,
  author    = {Klein, Nadja},
  journal   = {Annual Review of Statistics and Its Application},
  title     = {Distributional regression for data analysis},
  year      = {2024},
  volume    = {11},
  publisher = {Annual Reviews},
}

@Software{zanetta2024scoringrules,
  author = {Francesco Zanetta and Sam Allen},
  title  = {Scoringrules: a python library for probabilistic forecast evaluation},
  url    = {https://github.com/frazane/scoringrules},
  year   = {2024},
}

@Article{diebold2015comparing,
  author    = {Diebold, Francis X},
  journal   = {Journal of Business \& Economic Statistics},
  title     = {Comparing predictive accuracy, twenty years later: A personal perspective on the use and abuse of Diebold--Mariano tests},
  year      = {2015},
  number    = {1},
  pages     = {1--1},
  volume    = {33},
  publisher = {Taylor \& Francis},
}

@Article{diebold2002comparing,
  author    = {Diebold, Francis X and Mariano, Robert S},
  journal   = {Journal of Business \& economic statistics},
  title     = {Comparing predictive accuracy},
  year      = {2002},
  number    = {1},
  pages     = {134--144},
  volume    = {20},
  publisher = {Taylor \& Francis},
}

@article{pinson2013discrimination,
  title={Discrimination ability of the energy score},
  author={Pinson, Pierre and Tastu, Julija},
  year={2013},
  publisher={Technical University of Denmark}
}

@Article{scheuerer2015variogram,
  author  = {Scheuerer, Michael and Hamill, Thomas M},
  journal = {Monthly Weather Review},
  title   = {Variogram-based proper scoring rules for probabilistic forecasts of multivariate quantities},
  year    = {2015},
  number  = {4},
  pages   = {1321--1334},
  volume  = {143},
}

@article{alexander2024evaluating,
  title={Evaluating the discrimination ability of proper multi-variate scoring rules},
  author={Alexander, Carol and Coulon, Michael and Han, Yang and Meng, Xiaochun},
  journal={Annals of Operations Research},
  volume={334},
  number={1},
  pages={857--883},
  year={2024},
  publisher={Springer}
}

@Article{dawid1999coherent,
  author    = {Dawid, A Philip and Sebastiani, Paola},
  journal   = {Annals of Statistics},
  title     = {Coherent dispersion criteria for optimal experimental design},
  year      = {1999},
  pages     = {65--81},
  publisher = {JSTOR},
}

@Article{gneiting2007probabilistic,
  author    = {Gneiting, Tilmann and Balabdaoui, Fadoua and Raftery, Adrian E},
  journal   = {Journal of the Royal Statistical Society Series B: Statistical Methodology},
  title     = {Probabilistic forecasts, calibration and sharpness},
  year      = {2007},
  number    = {2},
  pages     = {243--268},
  volume    = {69},
  publisher = {Oxford University Press},
}

@inproceedings{angelosante2009online,
  title={Online coordinate descent for adaptive estimation of sparse signals},
  author={Angelosante, Daniele and Bazerque, Juan Andres and Giannakis, Georgios B},
  booktitle={2009 IEEE/SP 15th Workshop on Statistical Signal Processing},
  pages={369--372},
  year={2009},
  organization={IEEE}
}

@inproceedings{pierrot2021adaptive,
  title={Adaptive generalized logit-normal distributions for wind power short-term forecasting},
  author={Pierrot, Amandine and Pinson, Pierre},
  booktitle={2021 IEEE Madrid PowerTech},
  pages={1--6},
  year={2021},
  organization={IEEE}
}

@Article{messner2019online,
  author    = {Messner, Jakob W and Pinson, Pierre},
  journal   = {International Journal of Forecasting},
  title     = {Online adaptive lasso estimation in vector autoregressive models for high dimensional wind power forecasting},
  year      = {2019},
  number    = {4},
  pages     = {1485--1498},
  volume    = {35},
  publisher = {Elsevier},
}

@article{kath2021conformal,
  title={Conformal prediction interval estimation and applications to day-ahead and intraday power markets},
  author={Kath, Christopher and Ziel, Florian},
  journal={International Journal of Forecasting},
  volume={37},
  number={2},
  pages={777--799},
  year={2021},
  publisher={Elsevier}
}

@Article{kolkmann2024modeling,
  author    = {Kolkmann, Sven and Ostmeier, Lars and Weber, Christoph},
  journal   = {Energy Economics},
  title     = {Modeling multivariate intraday forecast update processes for wind power},
  year      = {2024},
  pages     = {107890},
  volume    = {139},
  publisher = {Elsevier},
}

@Article{soerensen2022recent,
  author    = {Sørensen, Mikkel L. and Nystrup, Peter and Bjerregård, Mathias B. and Møller, Jan K. and Bacher, Peder and Madsen, Henrik},
  journal   = {WIREs Energy and Environment},
  title     = {Recent developments in multivariate wind and solar power forecasting},
  year      = {2022},
  issn      = {2041-840X},
  month     = oct,
  number    = {2},
  volume    = {12},
  doi       = {10.1002/wene.465},
  publisher = {Wiley},
}

@article{marcjasz2023distributional,
  title={Distributional neural networks for electricity price forecasting},
  author={Marcjasz, Grzegorz and Narajewski, Micha{\l} and Weron, Rafa{\l} and Ziel, Florian},
  journal={Energy Economics},
  volume={125},
  pages={106843},
  year={2023},
  publisher={Elsevier}
}

@article{brusaferri2024nbmlss,
  title={NBMLSS: probabilistic forecasting of electricity prices via Neural Basis Models for Location Scale and Shape},
  author={Brusaferri, Alessandro and Ramin, Danial and Ballarino, Andrea},
  journal={arXiv preprint arXiv:2411.13921},
  year={2024}
}

@article{brusaferri2024line,
  title={On-line conformalized neural networks ensembles for probabilistic forecasting of day-ahead electricity prices},
  author={Brusaferri, Alessandro and Ballarino, Andrea and Grossi, Luigi and Laurini, Fabrizio},
  journal={Applied Energy},
  volume={398},
  pages={126412},
  year={2025},
  publisher={Elsevier}
}

@article{gibbs2021adaptive,
  title={Adaptive conformal inference under distribution shift},
  author={Gibbs, Isaac and Candes, Emmanuel},
  journal={Advances in Neural Information Processing Systems},
  volume={34},
  pages={1660--1672},
  year={2021}
}

@article{gibbs2024conformal,
  title={Conformal inference for online prediction with arbitrary distribution shifts},
  author={Gibbs, Isaac and Cand{\`e}s, Emmanuel J},
  journal={Journal of Machine Learning Research},
  volume={25},
  number={162},
  pages={1--36},
  year={2024}
}

@inproceedings{zaffran2022adaptive,
  title={Adaptive conformal predictions for time series},
  author={Zaffran, Margaux and F{\'e}ron, Olivier and Goude, Yannig and Josse, Julie and Dieuleveut, Aymeric},
  booktitle={International Conference on Machine Learning},
  pages={25834--25866},
  year={2022},
  organization={PMLR}
}

@article{lipiecki2024postprocessing,
  title={Postprocessing of point predictions for probabilistic forecasting of day-ahead electricity prices: The benefits of using isotonic distributional regression},
  author={Lipiecki, Arkadiusz and Uniejewski, Bartosz and Weron, Rafa{\l}},
  journal={Energy Economics},
  volume={139},
  pages={107934},
  year={2024},
  publisher={Elsevier}
}

@article{bjerregaard2021introduction,
  title={An introduction to multivariate probabilistic forecast evaluation},
  author={Bjerreg{\aa}rd, Mathias Blicher and M{\o}ller, Jan Kloppenborg and Madsen, Henrik},
  journal={Energy and AI},
  volume={4},
  pages={100058},
  year={2021},
  publisher={Elsevier}
}

@article{mashlakov2021assessing,
  title={Assessing the performance of deep learning models for multivariate probabilistic energy forecasting},
  author={Mashlakov, Aleksei and Kuronen, Toni and Lensu, Lasse and Kaarna, Arto and Honkapuro, Samuli},
  journal={Applied Energy},
  volume={285},
  pages={116405},
  year={2021},
  publisher={Elsevier}
}

@article{han2023probabilistic,
    title={Probabilistic Multivariate Time Series Forecasting and Robust Uncertainty Quantification with Applications in Electricity Price Prediction},
    author={Han, Jie},
    journal={Industrial, Manufacturing, and Systems Engineering Dissertations. University of Texas at Arlington.},
    volume={187},
    year={2023}
}

@article{kneib2023rage,
  title={Rage against the mean--a review of distributional regression approaches},
  author={Kneib, Thomas and Silbersdorff, Alexander and S{\"a}fken, Benjamin},
  journal={Econometrics and Statistics},
  volume={26},
  pages={99--123},
  year={2023},
  publisher={Elsevier}
}

@article{serinaldi2011distributional,
  title={Distributional modeling and short-term forecasting of electricity prices by generalized additive models for location, scale and shape},
  author={Serinaldi, Francesco},
  journal={Energy Economics},
  volume={33},
  number={6},
  pages={1216--1226},
  year={2011},
  publisher={Elsevier}
}

@article{grothe2023point,
  title={From point forecasts to multivariate probabilistic forecasts: The Schaake shuffle for day-ahead electricity price forecasting},
  author={Grothe, Oliver and K{\"a}chele, Fabian and Kr{\"u}ger, Fabian},
  journal={Energy Economics},
  volume={120},
  pages={106602},
  year={2023},
  publisher={Elsevier}
}

@Article{berrisch2024multivariate,
  author    = {Berrisch, Jonathan and Ziel, Florian},
  journal   = {International Journal of Forecasting},
  title     = {Multivariate probabilistic crps learning with an application to day-ahead electricity prices},
  year      = {2024},
  publisher = {Elsevier},
}

@Article{nowotarski2018recent,
  author    = {Nowotarski, Jakub and Weron, Rafa{\l}},
  journal   = {Renewable and Sustainable Energy Reviews},
  title     = {Recent advances in electricity price forecasting: A review of probabilistic forecasting},
  year      = {2018},
  pages     = {1548--1568},
  volume    = {81},
  publisher = {Elsevier},
}

@Article{green1984iteratively,
  author    = {Green, Peter J},
  journal   = {Journal of the Royal Statistical Society: Series B (Methodological)},
  title     = {Iteratively reweighted least squares for maximum likelihood estimation, and some robust and resistant alternatives},
  year      = {1984},
  number    = {2},
  pages     = {149--170},
  volume    = {46},
  publisher = {Wiley Online Library},
}

@Article{cole1992smoothing,
  author    = {Cole, Timothy J and Green, Pamela J},
  journal   = {Statistics in medicine},
  title     = {Smoothing reference centile curves: the LMS method and penalized likelihood},
  year      = {1992},
  number    = {10},
  pages     = {1305--1319},
  volume    = {11},
  publisher = {Wiley Online Library},
}

@article{salinas2019high,
  title={High-dimensional multivariate forecasting with low-rank gaussian copula processes},
  author={Salinas, David and Bohlke-Schneider, Michael and Callot, Laurent and Medico, Roberto and Gasthaus, Jan},
  journal={Advances in neural information processing systems},
  volume={32},
  year={2019}
}

@article{o2023variable,
  title={Variable selection using a smooth information criterion for distributional regression models},
  author={O’Neill, Meadhbh and Burke, Kevin},
  journal={Statistics and Computing},
  volume={33},
  number={3},
  pages={71},
  year={2023},
  publisher={Springer}
}

@Article{kock2023truly,
  title={Truly multivariate structured additive distributional regression},
  author={Kock, Lucas and Klein, Nadja},
  journal={Journal of Computational and Graphical Statistics},
  volume={34},
  number={4},
  pages={1189--1201},
  year={2025},
  publisher={Taylor \& Francis}
}

@Article{muschinski2022cholesky,
  author    = {Muschinski, Thomas and Mayr, Georg J and Simon, Thorsten and Umlauf, Nikolaus and Zeileis, Achim},
  journal   = {Econometrics and Statistics},
  title     = {Cholesky-based multivariate Gaussian regression},
  year      = {2022},
  publisher = {Elsevier},
}

@InProceedings{zhao2022online,
  author    = {Zhao, Yuxuan and Landgrebe, Eric and Shekhtman, Eliot and Udell, Madeleine},
  booktitle = {Proceedings of the AAAI Conference on Artificial Intelligence},
  title     = {Online missing value imputation and change point detection with the gaussian copula},
  year      = {2022},
  number    = {8},
  pages     = {9199--9207},
  volume    = {36},
}

\newpage
\appendix

\secondreview{
    \section{Overfitting and Regularization}\label{sec:overfitting}

    Modeling all elements of the scale matrix can lead to overfitting, especially in the high dimensional setting of energy markets. To counter this, we have proposed a path-based regularization approach in Section~\ref{sec:path_based_estimation}. Figure \ref{fig:overfitting} shows the results of a small experiment on the initial training set. We estimate the MODR-OLS-MCD-$\Sigma$ in a 8-fold cross-validation setting with 100 out-of-sample days without early stopping. We monitor the in-sample and  out-of-sample LS for the number of off-diagonals of $\mat{L}$ modeled in the path-wise estimation of $\mat{\Omega} = \transpose{\mat{L}}\mat{D}\mat{L}$. We see that the out-of-sample LS barely increases after modeling the first off-diagonal and starts to degrade after the 4th to 5th off-diagonal is included in the model for $\mat{\Omega}$. At the same time, the number of model coefficients increases with the amount of modeled elements. The information-criterion based early stopping suggests to include one (BIC and HQC) respectively four (AIC) off-diagonals. Furthermore, Figure~\ref{fig:path_based_estimation} shows the out-of-sample LS for the MODR-OLS-MCD-$\Sigma$ model for different levels of path-based regularization, which are well-aligned with the in-sample analysis, as the LS plateaus after including the first off-diagonal and starts to increase after the 2nd off-diagonal. In combination with the results from the previous section, we see that our approach allows for parsimonious, yet interpretable time-varying modeling of the dependence structure by exploiting the natural ordering of the hours in the day-ahead market.

    \begin{figure}[htb]
        \centering
        \includegraphics[width=0.6\textwidth]{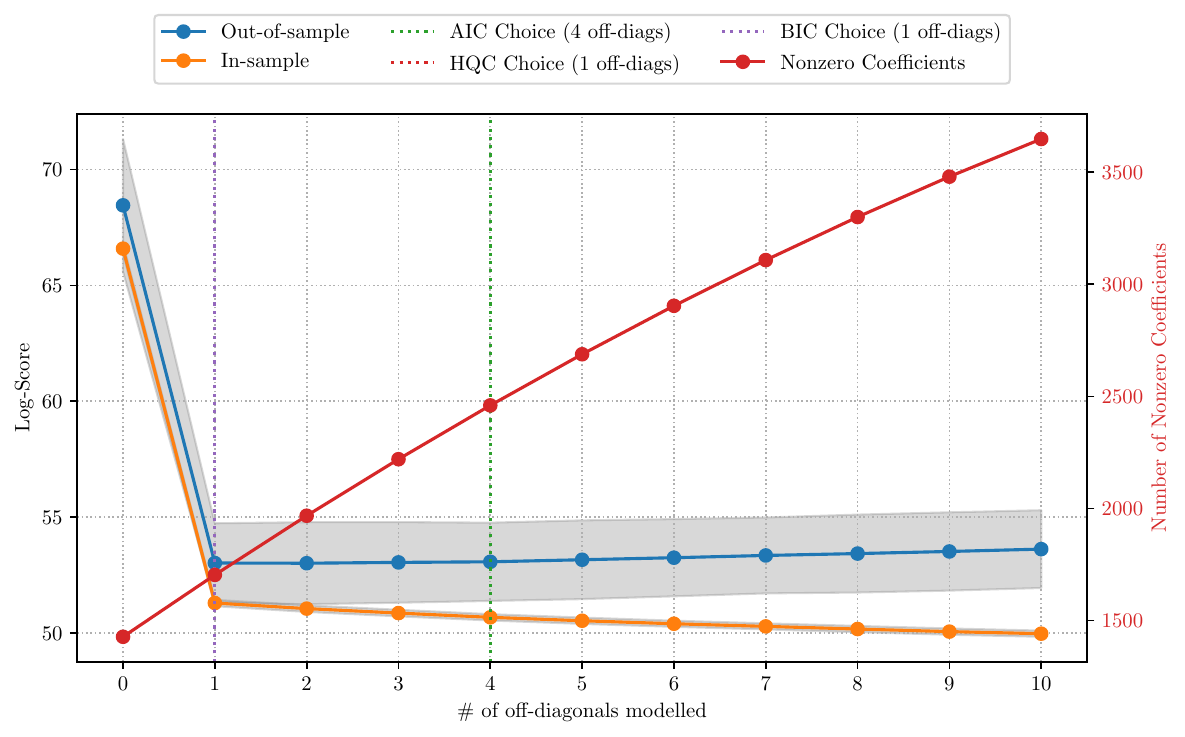}%
        \includegraphics[width=0.4\linewidth]{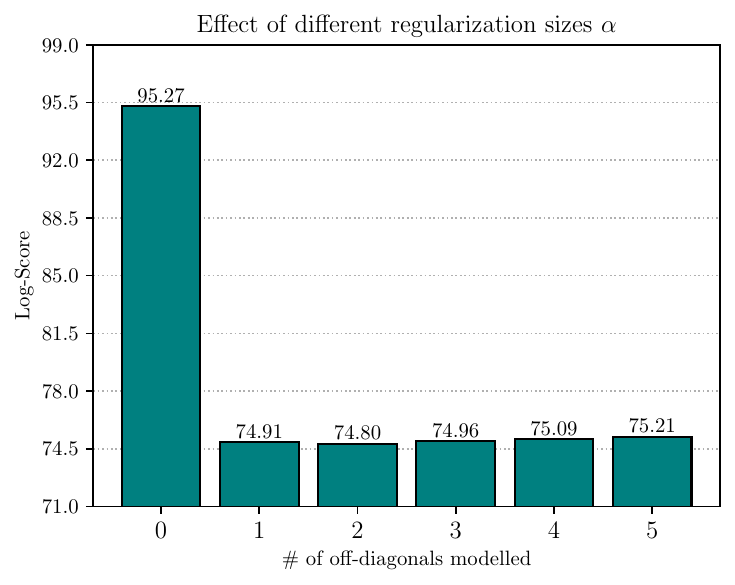}
        \caption{Analysis of Overfitting. 
            The left plot shows the result of the cross-validation study. We estimate the MODR-OLS-MCD-$\Sigma$ in a 8-fold cross-validation setting with 100 out-of-sample days without early stopping and monitor the in-sample and out-of-sample LS. Confidence bands are 95\%-confidence intervals based on the standard deviation of the LS. The number of model coefficients increases with the amount of modeled elements. The information-criterion based early stopping suggests to include one (BIC, HQC) respectively four (AIC) off-diagonals. The right plot shows the out-of-sample LS for the MODR-OLS-MCD-$\Sigma$ model for different levels of path-based regularization.
        }
        \label{fig:overfitting}
    \end{figure}
}
\FloatBarrier

\section{Additional Figures}\label{app:figures}

\begin{figure}
    \centering
    \includegraphics[width=\linewidth]{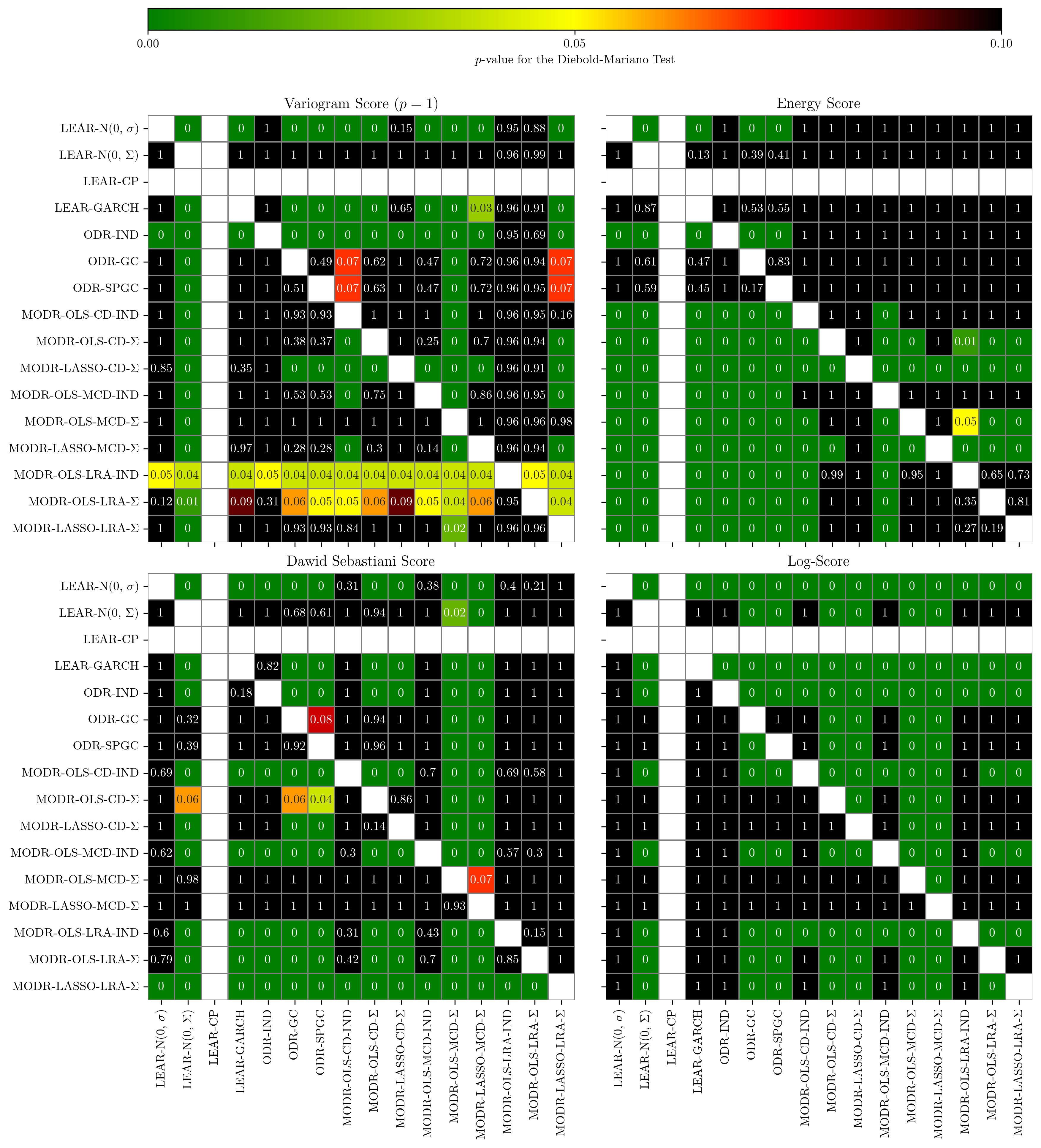}
    \caption{
        Diebold-Mariano Test Matrix. A $p$-value $p < 0.05$ implies that the forecasts given by a model on the column are significantly better than forecasts by a model on the row.
    }
    \label{fig:dm_tests}
\end{figure}

\clearpage


\section*{Supplementary material}

| | |
|---:|:---|
| AIC | Akaike Information Criterion |
| APS | Average Pinball Score |
| BIC | Bayesian Information Criterion |
| CD | Cholesky-Decomposition |
| CDF | Cumulative Density Function |
| DDNN | Distributional Deep Neural Networks |
| DSS | Dawid-Sebastiani Score |
| EPF | Electricity Price Forecasting |
| ES | Energy Score |
| GAMLSS | Generalized Additive Models for Location, Scale and Shape |
| GLM | Generalized Linear Model |
| HQC | Hannan-Quinn Criterion |
| IC | Information Criterion |
| IRLS | Iteratively Reweighted Least Squares |
| LASSO | Least Absolute Shrinkage and Selection Operator |
| LARX | LASSO-estimated AutoRegressive Model with eXogenous variables |
| LRA | Low-Rank Approximation |
| LS | Log-Score (= negative log-likelihood) |
| MAE | Mean Absolute Error |
| OCD | Online Coordinate Descent |
| OLS | Ordinary Least Squares |
| PDF | Probability Density Function |
| PEPF | Probabilistic Electricity Price Forecasting |
| PIT | Probability Integral Transformation |
| RMSE | Root Mean Squared Error |
| RS | Rigby & Stasinopolous (Algorithm) |
| VS | Variogram Score |

Table 6: Abbreviations used in the Paper.

# A Appendix

## A.1 Abbreviations

## A.2 Hyperparameters for the Multivariate Distributional Regression Model

To align with the reproducible research best practices, as described in e.g Lago et al. (2021), we publish the reproduction code on `GitHub` at https://github.com/simon-hirsch/online-mv-distreg, allowing for full reproducibility of all experiments. Additionally, we take the following paragraph to describe the hyperparameters of the model:

- *Information criteria and model selection:* We use the AIC for the multivariate distributional regression model and run the online coordinate descent on an exponential grid of 50 $\lambda$ values. We employ fast model selection based on the first derivatives for the CD-based models.

- *Link functions:* We use the identity link for the location for all models. For the CD-based distributional models, we use the `InverseSoftPlusLink`. For the LRA-based models, we employ the

29

`SqrtLink` for the diagonal matrix **A** as initial experiments showed a more robust convergence behavior and the `IdentityLink` for the matrix **V**. For the degrees of freedom $\nu$, we employ the `LogShiftTwoLink`, which ensures that $\nu > 2$ and hence the covariance matrix is positive definite.

- *Early stopping:* We employ early stopping for the path-based regularization of the scale matrix if the AIC does not improve, as described in Section 2.5. We limit the number of off-diagonals for the CD-based parameterization to max 6, however note that the algorithm breaks after fitting 2-3 off-diagonals. We do not limit the number of columns fitted in the LRA-based model and note that the algorithm breaks after fitting the full rank-2 matrix **V**.

- *Number of iterations, step-size and dampening:* We dampen the estimation in the first iteration for the scale parameters only. We generally allow for a maximum of 30 inner and 10 outer iterations in the initial fit and the update steps.

## A.3 Derivation of Equation 10 and 11 for Newton-Raphson Scoring

We aim to calculate $\partial \ell / \partial \eta$ and $\partial^2 \ell / \partial \eta^2$ for the calculation of the score and weight vectors (see Eq. 10 and Eq. 11) using the partial derivatives of the log-likelihood with respect to the distribution parameter (resp. the coordinate of the distribution parameter in case of matrix-valued parameters), $\partial \ell / \partial \theta$ and $\partial^2 \ell / \partial \theta^2$. For continuous, twice differentiable link functions $\eta = g(\theta)$, we have

$$\frac{\partial \ell}{\partial \eta} = \frac{\partial \ell}{\partial \theta} \left( \frac{\partial g(\theta)}{\partial \theta} \right)^{-1} \tag{32}$$

and

$$\frac{\partial^2 \ell}{\partial \eta^2} = \frac{\partial \left( \frac{\partial \ell}{\partial \theta} \left( \frac{\partial g(\theta)}{\partial \theta} \right)^{-1} \right)}{\partial \eta} = \frac{\partial \left( \frac{\partial \ell}{\partial \theta} \left( \frac{\partial g(\theta)}{\partial \theta} \right)^{-1} \right)}{\partial \theta} \left( \frac{\partial g(\theta)}{\partial \theta} \right)^{-1}$$

and by the quotient rule, we have

$$\frac{\partial^2 \ell}{\partial \eta^2} = \left( \frac{\frac{\partial^2 \ell}{\partial \theta^2} \left( \frac{\partial g(\theta)}{\partial \theta} \right) - \frac{\partial \ell}{\partial \theta} \left( \frac{\partial^2 g(\theta)}{\partial \theta^2} \right)}{\left( \frac{\partial g(\theta)}{\partial \theta} \right)^2} \left( \frac{\partial g(\theta)}{\partial \theta} \right)^{-1} \right) \tag{33}$$

and the simplification

$$\frac{\partial^2 \ell}{\partial \eta^2} = \left( \frac{\partial^2 \ell}{\partial \theta^2} \frac{\partial g(\theta)}{\partial \theta} - \frac{\partial \ell}{\partial \theta} \frac{\partial^2 g(\theta)}{\partial \theta^2} \right) \left( \frac{\partial g(\theta)}{\partial \theta} \right)^{-3} \tag{34}$$

concludes the derivation ∎

## A.4 Partial derivatives of the multivariate Gaussian Distribution

The probability density function of the multivariate normal distribution of dimension $D$ is given by:

$$f(\mathbf{y} \mid \boldsymbol{\mu}, \boldsymbol{\Sigma}) = \frac{1}{(2\pi)^{D/2} |\boldsymbol{\Sigma}|^{1/2}} \exp\left( -\frac{1}{2} (\mathbf{y} - \boldsymbol{\mu})^\top \boldsymbol{\Sigma}^{-1} (\mathbf{y} - \boldsymbol{\mu}) \right) \tag{35}$$

with the location or mean vector $\boldsymbol{\mu}$ and the scale respectively covariance matrix $\boldsymbol{\Sigma}$. We parameterize the PDF in terms of the inverse scale matrix $\boldsymbol{\Omega} = \boldsymbol{\Sigma}^{-1}$:

$$f(\mathbf{y} \mid \boldsymbol{\mu}, \boldsymbol{\Sigma}) = \frac{1}{(2\pi)^{D/2}} |\boldsymbol{\Omega}|^{1/2} \exp\left(-\frac{1}{2}(\mathbf{y} - \boldsymbol{\mu})^\top \boldsymbol{\Omega} (\mathbf{y} - \boldsymbol{\mu})\right) \tag{36}$$

and calculate the log-likelihood as $\ell(\mathbf{y} \mid \boldsymbol{\mu}, \boldsymbol{\Omega}) = \log(f(\mathbf{y} \mid \boldsymbol{\mu}, \boldsymbol{\Omega}))$, which reads:

$$\ell(\mathbf{y} \mid \boldsymbol{\mu}, \boldsymbol{\Sigma}) = -\frac{D}{2}\log(2\pi) - \frac{1}{2}\log(|\boldsymbol{\Sigma}|) - \frac{1}{2}(\mathbf{y} - \boldsymbol{\mu})^\top \boldsymbol{\Sigma}^{-1} (\mathbf{y} - \boldsymbol{\mu}) \tag{37}$$

and parameterize the inverse covariance matrix through the Cholesky-decomposition $\boldsymbol{\Sigma} = \mathbf{A}\mathbf{A}^\top$ and $\boldsymbol{\Sigma}^{-1} = \boldsymbol{\Omega} = (\mathbf{A}^{-1})^\top (\mathbf{A}^{-1})$, which yields:

$$\ell(\mathbf{y} \mid \boldsymbol{\mu}, \mathbf{A}^{-1}) = -\frac{D}{2}\log(2\pi) - \log(|\mathbf{A}^{-1}|) - \frac{1}{2}\mathbf{z}^\top \mathbf{z} \tag{38}$$

where $\mathbf{z} = \mathbf{A}^{-1}(\mathbf{y} - \boldsymbol{\mu})$ and $z = \mathbf{z}^\top \mathbf{z}$. The first derivatives with respect to the elements of $\boldsymbol{\mu}$ and $\mathbf{A}^{-1}$ are given in Muschinski et al. (2022) and read

$$\frac{\partial \ell}{\partial \boldsymbol{\mu}_i} = \sum_{k=0}^{D} \boldsymbol{\Omega}_{ik}(\mathbf{y}_k - \boldsymbol{\mu}_k) \tag{39}$$

$$\frac{\partial \ell}{\partial (\mathbf{A}^{-1})_{ij}} = \frac{1}{(\mathbf{A}^{-1})_{ij}} - (\mathbf{y}_i - \boldsymbol{\mu}_i) \sum_{k=0}^{D}(\mathbf{y}_k - \boldsymbol{\mu}_k)(\mathbf{A}^{-1})_{kj} \tag{40}$$

and the second derivatives are given by

$$\frac{\partial \ell^2}{\partial \boldsymbol{\mu}_i^2} = -\boldsymbol{\Omega}_{ii} \tag{41}$$

$$\frac{\partial \ell^2}{\partial (\mathbf{A}^{-1})_{ij}^2} = -\frac{1}{(\mathbf{A}^{-1})_{ij}^2} - (\mathbf{y}_i - \boldsymbol{\mu}_i)^2 \tag{42}$$

For the low-rank approximation, we parameterize Equation 36 in terms of the low-rank approximation $\boldsymbol{\Omega} = \mathbf{D} + \mathbf{V}^\top \mathbf{V}$, which yields:

$$\ell(\mathbf{y} \mid \boldsymbol{\mu}, \mathbf{U}, \mathbf{V}) = -\frac{D}{2}\log(2\pi) - \log(|(\mathbf{D} + \mathbf{V}^\top \mathbf{V})^{-1}|) - \frac{1}{2}(\mathbf{y} - \boldsymbol{\mu})^\top (\mathbf{D} + \mathbf{V}^\top \mathbf{V})(\mathbf{y} - \boldsymbol{\mu}) \tag{43}$$

we note that the derivatives with respect to the elements of the mean vector $\boldsymbol{\mu}_i$ remain the same:

$$\frac{\partial \ell}{\partial \boldsymbol{\mu}_i} = \sum_{k=0}^{D} \boldsymbol{\Omega}_{ik}(\mathbf{y}_k - \boldsymbol{\mu}_k) \tag{44}$$

$$\frac{\partial \ell^2}{\partial \boldsymbol{\mu}_i^2} = -\boldsymbol{\Omega}_{ii} \tag{45}$$

and the derivatives with respect to the elements of $\mathbf{D}_{ii}$ are given by:

$$\frac{\partial \ell}{\partial \mathbf{D}_{ii}} = \frac{1}{2}\left(\boldsymbol{\Sigma}_{ii} - (\mathbf{y}_k - \boldsymbol{\mu}_k)^2\right) \tag{46}$$

$$\frac{\partial \ell^2}{\partial \mathbf{D}_{ii}^2} = -\frac{1}{2}\boldsymbol{\Sigma}_{ii}^2. \tag{47}$$

The partial derivatives with respect to the elements of $\mathbf{V}$ are given by:

$$\frac{\partial \ell}{\partial \mathbf{V}_{ij}} = \sum_{k=0}^{D} \mathbf{\Sigma}_{ik} \mathbf{V}_{kj} \sum_{k=0}^{D} (\mathbf{y}_k - \boldsymbol{\mu}_k)(\mathbf{y}_i - \boldsymbol{\mu}_i) \mathbf{V}_{kj} \tag{48}$$

$$\frac{\partial^2 \ell}{\partial \mathbf{V}_{ij}^2} = \mathbf{\Sigma}_{ij} - \sum_{k=0}^{D}\sum_{q=0}^{D} \mathbf{\Sigma}_{ii} \mathbf{V}_{qj} \mathbf{\Sigma}_{qk} \mathbf{V}_{kj} - \sum_{k=0}^{D}\sum_{q=0}^{D} \mathbf{\Sigma}_{iq} \mathbf{V}_{qj} \mathbf{\Sigma}_{ik} \mathbf{V}_{kj} \tag{49}$$

$$- \left( (\mathbf{y}_i - \boldsymbol{\mu}_i)^2 - \left( \sum_{k=0}^{D} (\mathbf{y}_k - \boldsymbol{\mu}_k)(\mathbf{y}_i - \boldsymbol{\mu}_i) \mathbf{V}_{kj} \right)^2 \right)$$

which concludes the derivation of the partial derivatives ∎

### A.5 Partial derivatives of the multivariate *t*-distribution

The probability density function (PDF) of the multivariate *t*-distribution of dimension $D$ is given by:

$$f(\mathbf{y} \mid \boldsymbol{\mu}, \Sigma, \nu) = \frac{\Gamma((\nu+D)/2)}{\Gamma(\nu/2)\nu^{D/2}\pi^{D/2}|\boldsymbol{\Sigma}|^{1/2}} \left(1 + \frac{1}{\nu}(\mathbf{y}-\boldsymbol{\mu})^\top \boldsymbol{\Sigma}^{-1}(\mathbf{y}-\boldsymbol{\mu})\right)^{-(\nu+D)/2}$$

with the location vector $\mu$, the shape matrix $\boldsymbol{\Sigma}$ and the degrees of freedom $\nu$. We parameterize the PDF in terms of the inverse shape matrix $\boldsymbol{\Omega} = \boldsymbol{\Sigma}^{-1}$:

$$f(\mathbf{y} \mid \boldsymbol{\mu}, \boldsymbol{\Omega}, \nu) = \frac{\Gamma((\nu+D)/2)}{\Gamma(\nu/2)\nu^{D/2}\pi^{D/2}} |\boldsymbol{\Omega}|^{1/2} \left(1 + \frac{1}{\nu}(\mathbf{y}-\boldsymbol{\mu})^\top \boldsymbol{\Omega}(\mathbf{y}-\boldsymbol{\mu})\right)^{-(\nu+D)/2}. \tag{50}$$

We start with the partial derivatives for the CD-based parametrization. We have the Choleksy-decomposition $\boldsymbol{\Sigma} = \mathbf{A}\mathbf{A}^\top$ and $\boldsymbol{\Sigma}^{-1} = \boldsymbol{\Omega} = (\mathbf{A}^{-1})^\top (\mathbf{A}^{-1})$, which yields:

$$f(\mathbf{y} \mid \boldsymbol{\mu}, \mathbf{A}^{-1}, \nu) = \frac{\Gamma((\nu+D)/2)}{\Gamma(\nu/2)\nu^{D/2}\pi^{D/2}} |(\mathbf{A}^{-1})| \left(1 + \frac{1}{\nu}(\mathbf{y}-\boldsymbol{\mu})^\top (\mathbf{A}^{-1})^\top (\mathbf{A}^{-1})(\mathbf{y}-\boldsymbol{\mu})\right)^{-(\nu+D)/2}.$$

Let us introduce some notation to simply the following derivatives. Define:

$$\mathbf{z} = \mathbf{A}^{-1}(\mathbf{y}-\boldsymbol{\mu}) \tag{51}$$

$$z = \mathbf{z}^\top \mathbf{z} \tag{52}$$

The log-likelihood is given by $\ell(\mathbf{y} \mid \boldsymbol{\mu}, \mathbf{A}^{-1}, \nu) = \log(f(\mathbf{y} \mid \boldsymbol{\mu}, \mathbf{A}^{-1}, \nu))$ and reads:

$$\ell(\mathbf{y} \mid \boldsymbol{\mu}, \mathbf{A}^{-1}, \nu) = \log\left(\frac{\Gamma((\nu+D)/2)}{\Gamma(\nu/2)\nu^{D/2}\pi^{D/2}}\right) + \log\left(|(\mathbf{A}^{-1})|\right) + \log\left(\left(1 + \frac{1}{\nu}(\mathbf{z}^T \mathbf{z})\right)^{-(\nu+D)/2}\right) \tag{53}$$

For the partial derivatives with respect to the elements of $\boldsymbol{\mu}$ and $\mathbf{A}^{-1}$, we notice that $\mathbf{z}^\top \mathbf{z}$ can be treated as a function of these elements and employ the chain rule. We see that:

$$\frac{\partial(\mathbf{z}^\top \mathbf{z})}{\partial \boldsymbol{\mu}_i} = 2 \sum_{j=1}^{D} \boldsymbol{\Omega}_{ij}(\mathbf{y}_j - \boldsymbol{\mu}_j) \tag{54}$$

$$\frac{\partial^2(\mathbf{z}^\top \mathbf{z})}{\partial \boldsymbol{\mu}_i^2} = -2\boldsymbol{\Omega}_{ij} \tag{55}$$

$$\frac{\partial(\mathbf{z}^\top \mathbf{z})}{\partial (\mathbf{A}^{-1})_{ij}} = 2(\mathbf{y}_i - \boldsymbol{\mu}_i) \sum_{m=1}^{M=j} (\mathbf{y}_m - \boldsymbol{\mu}_m)(\mathbf{A}^{-1})_{mj} \tag{56}$$

$$\frac{\partial(\mathbf{z}^\top \mathbf{z})^2}{\partial (\mathbf{A}^{-1})_{ij}^2} = (\mathbf{y}_i - \boldsymbol{\mu}_i)^2 \tag{57}$$

The chain rule for the last term of Equation 53 yields:

$$\left[\log\left(\left(1 + \frac{1}{\nu}(\mathbf{z}^\top \mathbf{z})\right)^{-(\nu+D)/2}\right)\right]' = \frac{(D+\nu)}{2((\mathbf{z}^\top \mathbf{z}) + \nu)}(\mathbf{z}^\top \mathbf{z})' \tag{58}$$

$$\left[\log\left(\left(1 + \frac{1}{\nu}(\mathbf{z}^\top \mathbf{z})\right)^{-(\nu+D)/2}\right)\right]'' = -\frac{(D+\nu)(((\mathbf{z}^\top \mathbf{z}) + \nu)(\mathbf{z}^\top \mathbf{z})'' - ((\mathbf{z}^\top \mathbf{z})')^2)}{2((\mathbf{z}^\top \mathbf{z}) + \nu)^2} \tag{59}$$

and plugging in the according partial derivatives in Equations 54 to 57 and applying integration by parts for the remainder of Equation 53, we have:

$$\frac{\partial l}{\partial \boldsymbol{\mu}_i} = \frac{(D+\nu)}{2(z+\nu)}\left(2 \sum_{j=1}^{D} \boldsymbol{\Omega}_{ij}(\mathbf{y}_j - \boldsymbol{\mu}_j)\right) \tag{60}$$

$$\frac{\partial^2}{\partial \boldsymbol{\mu}_i^2} = -\frac{(D+\nu)\left((z+\nu)(-2\boldsymbol{\Omega}_{ij}) - \left(2\sum_{j=1}^{D} \boldsymbol{\Omega}_{ij}(\mathbf{y}_j - \boldsymbol{\mu}_j)\right)^2\right)}{2(z+\nu)^2} \tag{61}$$

$$\frac{\partial \ell}{\partial (\mathbf{A}^{-1})_{ij}} = \frac{1}{(\mathbf{A}^{-1})_{ij}}\mathbf{1}_{i=j} + \frac{(D+\nu)}{2(z+\nu)}\left(2(\mathbf{y}_i - \boldsymbol{\mu}_i) \sum_{m=1}^{M=i} (\mathbf{y}_m - \boldsymbol{\mu}_m)(\mathbf{A}^{-1})_{mj}\right) \tag{62}$$

$$\frac{\partial^2 l}{\partial (\mathbf{A}^{-1})ij^2} = -\frac{1}{(\mathbf{A}^{-1})_{ij}^2}\mathbf{1}_{i=j} - \frac{(D+\nu)\left((z+\nu)(\mathbf{y}_i - \boldsymbol{\mu}_i)^2 - \left(2\sum_{j=1}^{D} \boldsymbol{\Omega}_{ij}(\mathbf{y}_j - \boldsymbol{\mu}_j)\right)^2\right)}{2((\mathbf{z}^\top \mathbf{z}) + \nu)^2} \tag{63}$$

Where **1** is the indicator function for $i = j$, since the partial derivative of $\log\left(|\mathbf{A}^{-1}|\right)$ are only relevant for the partial derivatives of the diagonal elements of $\mathbf{A}^{-1}$. For the partial derivatives with respect to the degrees of freedom $\nu$, integration by parts yields:

$$\frac{\partial l}{\partial \nu} = -\frac{-\nu \, \text{digamma}(\frac{D+\nu}{2}) + D + \nu \, \text{digamma}(\frac{\nu}{2})}{2\nu} + \frac{1}{2}\left(\frac{z(D+\nu)}{\nu(\nu+z)} - \log\left(\frac{(\nu+z)}{\nu}\right)\right) \tag{64}$$

$$\frac{\partial^2 l}{\partial \nu^2} = \frac{1}{4}\left(\frac{2k}{\nu^2} + \text{trigamma}(\frac{D+\nu}{2}) - \text{trigamma}(\frac{\nu}{2})\right) + \frac{z(\nu z - D(2\nu+z))}{2\nu^2(\nu+z)^2}. \tag{65}$$

For the low-rank approximation, we follow a similar notation. The LRA is given by $\mathbf{\Omega} = \mathbf{D} + \mathbf{V}^\top\mathbf{V}$ and hence the PDF is given by:

$$f(\mathbf{y} \mid \boldsymbol{\mu}, \mathbf{D}, \mathbf{V}, \nu) = \frac{\Gamma((\nu + D)/2)}{\Gamma(\nu/2)\nu^{D/2}\pi^{D/2}}|\mathbf{D} + \mathbf{V}^\top\mathbf{V}|^{1/2}\left(1 + \frac{1}{\nu}(\mathbf{y}-\boldsymbol{\mu})^\top(\mathbf{D}+\mathbf{V}^\top\mathbf{V})(\mathbf{y}-\boldsymbol{\mu})\right)^{-(\nu+D)/2} \quad (66)$$

and the log-likelihood is given by

$$\ell(\mathbf{y} \mid \boldsymbol{\mu}, \mathbf{A}^{-1}, \nu) = \log\left(\frac{\Gamma((\nu+D)/2)}{\Gamma(\nu/2)\nu^{D/2}\pi^{D/2}}\right) + \log\left(|(\mathbf{D}+\mathbf{V}^\top\mathbf{V})|\right) + $$
$$\log\left(\left(1 + \frac{1}{\nu}(\mathbf{y}-\boldsymbol{\mu})^\top(\mathbf{D}+\mathbf{V}^\top\mathbf{V})(\mathbf{y}-\boldsymbol{\mu})\right)^{-(\nu+D)/2}\right) \quad (67)$$

we follow a similar strategy as above and see that the partial derivatives with respect to the elements of $\boldsymbol{\mu}$ and with respect to the degrees of freedom $\nu$ are the same as above:

$$\frac{\partial \ell}{\partial \boldsymbol{\mu}_i} = \frac{(D+\nu)}{2(z+\nu)}\left(2\sum_{j=1}^{D}\boldsymbol{\Omega}_{ij}(\mathbf{y}_j - \boldsymbol{\mu}_j)\right) \quad (68)$$

$$\frac{\partial^2 \ell}{\partial \boldsymbol{\mu}_i^2} = -\frac{(D+\nu)\left((z+\nu)(-2\boldsymbol{\Omega}_{ij}) - \left(2\sum_{j=1}^{D}\boldsymbol{\Omega}_{ij}(\mathbf{y}_j-\boldsymbol{\mu}_j)\right)^2\right)}{2(z+\nu)^2} \quad (69)$$

$$\frac{\partial l}{\partial \nu} = -\frac{-\nu\,\text{digamma}(\frac{D+\nu}{2}) + D + \nu\,\text{digamma}(\frac{\nu}{2})}{2\nu} + \frac{1}{2}\left(\frac{z(D+\nu)}{\nu(\nu+z)} - \log\left(\frac{(\nu+z)}{\nu}\right)\right) \quad (70)$$

$$\frac{\partial^2 l}{\partial \nu^2} = \frac{1}{4}\left(\frac{2k}{\nu^2} + \text{trigamma}(\frac{D+\nu}{2}) - \text{trigamma}(\frac{\nu}{2})\right) + \frac{z(\nu z - D(2\nu+z))}{2\nu^2(\nu+z)^2}. \quad (71)$$

where $z$ is now defined as

$$z = (\mathbf{y}-\boldsymbol{\mu})^\top(\mathbf{D}+\mathbf{V}^\top\mathbf{V})(\mathbf{y}-\boldsymbol{\mu}). \quad (72)$$

For the partial derivatives with respect to the elements of $\mathbf{D}$, we note that the partial derivatives of the second term are given by:

$$\frac{\partial}{\partial \mathbf{D}_{ii}} = \frac{1}{2}\boldsymbol{\Sigma}_{ii} \quad (73)$$

$$\frac{\partial}{\partial (\mathbf{D}_{ii})^2} = -\frac{1}{2}(\boldsymbol{\Sigma}_{ii})^2 \quad (74)$$

and the partial deriviatives of the third term are given by

$$\frac{\partial}{\partial \mathbf{D}_{ii}} = \frac{D+\nu}{2(z+\nu)}(\mathbf{y}_i - \boldsymbol{\mu}_i)^2 \quad (75)$$

$$\frac{\partial}{\partial \mathbf{D}_{ii}^2} = 0 \quad (76)$$

and the second defaults to 0. The partial derivatives of the log-likelihood with respect to the elements of $\mathbf{D}$ are hence given by:

$$\frac{\partial \ell}{\partial \mathbf{D}_{ii}} = \frac{1}{2}\boldsymbol{\Sigma}_{ii} - \frac{D+\nu}{2(z+\nu)}(\mathbf{y}_i - \boldsymbol{\mu}_i)^2 \quad (77)$$

$$\frac{\partial^2 \ell}{\partial \mathbf{D}_{ii}^2} = -\frac{1}{2}(\boldsymbol{\Sigma}_{ii})^2 - \frac{D+\nu}{2(z+\nu)^2}\left((\mathbf{y}_i - \boldsymbol{\mu}_i)^2\right)^2. \quad (78)$$

For the partial derivatives with respect to the elements of $\mathbf{V}$, we have a more complex formulation for the second term involving the determinant:

$$\frac{\partial}{\partial V_{ij}} = \tag{79}$$

$$\frac{\partial}{\partial V_{ij}^2} = \mathbf{\Sigma}_{ij} - \sum_{k=0}^{D}\sum_{q=0}^{D} \mathbf{\Sigma}_{ii}\mathbf{V}_{qj}\mathbf{\Sigma}_{qk}\mathbf{V}_{kj} - \sum_{k=0}^{D}\sum_{q=0}^{D} \mathbf{\Sigma}_{iq}\mathbf{V}_{qj}\mathbf{\Sigma}_{ik}\mathbf{V}_{kj} \tag{80}$$

and the partial derivatives of the third term are given by

$$\frac{\partial}{\partial V_{ij}} = \frac{D+\nu}{(z+\nu)}\sum_{k=0}^{D}(\mathbf{y}_k - \boldsymbol{\mu}_k)(\mathbf{y}_i - \boldsymbol{\mu}_i)\mathbf{V}_{kj} \tag{81}$$

$$\frac{\partial}{\partial V_{ij}^2} = \frac{D+\nu}{(z+\nu)^2}(\mathbf{y}_i - \boldsymbol{\mu}_i)^2 \tag{82}$$

and hence integration by parts again gives us, similiar to the partial derivatives for the multivariate normal distribution in Equation 48 and 49:

$$\frac{\partial \ell}{\partial \mathbf{V}_{ij}} = \sum_{k=0}^{D} \mathbf{\Sigma}_{ik}\mathbf{V}_{kj} + \frac{D+\nu}{(z+\nu)}\sum_{k=0}^{D}(\mathbf{y}_k - \boldsymbol{\mu}_k)(\mathbf{y}_i - \boldsymbol{\mu}_i)\mathbf{V}_{kj} \tag{83}$$

$$\frac{\partial^2 \ell}{\partial \mathbf{V}_{ij}^2} = \mathbf{\Sigma}_{ij} - \sum_{k=0}^{D}\sum_{q=0}^{D} \mathbf{\Sigma}_{ii}\mathbf{V}_{qj}\mathbf{\Sigma}_{qk}\mathbf{V}_{kj} - \sum_{k=0}^{D}\sum_{q=0}^{D} \mathbf{\Sigma}_{iq}\mathbf{V}_{qj}\mathbf{\Sigma}_{ik}\mathbf{V}_{kj} \tag{84}$$

$$- \frac{D+\nu}{(z+\nu)^2}\left((\mathbf{y}_i - \boldsymbol{\mu}_i)^2 - \left(\sum_{k=0}^{D}(\mathbf{y}_k - \boldsymbol{\mu}_k)(\mathbf{y}_i - \boldsymbol{\mu}_i)\mathbf{V}_{kj}\right)^2\right)$$

which concludes the derivation of the partial derivatives for the multivariate $t$-distribution ∎



\end{document}